\documentclass{article} 
\usepackage{iclr2024_conference,times}
\AtBeginDocument{\RenewCommandCopy\qty\SI}

\usepackage{amsmath,amsfonts,bm}

\def\eqref#1{equation~\ref{#1}}

\def\1{\bm{1}}

\def\vb{{\bm{b}}}

\DeclareMathAlphabet{\mathsfit}{\encodingdefault}{\sfdefault}{m}{sl}
\SetMathAlphabet{\mathsfit}{bold}{\encodingdefault}{\sfdefault}{bx}{n}

\usepackage{hyperref}
\usepackage{url}
\usepackage{tcolorbox}
\usepackage{wrapfig}

\usepackage{blindtext}
\usepackage{titlesec}
\usepackage{titletoc}

\title{Localizing and Editing Knowledge in \\ Text-to-Image Generative Models}

\author{Samyadeep Basu$^{1}$, Nanxuan Zhao$^{2}$, Vlad Morariu$^{2}$, Soheil Feizi*$^{1}$, Varun Manjunatha*$^{2}$ \\
$^{1}$: University of Maryland, $^{2}$: Adobe Research\\
\texttt{Correspondence to:sbasu12@umd.edu} \\
}

\usepackage{algorithm}
\usepackage{algpseudocode}

\usepackage{hyperref}
\usepackage{cleveref}
\crefname{figure}{Fig}{Figs}%
\crefname{algorithm}{Algorithm}{Algo}%

\usepackage{booktabs}
\usepackage{esrelation}
\usepackage{tikz}
\usepackage{graphics}

\usepackage{siunitx}
\usepackage{graphicx}
\usepackage{xcolor}
\usepackage{amsfonts} 
\usepackage{cleveref} 
\usepackage{amsmath}
\usepackage{siunitx,letltxmacro}
\LetLtxMacro{\svqty}{\qty}
\usepackage{physics}
\LetLtxMacro{\qty}{\svqty}
\AtBeginDocument{\RenewCommandCopy\qty\SI}
\usepackage{newfloat}
\usepackage{listings}
\usepackage{color, colortbl}
\definecolor{LightCyan}{rgb}{0.88,1,1}

\usepackage[T1]{fontenc}
\floatstyle{ruled}
\newfloat{listing}{tb}{lst}{}
\floatname{listing}{Listing}
\newcommand{\difffix}{\textsc{Diff-QuickFix}}

\newcommand{\violet}[1]{\textcolor{violet}{#1}}

\iclrfinalcopy 
\begin{document}

\maketitle

\begin{abstract}
Text-to-Image Diffusion Models such as Stable-Diffusion and Imagen have achieved unprecedented quality of photorealism with state-of-the-art FID scores on MS-COCO and other generation benchmarks. 
Given a caption, image generation requires fine-grained knowledge about attributes such as object structure, style, and viewpoint amongst others. {\it Where does this information reside in text-to-image generative models?} 
In our paper, we tackle this question and understand how knowledge corresponding to distinct visual attributes is stored in large-scale text-to-image diffusion models. We adapt Causal Mediation Analysis for text-to-image models and trace knowledge about distinct visual attributes to various (causal) components in the (i) UNet and (ii) text-encoder of the diffusion model. 
In particular, we show that unlike generative large-language models, knowledge about different attributes is not localized in isolated components, but is instead distributed amongst a set of components in the conditional UNet. These sets of components are often distinct for different visual attributes (e.g., {\it style} / {\it objects}).  
Remarkably, we find that the CLIP text-encoder in public text-to-image models such as Stable-Diffusion contains {\it only} one causal state across different visual attributes, and this is the first self-attention layer corresponding to the last subject token of the attribute in the caption. 
This is in stark contrast to the causal states in other language models which are often the mid-MLP layers. 
Based on this observation of {\it only} one causal state in the text-encoder, we introduce a fast, data-free model editing method \difffix{} which can effectively edit concepts (remove or update knowledge) in text-to-image models.~\difffix{} can edit (ablate) concepts in under a second with a closed-form update, providing a significant 1000x speedup and comparable editing performance to existing fine-tuning based editing methods.

\end{abstract}
\section{Introduction}
Text-to-Image generative models such as Stable-Diffusion~\citep{sd_main}, Imagen~\citep{saharia2022photorealistic} and DALLE~\citep{dalle}  have revolutionized conditional image generation in the last few years. These models have attracted a lot of attention due to their impressive image generation and editing capabilities, obtaining state-of-the-art FID scores on common generation benchmarks such as MS-COCO~\citep{coco}. Text-to-Image generation models are generally trained on billion-scale image-text pairs such as LAION-5B~\citep{schuhmann2022laion5b} which typically consist of a plethora of visual concepts encompassing color, artistic styles, objects, and famous personalities, amongst others. Prior works~\citep{carlini2023extracting, Somepalli_2023_CVPR, somepalli2023understanding} have shown that text-to-image models such as Stable-Diffusion memorize various aspects of the pre-training dataset. For example, given a caption from the LAION dataset, a model can generate an exact image from the training dataset corresponding to the caption in certain cases~\citep{carlini2023extracting}. These observations reinforce that some form of knowledge corresponding to visual attributes is stored in the parameter space of text-to-image model. 

When an image is generated, it possesses visual attributes such as (but not limited to) the presence of distinct objects with their own characteristics (such as color or texture), artistic style or scene viewpoint. This attribute-specific information is usually specified in the conditioning textual prompt to the UNet in text-to-image models which is used to pull relevant knowledge from the UNet to construct and subsequently generate an image. This leads to an important question: {\it How and where is knowledge corresponding to various visual attributes stored in text-to-image models?} 

In this work, we empirically study this question towards understanding how knowledge corresponding to different visual attributes is stored in text-to-image models, using Stable Diffusion\citep{sd_main} as a representative model. In particular, we adapt Causal Mediation Analysis~\citep{vig_nlp, pearl2013direct} for large-scale text-to-image diffusion models to identify specific causal components in the (i) UNet and (ii) the text-encoder where visual attribute knowledge resides. Previously, Causal Meditation Analysis has been used for understanding where factual knowledge is stored in LLMs. In particular, ~\citep{meng2023locating} find that factual knowledge is localized and stored in the mid-MLP layers of a LLM such as GPT-J~\citep{gpt-j}. Our work, however, paints a different picture - for multimodal text-to-image models, we specifically find that knowledge is not localized to one particular component. Instead, there exist various components in the UNet where knowledge is stored. However, each of these components store attribute information with a different efficacy and often different attributes have a distinct set of causal components where knowledge is stored. For e.g., for {\it style} -- we find that the first self-attention layer in the UNet stores {\it style} related knowledge, however it is not causally important for other attributes such as {\it objects}, {\it viewpoint} or {\it action}. To our surprise, we specifically find that the cross-attention layers are not causally important states and a significant amount of knowledge is in fact stored in components such as the ResNet blocks and the self-attention blocks. 

Remarkably, in the text-encoder, we find that knowledge corresponding to distinct attributes is strongly localized, contrary to the UNet. However unlike generative language models~\citep{meng2023locating} where the mid MLP layers are causal states, we find that the first self-attention layer is causal in the CLIP based text-encoders of public text-to-image generative models (e.g., Stable-Diffusion). 
\begin{figure*}
    \hskip -0.2cm
  \includegraphics[width=14.5cm, height=6.2cm]{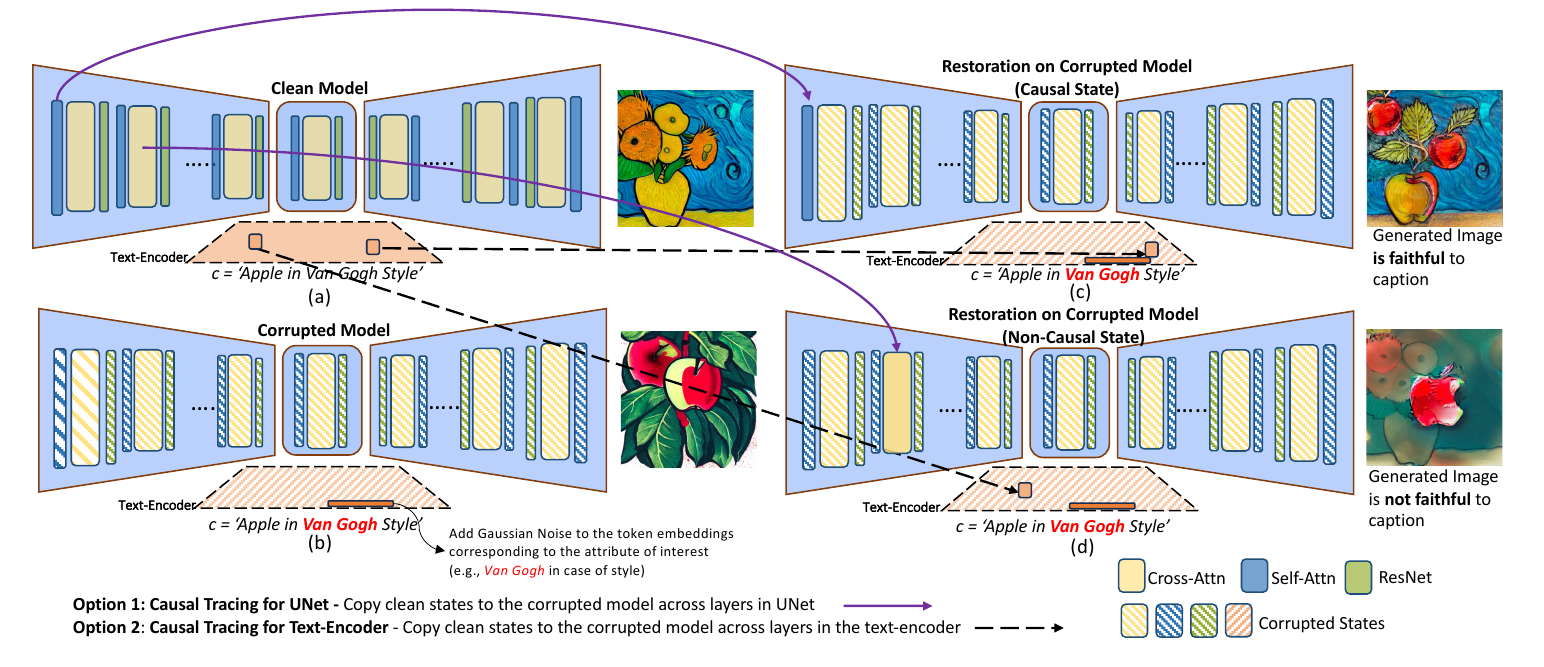}
  \vspace{-0.5cm}
    \caption{\label{teaser} \textbf{Causal Tracing in Text-to-Image Models for (i) UNet and (ii) Text-Encoder shows that knowledge location matters, i.e., restoring causal layers in a corrupted model causes the model to obey the prompt again, while restoring non-causal layers does not.}
    (a) {\it Clean Model}: We prompt a Stable-Diffusion model in the conventional way and generate an image as output. (b) {\it Corrupted Model}: Token embeddings corresponding to attribute of interest are corrupted, leading to a generated image that does not obey the prompt. (c) {\it Restored (Causal) Model}: Causal layer activations are now copied from the clean model to the corrupted model. We observe that the corrupted model can now generate images with high fidelity to the original caption. (d) {\it Restored (Non-Causal) Model}: Non-causal layer activations are copied from the clean model to the corrupted model, but we now observe that the generated image does not obey the prompt.
    Note that a single layer is copied at a time, and it can be from either the UNet (Option 1, \violet{solid violet arrow}) or the text-encoder (Option 2, broken black arrow).
    }%
    \vspace{-0.7cm}
\end{figure*}

Identification of local causal states in a given model has a crucial benefit: it allows for incorporating controlled edits to the model by updating {\it only} a tiny fraction of the model parameters without any fine-tuning. Using our observation that the text-encoder hosts {\it only} one localized causal state, we introduce a new data-free and fast model editing method - \difffix{} which can edit concepts in text-to-image models effectively using a closed-form update. In particular, we show that~\difffix{} can (i) remove copyrighted styles, (ii) trademarked objects as well as (iii) update stale knowledge 1000x faster than existing fine-tuning based editing methods such as~\citep{kumari2023ablating, gandikota2023erasing} with comparable or even better performance in some cases. 

In summary, our contributions are as follows:
\begin{itemize}
    \item We adapt Causal Mediation Analysis~\citep{pearl2013direct, vig_nlp} to large-scale text-to-image models (with Stable-Diffusion as a representative model), and use it to trace knowledge corresponding to various visual attributes in the UNet and text-encoder.
    \item We perform large-scale analysis of the identified causal components and shed light on the knowledge flow corresponding to various visual attributes in the UNet and the text-encoder. 
    \item Leveraging the interpretability observations of localized causal states in the text-encoder, we develop a light-weight method~\difffix{} which can edit various concepts in text-to-image models in under a second, 1000x faster than existing concept ablating methods~\cite{kumari2023ablating, gandikota2023erasing}.
    
\end{itemize}
\section{Related Works}
\textbf{Text-to-Image Diffusion Models. } In the last year, a large number of text-to-image models such as Stable-Diffusion~\citep{sd_main}, DALLE~\citep{dalle} , Imagen~\citep{saharia2022photorealistic} and others~\citep{balaji2023ediffi, chang2023muse, ding2022cogview2, kang2023scaling} have been released. In addition, the open-source community has released DeepFloyd\footnote{https://www.deepfloyd.ai} and Midjourney\footnote{https://www.midjourney.com/} which can generate photorealistic images given a text prompt. While most of these models operate in the latent space of the images, they differ in the text-encoder used. For e.g., Stable-Diffusion uses CLIP for the text-encoder, whereas Imagen uses T5. These text-to-image diffusion models have been used as a basis for various applications such as image-editing, semantic-segmentation, object-detection, image restoration and zero-shot classification. 

\textbf{Intepretability of Text-to-Image Models. } 
To our knowledge, few works delve into the mechanisms of large text-to-image models like Stable-Diffusion. DAAM~\citep{tang2022daam} interprets diffusion models by analyzing cross-attention maps between text tokens and images, emphasizing their semantic accuracy for interpretation. In contrast, our approach focuses on comprehending the inner workings of diffusion models by investigating the storage of visual knowledge related to different attributes. We explore various model layers beyond just the cross-attention layer.

\textbf{Editing Text-to-Image Models.}  
Understanding knowledge storage in diffusion models has significant implications for model editing. This ability to modify a diffusion model's behavior without retraining from scratch were first explored in Concept-Ablation~\citep{kumari2023ablating} and Concept-Erasure~\citep{gandikota2023erasing}. TIME~\citep{orgad2023editing} is another model editing method which translates between concepts by modifying the key and value matrices in cross-attention layers. However, the experiments in~\citep{orgad2023editing} do not specifically target removing or updating concepts such as those used in~\citep{kumari2023ablating, gandikota2023erasing}. We also acknowledge concurrent works~\citep{gandikota2023unified} and ~\citep{arad2023refact} use a closed-form update on the cross-attention layers and text-encoder respectively to ablate concepts. However, we note that our work focuses primarily on first understanding how knowledge is stored in text-to-image models and subsequently using this information to design a closed-form editing method for editing concepts.

\vspace{-0.4cm}
\section{Causal Tracing for Text-to-Image Generative Models}
In this section, we first provide a brief overview of diffusion models in~Sec.(\ref{background}). We then describe how causal tracing is adapted to multimodal diffusion models such as Stable-Diffusion.
\vspace{-0.3cm}
\subsection{Background}
\label{background}
Diffusion models are inspired by non-equilibrium thermodynamics and specifically aim to learn to denoise data through a number of steps. Usually, noise is added to the data following a Markov chain across multiple time-steps $t \in [0, T]$. Starting from an initial random real image $\vb{x}_{0}$, the noisy image at time-step $t$ is defined as $\vb{x}_{t} = \sqrt{\alpha_{t}}\vb{x}_{0} + \sqrt{(1-\alpha_{t})}\vb{\epsilon}$. In particular, $\alpha_{t}$ determines the strength of the random Gaussian noise and it gradually decreases as the time-step increases such that $\vb{x}_{T}\sim \mathcal{N}(0, I)$. The denoising network denoted by $\epsilon_{\theta}(\vb{x}_{t}, \vb{c}, t)$ is pre-trained to denoise the noisy image $\vb{x}_{t}$ to obtain $\vb{x}_{t-1}$. Usually, the conditional input $\vb{c}$ to the denoising network $\epsilon_{\theta}(.)$ is a text-embedding of a caption $c$ through a text-encoder $\vb{c} = v_{\gamma}(c)$ which is paired with the original real image $\vb{x}_{0}$. The pre-training objective for diffusion models can be defined as follows for a given image-text pair denoted by ($\vb{x}$, $\vb{c}$):
\begin{equation}
    \label{og_diffusion}
    \mathcal{L}(\vb{x}, \vb{c}) = \mathbb{E}_{\epsilon, t} || \epsilon - \epsilon_{\theta}(\vb{x}_{t}, \vb{c}, t) ||_{2}^{2},
\end{equation}
where $\theta$ is the set of learnable parameters. For better training efficiency, the noising as well as the denoising operation occurs in a latent space defined by $\vb{z} = \mathcal{E}(\vb{x})$~\cite{sd_main}. In this case, the pre-training objective learns to denoise in the latent space as denoted by:
\begin{equation}
    \label{og_diffusion_latent}
    \mathcal{L}(\vb{x}, \vb{c}) = \mathbb{E}_{\epsilon, t} || \epsilon - \epsilon_{\theta}(\vb{z}_{t}, \vb{c}, t) ||_{2}^{2},
\end{equation}
where $\vb{z}_{t} = \mathcal{E}(\vb{x}_{t})$ and $\mathcal{E}$ is an encoder such as VQ-VAE~\citep{oord2018neural}. During inference, where the objective is to synthesize an image given a text-condition $\vb{c}$, a random Gaussian noise $\vb{x}_{T} \sim \mathcal{N}(0,I)$ is iteratively denoised for a fixed range of time-steps in order to produce the final image.  We provide more details on the pre-training and inference steps in~\Cref{pretrain_details}. 
\subsection{Adapting Causal Tracing For Text-to-Image Diffusion Models}
\label{causal_adaptation}
Causal Mediation Analysis~\citep{pearl2013direct, vig_nlp} is a method from causal inference that studies the change in a response variable following an intervention on intermediate variables of interest (mediators). 
One can think of the internal model components (e.g., specific neurons or layer activations) as mediators along a directed acyclic graph between the input and output. For text-to-image diffusion models, we use Causal Mediation Analysis to trace the causal effects of these internal model components within the UNet and the text-encoder which contributes towards the generation of images with specific visual attributes (e.g., {\it objects, style}).  For example, we find the subset of model components in the text-to-image model which are causal for generating images with specific {\it objects}, {\it styles}, {\it viewpoints}, {\it action} or {\it color}.  

\textbf{Where is Causal Tracing Performed?} We identify the causal model components in both the UNet $\epsilon_{\theta}$ and the text-encoder $v_{\gamma}$. For $\epsilon_{\theta}$, we perform the causal tracing at the granularity of layers, whereas for the text-encoder, causal tracing is performed at the granularity of hidden states of the token embeddings in $\vb{c}$ across distinct layers. The UNet $\epsilon_{\theta}$ consists of 70 unique layers distributed amongst three types of blocks: (i) \texttt{down-block}; (ii) \texttt{mid-block} and (iii) \texttt{up-block}. Each of these blocks contain varying number of cross-attention layers, self-attention layers and residual layers. ~\cref{teaser} visualizes the internal states of the UNet and how causal tracing for knowledge attribution is performed.  
For the text-encoder $v_{\gamma}$, there are 12 blocks in total with each block consisting of a self-attention layer and a MLP layer (see~\cref{teaser}). 
We highlight that the text-encoder in text-to-image models such as Stable-Diffusion has a GPT-style architecture with a causal self-attention, though it's pre-trained without a language modeling objective. More details on the layers used in~\Cref{attribution_layers}.

Given a caption $c$, an image $\vb{x}$ is generated starting from some random Gaussian noise. This image $\vb{x}$ encapsulates the visual properties embedded in the caption $c$. For e.g., the caption $c$ can contain information corresponding from {\it objects} to {\it action} etc. We specifically identify distinct components in the UNet and the text-encoder which are causally responsible for these properties. 
\begin{figure*}
    \hskip -0.1cm
  \includegraphics[width=14.2cm, height=8.2cm]{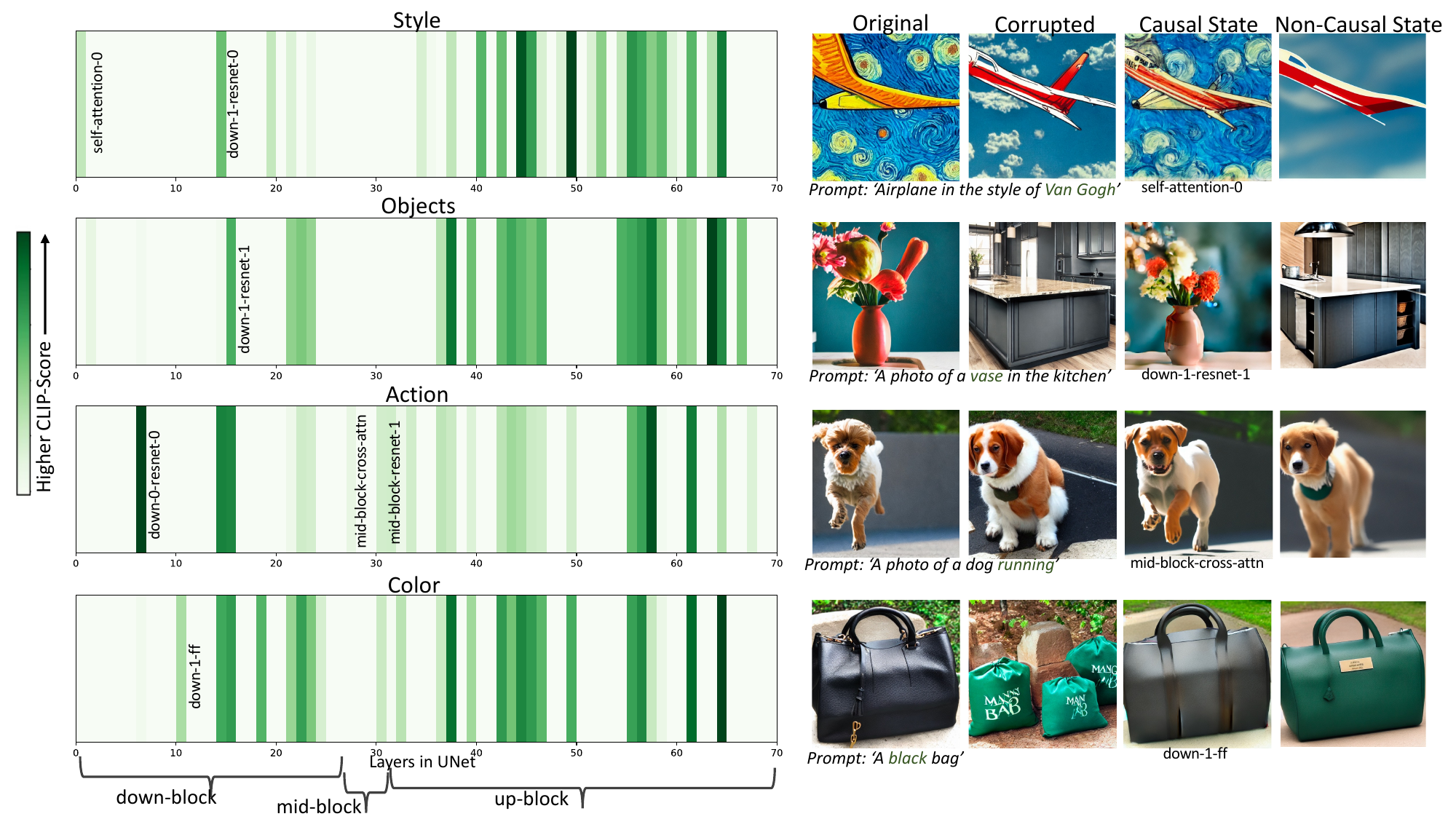}
  \vspace{-0.4cm}
    \caption{\label{tracing_1} \textbf{Causal Tracing Results for the UNet: Knowledge is Distributed.} The intensity of the bars indicate the \texttt{CLIP-Score} between the generated image (after causal intervention) and the original caption. For each attribute, we find that the causal states are distributed across the UNet and the distribution varies amongst distinct attributes. For e.g., self-attn in the first layer is causal for {\it style}, but not for {\it objects}, {\it action} or {\it color}. Similarly, mid-block cross-attn is causal for {\it action}, but not for the other attributes. On the right-side, we visualize the images generated by (i) Original model; (ii) Corrupted Model; (iii) Restored causal states and (iv) Restored non-causal states in the UNet for {\it style, action, object, color} attributes. 
    }%
    \vspace{-0.50cm}
\end{figure*}

\textbf{Creating the Probe Captions.} 
We primarily focus on four different visual attributes for causal tracing: (i) {\it objects}; (ii) {\it style}; (iii) {\it color}; and (iv) {\it action}. 
In particular, identifying the location of knowledge storage for {\it objects} and {\it style} can be useful to perform post-hoc editing of diffusion models to edit concepts (e.g., delete or update certain concepts). We provide the complete details about the probe dataset used for causal tracing in~\Cref{probe_dataset}. The probe dataset also contains additional captions for {\it viewpoint} and {\it count} attribute. However, we do not focus on them as often the generations from the unedited model are erroneous for these attributes (see~\Cref{trace_viewpoint_count} for details). 
\vspace{-0.3cm}
\subsection{Tracing Knowledge in UNet}
\label{trace_unet}
During inference, classifier-free guidance~\citep{ho2022classifierfree} is used to regulate image-generation by incorporating scores from the conditional and unconditional diffusion model at each of the time-steps. In particular, at each time-step, classifier-free guidance is used in the following way to combine the conditional ($\epsilon_{\theta}(\vb{z}_{t}, \vb{c}, t)$) and unconditional score estimates ($\epsilon_{\theta}(\vb{z}_{t},t)$) at each time-step $t$ to obtain the combined score denoted as $\hat{\epsilon}(\vb{z}_{t}, \vb{c}, t)$:
\begin{equation}
    \label{cf_free_guidance}
    \hat{\epsilon}_{\theta}(\vb{z_{t}}, \vb{c}, t) = \epsilon_{\theta}(\vb{z_{t}}, \vb{c}, t) + \alpha (\epsilon_{\theta}(\vb{z_{t}}, \vb{c}, t) - \epsilon_{\theta}(\vb{z_{t}}, t)), \hspace{1em}  \forall t\in [T, 1].
\end{equation}
This combined score is used to update the latent $\vb{z}_{t}$ using DDIM sampling~\citep{DBLP:journals/corr/abs-2010-02502} at each time-step iteratively to obtain the final latent code $\vb{z}_{0}$. 

To perform causal tracing on the UNet $\epsilon_{\theta}$~(see \Cref{teaser} for visualization), we perform a sequence of operations that is somewhat analogous to earlier work from ~\citep{meng2023locating} which investigated knowledge-tracing in large language models. We consider three types of model configurations: (i) a clean model $\epsilon_{\theta}$, where classifier-free guidance is used as default; (ii) a corrupted model $\epsilon_{\theta}^{corr}$, where the word embedding of the subject (e.g., {\it Van Gogh}) of a given attribute (e.g., {\it style}) corresponding to a caption $c$ is corrupted with Gaussian Noise; and, 
 (iii) a restored model $\epsilon_{\theta}^{restored}$, which is similar to $\epsilon_{\theta}^{corr}$ except that one of its layers is restored from the clean model at each time-step of the classifier-free guidance. 
Given a list of layers $\mathcal{A}$, let $a_{i} \in \mathcal{A}$ denote the $i^{th}$ layer whose importance needs to be evaluated. Let $\epsilon_{\theta}[a_{i}]$, $\epsilon_{\theta}^{corr}[a_{i}]$ and $\epsilon_{\theta}^{restored}[a_{i}]$ denote the activations of layer $a_{i}$. To find the importance of layer $a_{i}$  for a particular attribute embedded in a caption $c$, we perform the following replacement operation on the corrupted model $\epsilon_{\theta}^{corr}$ to obtain the restored model $\epsilon_{\theta}^{restored}$: 
\begin{equation}
    \label{restoration_operation}
    \epsilon_{\theta}^{restored}[a_{i}]: \epsilon_{\theta}^{corr}[a_{i}] = \epsilon_{\theta}[a_{i}].
\end{equation}
Next, we obtain the restored model by replacing the activations of layer $a_{i}$ of the corrupted model with those of the clean model to get a restored layer $\epsilon_{\theta}^{restored}[a_{i}]$. We run classifier-free guidance to obtain the combined score estimate:
\begin{equation}
    \label{cf_free_guidance_latent}
    \hat{\epsilon}_{\theta}^{restored}(\vb{z_{t}}, \vb{c}, t) = \epsilon_{\theta}^{restored}(\vb{z_{t}}, \vb{c}, t) + \alpha (\epsilon_{\theta}^{restored}(\vb{z_{t}}, \vb{c}, t) - \epsilon_{\theta}^{restored}(\vb{z_{t}}, t)), \hspace{1em}  \forall t\in [T, 1].
\end{equation}
The final latent $\vb{z}_{0}$ is obtained with the score  from~\Cref{cf_free_guidance_latent} at each time-step using DDIM~\citep{DBLP:journals/corr/abs-2010-02502} and passed through the VQ-VAE decoder to obtain the final image $\vb{x}_{0}^{restored}$.  
\begin{figure*}
    \hskip -0.35cm
  \includegraphics[width=14.5cm, height=4.0cm]{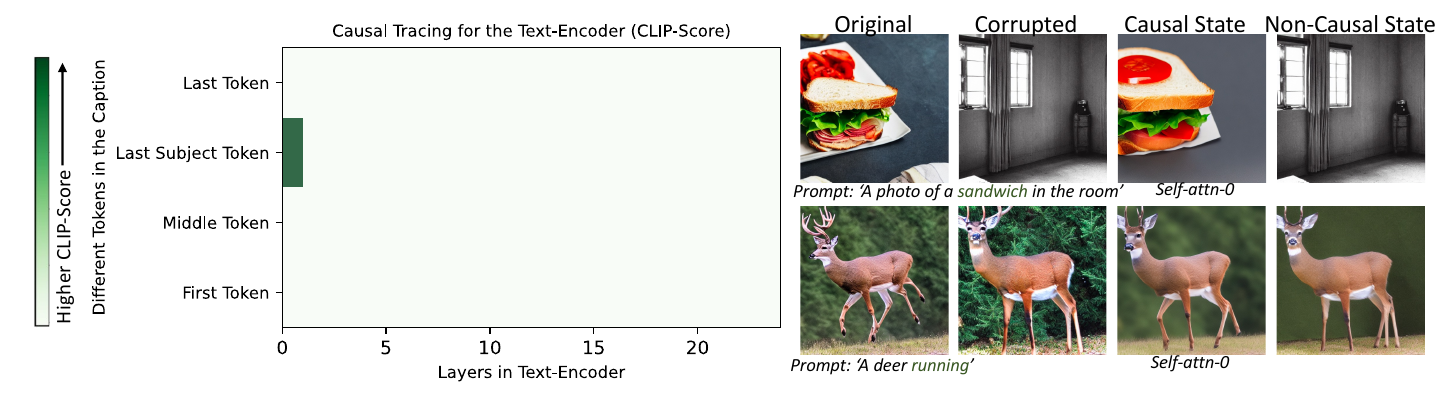}
  \vspace{-0.4cm}
    \caption{\label{text_tracing} \textbf{Causal Tracing in the Text-Encoder: Knowledge is Localized.} In the CLIP text-encoder used for Stable-Diffusion, we find the existence of {\it only} one causal state, which is the first self-attention layer corresponding to the last subject token. The \texttt{CLIP-Score}(Left) is computed across all the four visual attributes. Visualizations (Right) further illustrate that restoring the sole causal state (self-attn-0) leads to image generation with high fidelity to the original captions.}%
    \vspace{-0.3cm}
\end{figure*}
\subsection{Tracing Knowledge in the Text-Encoder}
\label{trace_text_encoder}
The text-encoder in public text-to-image models such as Stable-Diffusion is a CLIP-ViT-L/336px text-encoder~\cite{sd_main}. Similar to Sec.(\ref{trace_unet}), we define three states of the CLIP text-encoder: (i) Clean model denoted by $v_{\gamma}$; (ii) Corrupted model $v_{\gamma}^{corr}$ where the word embedding of the subject in a given caption $c$ is corrupted; (iii) Restored model $v_{\gamma}^{restored}$ which is similar to $v_{\gamma}^{corr}$ except that one of its layers is copied from $v_{\gamma}$. Similar to Sec.(\ref{trace_unet}), to find the effect of the layer $a_{i} \in \mathcal{A}$, where $\mathcal{A}$ consists of all the layers to probe in the CLIP text-encoder:
\begin{equation}
    v_{\gamma}^{restored}[a_{i}]: v_{\gamma}^{corr}[a_{i}] = v_{\gamma}[a_{i}],
\end{equation}
We then use the restored text-encoder $v_{\gamma}^{restored}$ with classifier-free guidance to obtain the final score estimate:
\begin{equation}
    \label{cf_free_guidance_text}
    \hat{\epsilon}_{\theta}(\vb{z_{t}}, \vb{c}', t) = \epsilon_{\theta}(\vb{z_{t}}, \vb{c}', t) + \alpha (\epsilon_{\theta}(\vb{z_{t}}, \vb{c}', t) - \epsilon_{\theta}(\vb{z_{t}}, t)), \hspace{1em}  \forall t\in [T, 1]
\end{equation}
where $\vb{c}' = v_{\gamma}^{restored}[a_{i}](c)$ for a given caption $c$. This score estimate $\hat{\epsilon}_{\theta}(\vb{z_{t}}, \vb{c}', t)$ at each time-step $t$ is used to obtain the final latent code $\vb{z}_{0}$ which is then used with the VQ-VAE decoder to obtain the final image $\vb{x}_{0}^{restored}$.
\vspace{-0.3cm}
\subsection{Extracting Causal States Using CLIP-Score}
\label{verify}
In this section, we discuss details on how to retrieve causal states using automated metrics such as \texttt{CLIP-Score}~\citep{clip_s_reference}. Let $\vb{x}^{restored}_{0}(a_{i})$ be the final image generated by the diffusion model after intervening on layer $a_{i}$, $\vb{x}_{0}$ be the image generated by the clean diffusion model and $\vb{x}^{corr}$ be the final image generated by the corrupted model. In particular, we are interested in the average indirect effect~\citep{vig_nlp, pearl2013direct} which measures the difference between the corrupted model and the restored model. Intuitively, a higher value of average indirect effect (AIE) signifies that the restored model deviates from the corrupted model. To compute the average indirect effect with respect to causal mediation analysis for text-to-image models such as Stable-Diffusion, we use \texttt{CLIP-Score} which computes the similarity between an image embedding and a caption embedding. In particular, AIE = $|\texttt{CLIPScore}(\vb{x}_{0}^{restored}, c) - \texttt{CLIPScore}(\vb{x}_{0}^{corr}, c)|$. Given $\vb{x}_{0}^{corr}$ is common across all the layers for a caption, we can use $\texttt{CLIPScore}(\vb{x}_{0}^{restored}, c)$ as the AIE.

\textbf{Selecting Threshold for \texttt{CLIP-Score}.} In order to determine the optimal threshold value for \texttt{CLIP-Score}, we select a small validation set of 10 prompts per attribute. To this end, we establish a concise user study interface (refer to~\Cref{validation_design} for details). 
Through human participation, we collect binary ratings if an image generated by restoring a particular layer is faithful to the original captions. We then extract the common causal states across all the prompts for a given attribute and find the average (across all the prompts) \texttt{CLIP-Score} for each causal state. We then use the lowest average \texttt{CLIP-Score} corresponding to a causal state as the threshold, which we apply on the probe dataset in~\Cref{probe_dataset} to filter the causal states at scale for each attribute separately. 
\vspace{-0.4cm}
\section{How is Knowledge Stored in Text-to-Image Models? }
\vspace{-0.4cm}
\label{automated_metrics}
In this section, we discuss the results of tracing knowledge across various components of the text-to-image model in details. 

\textbf{Tracing Results for UNet.} 
In~\cref{tracing_1}, we illustrate the distribution of causal states across different visual attributes within the UNet architecture using the \texttt{CLIP-Score} metric. This metric evaluates the faithfulness of the image produced by the restored state $\mathbf{x}_{0}^{restored}$ compared to the original caption $c$. 
From the insights derived in ~\cref{tracing_1}, it becomes evident that causal states are spread across diverse components of the UNet. In particular, we find that the density of the causal states are more in the \texttt{up-block} of the UNet when compared to the \texttt{down-block} or the \texttt{mid-block}. 
Nonetheless, a notable distinction emerges in this distribution across distinct attributes. For instance, when examining the {\it style} attribute, the initial self-attention layer demonstrates causality, whereas this causal relationship is absent for other attributes. Similarly, in the context of the {\it action} attribute, the cross-attention layer within the mid-block exhibits causality, which contrasts with its non-causal behavior concerning other visual attributes. ~\cref{tracing_1} showcases the images generated by restoring both causal and non-causal layers within the UNet. A comprehensive qualitative enumeration of both causal and non-causal layers for each visual attribute is provided in~\Cref{qual_unet}. 
Our findings underscore the presence of information pertaining to various visual attributes in regions beyond the cross-attention layers. Importantly, we observe that the distribution of information within the UNet diverges from the patterns identified in extensive generative language models, as noted in prior research~\citep{meng2023locating}, where attribute-related knowledge is confined to a few proximate layers. In~\Cref{fine_grained_interpret}, we provide additional causal tracing results, where we add Gaussian noise to the entire text-embedding. Even in such a case, certain causal states can restore the model close to its original configuration, highlighting that the conditional information can be completely bypassed if certain causal states are active.

\textbf{Tracing Results for Text-Encoder.} In~\Cref{text_tracing}, we illustrate the causal states in the text-encoder for Stable-Diffusion corresponding to various visual attributes. At the text-encoder level, we find that the causal states are localized to the first self-attention layer corresponding to the last subject token across all the attributes. In fact, there exists {\it only one} causal state in the text-encoder. Qualitative visualizations in~\Cref{text_tracing} and~\Cref{qual_text_encoder} illustrate that the restoration of layers other than the first self-attention layer corresponding to the subject token does not lead to images with high fidelity to the original caption. Remarkably, this observation is distinct from generative language models where factual knowledge is primarily localized in the proximate mid MLP layers~\cite{meng2023locating}. 
\begin{tcolorbox}
\vspace{-0.2cm}
\textbf{General Takeaway.} Causal components corresponding to various visual attributes are dispersed (with a {\it different distribution} between {\it distinct} attributes) in the UNet, whereas there exists {\it only} one causal component in the text-encoder. 
\vspace{-0.2cm}
\end{tcolorbox}

The text-encoder's strong localization of causal states for visual attributes enables controlled knowledge manipulation in text-to-image models, facilitating updates or removal of concepts. However, since attribute knowledge is dispersed in the UNet, targeted editing is challenging without layer interference. While fine-tuning methods for UNet model editing exist~\citep{gandikota2023erasing, kumari2023ablating}, they lack scalability and don't support simultaneous editing of multiple concepts. In the next section, we introduce a closed-form editing method, \difffix{}, leveraging our causal tracing insights to efficiently edit various concepts in text-to-image models.
\vspace{-0.3cm}
\section{\difffix{}: Fast Model Editing for Text-to-Image Models}
\vspace{-0.3cm}
\label{diff_fix}
\subsection{Editing Method}
Recent works such as~\citep{kumari2023ablating, gandikota2023erasing} edit concepts from text-to-image diffusion models by fine-tuning the UNet. They generate training data for fine-tuning using the pre-trained diffusion model itself. While both methods are effective at editing concepts, fine-tuning the UNet can be expensive due to backpropogation of gradients through the UNet. To circumvent this issue, we design a fast, data-free model editing method leveraging our interpretability observations in~\Cref{automated_metrics}, where we find that there exists only one causal state (the very first self-attention layer) in the text-encoder for Stable-Diffusion. 

Our editing method~\difffix{} can update text-to-image diffusion models in a targeted way in under $1s$ through a closed-form update making it 1000x faster than existing fine-tuning based concept ablating methods such as~\citep{kumari2023ablating, gandikota2023erasing}.  The first self-attention layer in the text-encoder for Stable-Diffusion contains four updatable weight matrices: $W_{k}, W_{q}, W_{v}$ and $W_{out}$, where $W_{k}, W_{q}, W_{v}$ are the projection matrices for the key, query and value embeddings respectively. $W_{out}$ is the projection matrix before the output from the \texttt{self-attn-0} layer after the attention operations. \difffix{} specifically updates this $W_{out}$ matrix by collecting caption pairs $(c_{k}, c_{v})$ where $c_{k}$ (key) is the original caption and $c_{v}$ (value) is the caption to which $c_{k}$ is mapped. For e.g., to remove the style of {\it `Van Gogh'}, we set $c_{k} = \text{`{\it Van Gogh}'}$ and $c_{v} = \text{`{\it Painting}'}$.  In particular, to update $W_{out}$, we solve the following optimization problem:
\begin{equation}
    \label{opt_edit}
    \min_{W_{out}} \sum_{i=1}^{N} \|W_{out}k_{i} - v_{i} \|_{2}^{2} + \lambda \|W_{out} - W_{out}' \|_{2}^{2},
\end{equation}
where $\lambda$ is a regularizer to not deviate significantly from the original pre-trained weights $W_{out}'$, $N$ denotes the total number of caption pairs containing the last subject token embeddings of the key and value. $k_{i}$ corresponds to the embedding of $c_{k_{i}}$ after the attention operation using $W_{q}, W_{k}$ and $W_{v}$ for the $i^{th}$ caption pair. $v_{i}$ corresponds to the embedding of $c_{v_{i}}$ after the original pre-trained weights $W_{out}^{'}$ acts on it. 

One can observe that Eq. (\ref{opt_edit}) has a closed-form solution due to the absence of any non-linearities. In particular, the optimal $W_{out}$ can be expressed as the following: 
\begin{equation}
    W_{out} = (\lambda W_{out}' + \sum_{i=1}^{N} v_{i}k_{i}^{T})(\lambda I + \sum_{i=1}^{N}k_{i}k_{i}^{T})^{-1},
\end{equation}
In \Cref{exp_results}, we show qualitative as well as quantitative results using \difffix{} for editing various concepts in text-to-image models. 
\vspace{-0.2cm}
\subsection{Experimental Setup}
\label{exp_setup}
We validate ~\difffix{} by applying edits to a Stable-Diffusion~\citep{sd_main} model and quantifying the \emph{efficacy} of the edit. For removing concepts such as artistic styles or objects using~\difffix{}, we use the prompt dataset from~\citep{kumari2023ablating}. For updating knowledge (e.g., {\it President of a country}) in text-to-image models, we add newer prompts to the prompt dataset from~\citep{kumari2023ablating} and provide further details in~\Cref{editing _dataset}.
\begin{figure*}
    \hskip -0.4cm
  \includegraphics[width=14.7cm, height=3.9cm]{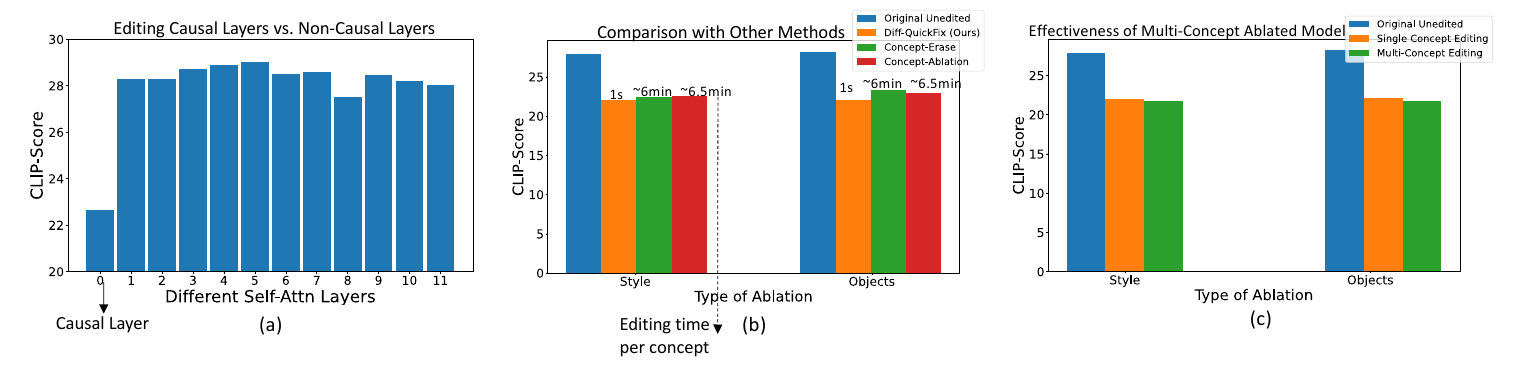}
  \vspace{-1cm}
\caption{\label{edit_composite_clip} \textbf{Quantitative Analysis of \difffix{}}. (a) Editing Causal vs. Non-Causal Layers (Averaged across {\it Objects, Style and Facts}): Lower \texttt{CLIP-Score} for causal layer indicates successful edits; (b) Efficacy of \difffix{} when compared to other methods -- Our method leads to comparable \texttt{CLIP-Scores} to fine-tuning based approaches, but can edit concepts 1000x faster; (c) \difffix{} can be used to effectively edit multiple concepts at once, shown by comparable \texttt{CLIP-Scores} to the single-concept edited ones.}%
    \vspace{-0.6cm}
\end{figure*}
We compare our method with (i) Original Stable-Diffusion; (ii) Editing methods from~\citep{kumari2023ablating} and~\citep{gandikota2023erasing}. To validate the effectiveness of editing methods including our~\difffix{}, we perform evaluation using automated metrics such as \texttt{CLIP-Score}. In particular, we compute the \texttt{CLIP-Score} between the images from the edited model and the concept corresponding to the visual attribute which is edited. A low \texttt{CLIP-Score} therefore indicates correct edits.
\vspace{-0.3cm}
\subsection{Editing Results}
\label{exp_results}
\vspace{-0.2cm}
\textbf{Editing Non-causal Layers Does Not Lead to Correct Edits. } We use~\difffix{} with the non-causal self-attention layers in the text-encoder to ablate {\it styles}, {\it objects} and update {\it facts}. In~\Cref{edit_composite_clip}-(a), we compute the \texttt{CLIP-Score} between the generated images and the attribute from the original captions (e.g., {\it van gogh} in the case of {\it style}). 
In particular, we find that editing the non-causal layers does not lead to any intended model changes -- highlighted by the high \texttt{CLIP-Scores} consistently across non-causal layers (layers numbered 1 to 11). However, editing the sole causal layer (layer-0) leads to correct model changes, highlighted by the lower \texttt{CLIP-Score} between the generated images from the edited model and the attribute from the original captions. This shows that identifying the causal states in the model is particularly important to perform targeted model editing for ablating concepts. In ~\Cref{qual_editing_non_causal}, we show additional qualitative visualizations highlighting that editing the non-causal states lead to similar model outputs as the unedited model. 
\begin{figure*}
    \hskip 0.0cm
    \vspace{-0.4cm}
  \includegraphics[width=14.0cm, height=4.3cm]{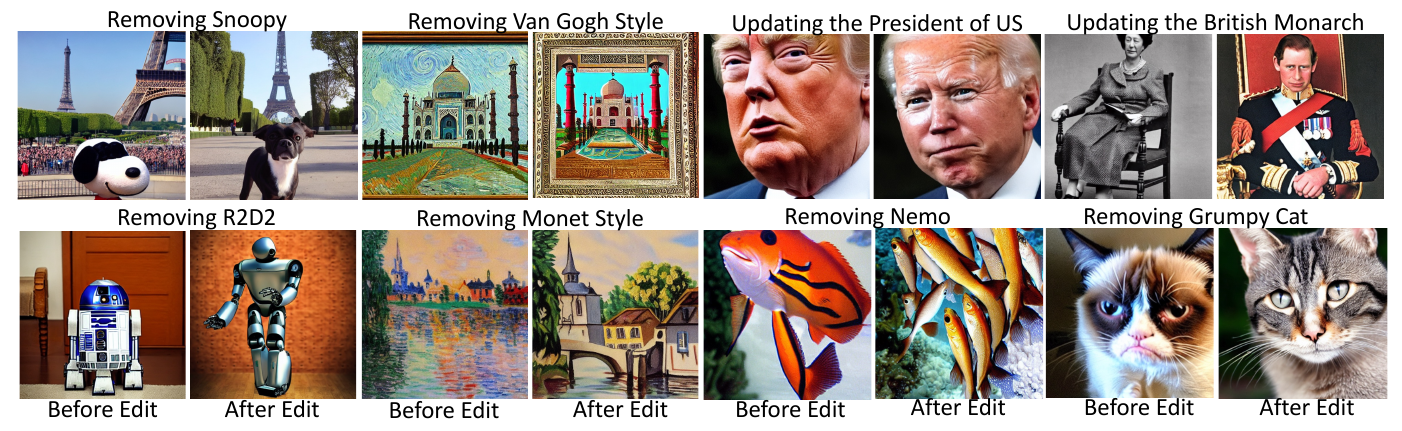}
  \vspace{-0.4cm}
    \caption{\label{edit_viz} \textbf{Qualitative Examples with using \difffix{}} to ablate {\it style}, {\it objects} and update {\it facts} in text-to-image models. More qualitative examples in the~\Cref{qual_editing_concepts}. }%
    \vspace{-0.2cm}
\end{figure*}

\textbf{Efficacy in Removing Styles and Objects}. ~\Cref{edit_composite_clip}-(b) shows the average \texttt{CLIP-Score} of the generated images from the edited model computed with the relevant attributes from the original captions. We find that the \texttt{CLIP-Score}  from the edited model with~\difffix{} decreases when compared to the generations from the unedited model. We also find that our editing method has comparable \texttt{CLIP-Scores} to other fine-tuning based approaches such as Concept-Erase~\citep{gandikota2023erasing} and Concept-Ablation~\citep{kumari2023ablating}, which are more computationally expensive.
~\Cref{edit_viz} shows qualitative visualizations corresponding to images generated by the text-to-image model before and after the edit operations.
Together, these quantitative and qualitative results show that~\difffix{} is able to effectively remove various {\it styles} and {\it objects} from an underlying text-to-image model. In~\Cref{qual_editing_concepts} we provide additional qualitative visualizations and in~\Cref{surrounding_concepts} we show additional results showing that our editing method does not harm surrounding concepts (For e.g., removing the style of {\it Van Gogh} does not harm the style of {\it Monet}).


\textbf{Efficacy in Updating Stale Knowledge}. The \texttt{CLIP-Score} between the generated images and a caption designating the incorrect fact (e.g., {\it Donald Trump} as the  {\it President of the US}) decreases from 0.28 to 0.23 after editing with~\difffix{}, while the \texttt{CLIP-Score} with the correct fact (e.g., {\it Joe Biden} as the {\it President of the US}) increases from 0.22 to 0.29 after the relevant edit. This shows that the incorrect fact is updated with the correct fact in the text-to-image model. 
Additional qualitative visualizations are provided in ~\Cref{edit_viz} and~\Cref{qual_editing_concepts}. 

 
\textbf{Multiple Edits using~\difffix{}}.
An important feature of \difffix{} is its capability to ablate multiple concepts simultaneously. In~\Cref{edit_composite_clip}-(c), our framework demonstrates the removal of up to 10 distinct styles and objects at once. This multi-concept ablation results in lower \texttt{CLIP-Scores} compared to the original model, similar \texttt{CLIP-Scores} to single concept editing. This scalability suggests our framework's potential for large-scale multi-concept editing. In ~\Cref{multi_concept_ablated}, we provide qualitative visualizations of generations from the multi-concept ablated model, showcasing the effectiveness of our editing method in removing multiple concepts. Additionally, we highlight \difffix{}'s efficiency in eliminating a larger number of artistic styles, successfully removing 50 top artistic styles from Stable-Diffusion.

\vspace{-0.45cm}
\section{Conclusion}
Through the lens of Causal Mediation Analysis, we present methods for understanding the storage of knowledge corresponding to diverse visual attributes in text-to-image diffusion models. Notably, we find a distinct distribution of causal states across visual attributes in the UNet, while the text-encoder maintains a single causal state. This differs significantly from observations in language models like GPT, where factual information is concentrated in mid-MLP layers. In contrast, our analysis shows that public text-to-image models like Stable-Diffusion concentrate multiple visual attributes within the first self-attention layer of the text-encoder. Harnessing the insights from these observations, we design a fast model editing method ~\difffix{}. This approach outpaces existing editing methods by a factor of 1000, successfully ablating concepts from text-to-image models. The potency of ~\difffix{} is manifested through its adeptness in removing artistic styles, objects, and updating outdated knowledge all accomplished data-free and in less than a second, making ~\difffix{} a practical asset for real-world model editing scenarios.
\section{Acknowledgements}
This work was started and majorly done during Samyadeep's internship at Adobe Research. At UMD, Samyadeep Basu and Soheil Feizi are supported in part by a grant from an NSF CAREER AWARD 1942230, ONR YIP award N00014-22-1-2271, ARO’s Early Career Program Award 310902-00001, Meta grant 23010098, HR00112090132 (DARPA/RED), HR001119S0026 (DARPA/GARD), Army Grant No. W911NF2120076, NIST 60NANB20D134, the NSF award CCF2212458, an Amazon Research Award and an award from Capital One. The authors would like to thank Ryan Rossi for proofreading the draft.
\bibliography{iclr2024_conference}
\bibliographystyle{iclr2024_conference}

\newpage 
\appendix
\startcontents[mainsections]
\printcontents[mainsections]{l}{1}{\section*{Appendix Sections}\setcounter{tocdepth}{2}}

\section{Probe Dataset Design Details}
\label{probe_dataset}
In this section, we provide detailed descriptions of the probe dataset $\mathcal{P}$ which is used for causal tracing for both the UNet and the text-encoder. We primarily focus on four visual attributes : {\it style, color, objects} and {\it action}. In addition, we also use the causal tracing framework adapted for text-to-image diffusion models to analyse the {\it viewpoint} and {\it count} attribute. The main reason for focusing on {\it style, color, objects} and {\it action} is the fact that generations from current text-to-image models have a strong fidelity to these attributes, whereas the generations corresponding to {\it viewpoint} or {\it count} are often error-prone.  We generate probe captions for each of the attribute in the following way:

\begin{itemize}
    \item {\it \textbf{Objects}.} We select a list of 24 objects and 7 locations (e.g., {\it beach, forest, city, kitchen, mountain, park, room}) to create a set of 168 unique captions. The objects are : {\it \{ `dog', `cat', `bicycle', `oven', `tv', `bowl', `banana', `bottle', `cup', `fork', `knife', `apple', `sandwich', `pizza', `bed', `tv', `laptop', `microwave', `book', `clock', `vase', `toothbrush', `donut', `handbag' \}} . We then use the template: {\it `A photo of <object> in <location>.'} to generate multiple captions to probe the text-to-image model. These objects are selected from the list of objects present in MS-COCO. 
    \item {\it \textbf{Style}.} We select the list of 80 unique objects from MS-COCO and combine it with an artistic style from : {\it \{monet, pablo picasso, salvador dali, van gogh, baroque, leonardo da vinci, michelangelo\} }. The template used is: {\it `A <object> in the style of <artistic-style>'}. In total, using this template we generate 560 captions. 
    \item {\it \textbf{Color}.} We select the list of 80 unique objects from MS-COCO and combine it with a color from {\it \{ blue, red, yellow, brown, black, pink, green\}}. We then use the template: {\it `A <color> <object>'} to generate 560 unique captions to probe the text-to-image model. 
    \item {\it \textbf{Action}.} We first choose certain actions such as {\it eating, running, sitting, sprinting} and ask ChatGPT\footnote{We use version 3.5 for ChatGPT.} to list a set of animals who can perform these actions. From this list, we choose a set of 14 animals: {\it \{ bear, cat, deer, dog, giraffe, goat, horse, lion, monkey, mouse, ostrich, sheep, tiger, wolf\}}. In total we use the template: {\it `An <animal><action>'} to create a set of 56 unique captions for probing. 
    \item {\it \textbf{Viewpoint}. } For viewpoint, we use the same set of 24 objects from the {\it Objects} attribute and combine it with a viewpoint selected from {\it \{front, back, top, bottom\}} to create 96 captions in the template of: {\it `A <object> from the <viewpoint>'}. 
    \item {\it \textbf{Count}.} We use the same 24 objects from {\it Objects} attribute and introduce a count before the object in the caption. We specifically use the following template to generate captions: {\it `A photo of <count> objects in a room.'}, where we keep the location fixed. We select a count from \{2,4,6,8\} to create 96 unique captions in total. 
\end{itemize}

\begin{table}[ht]
  \centering
  \resizebox{\columnwidth}{!}{\begin{tabular}{lllllllll}
    \toprule
     & \multicolumn{5}{c} {Description of Probe Dataset for Causal Tracing} & & &\\
    \cmidrule(r){2-6}
    Attribute     & Description  & Example 1 & Example 2  & Example 3 \\  
    \midrule
    Objects   & Prompt containing an object in a location & photo of a {\it vase} in a room & a photo of a {\it car} in a desert  & a photo of a {\it house} in a forest \\
    Style  & An object drawn in a particular artistic style & airplane in the style of {\it van gogh} & town in the style of {\it monet} & bicycle in the style of {\it baroque} \\
    Color    & An object in a particular color & a {\it blue} car  & a {\it black} vase &  a {\it pink} bag \\
    Action   & An animal in a particular action & A giraffe {\it eating} & A tiger {\it running} & A cat {\it standing} \\
    Viewpoint     & An object in a particular viewpoint & A sofa from the {\it back} & A car from the {\it front} & A bus from the {\it side} \\
    Count  & Number of objects in a location & There are {\it 10} cars on the road & {\it 5 bags} in the room & {\it 6} laptops on a table  & & \\
    
    \bottomrule
  \end{tabular}}
 \caption{\textbf{Examples from the Probe Dataset Used For Causal Tracing.} The attributes in the captions are marked in {\it italics}.}
\end{table}
\section{Qualitative Visualizations for Causal Tracing (UNet)}
\label{qual_unet}
\subsection{Objects}
\begin{figure}[H]
    \hskip -2.6cm
  \includegraphics[width=18.5cm, height=19.3cm]{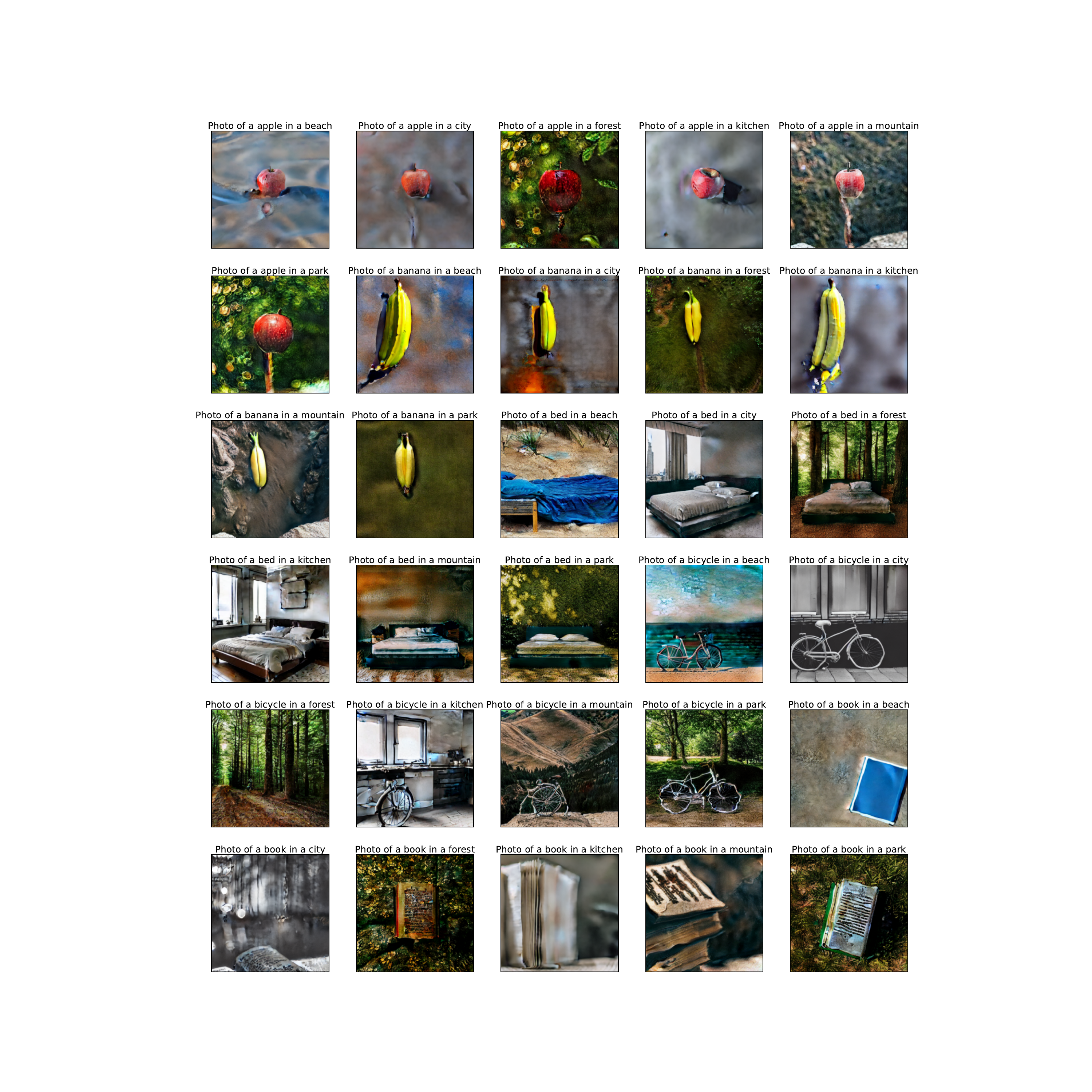}
  \vspace{-0.2cm}
    \caption{\label{causal_down1_resnet1} \textbf{Causal State: down-1-resnet-1.} We find that restoring the down-1-resnet-1 block in the UNet leads to generation of images with strong fidelity to the original caption. }%
\end{figure}
\begin{figure}[H]
    \hskip -2.6cm
  \includegraphics[width=18.5cm, height=19.3cm]{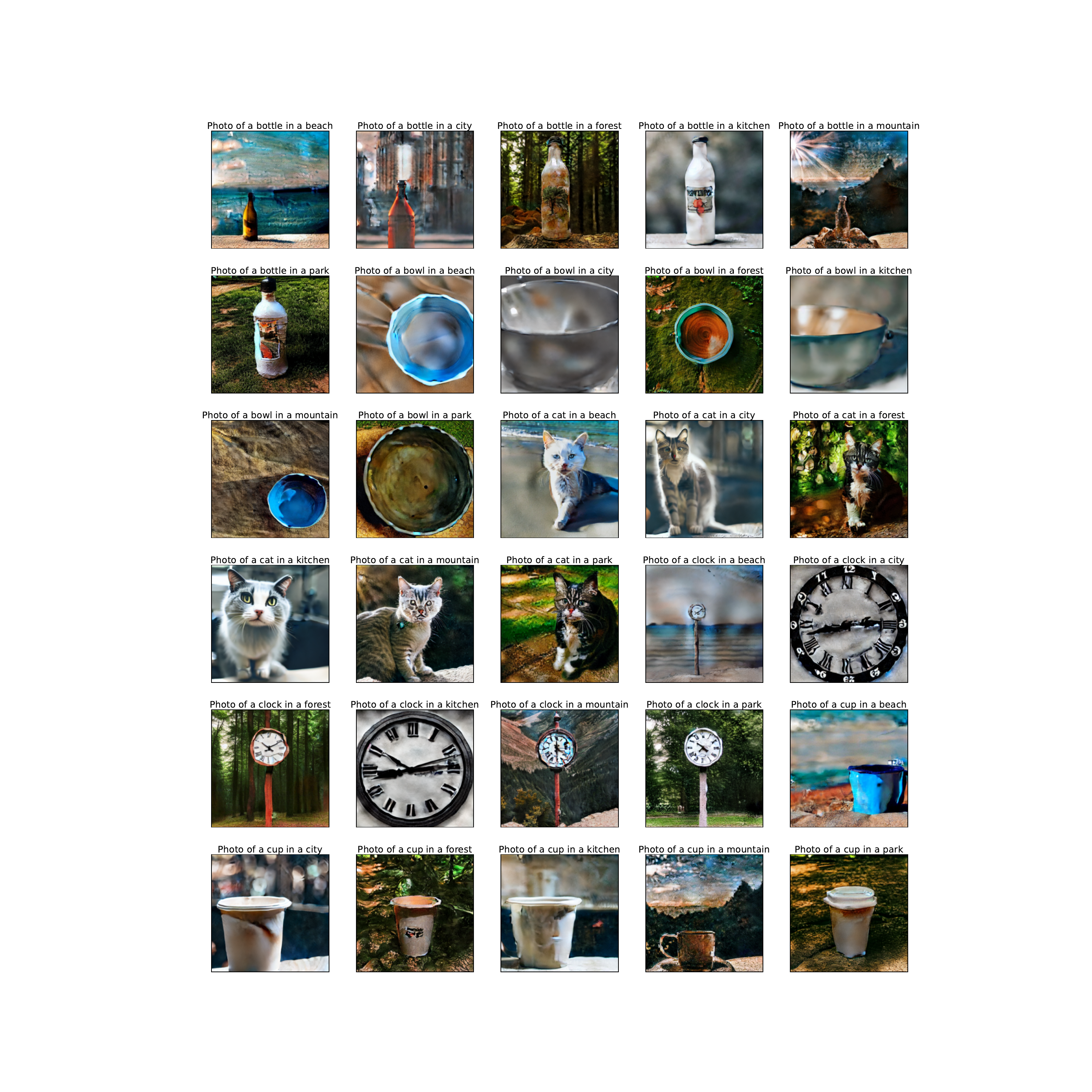}
  \vspace{-0.2cm}
    \caption{\label{causal_down1_resnet1_2} \textbf{Causal State: down-1-resnet-1.} We find that restoring the down-1-resnet-1 block in the UNet leads to generation of images with strong fidelity to the original caption. }%
\end{figure}
\begin{figure}[H]
    \hskip -2.6cm
  \includegraphics[width=18.5cm, height=19.3cm]{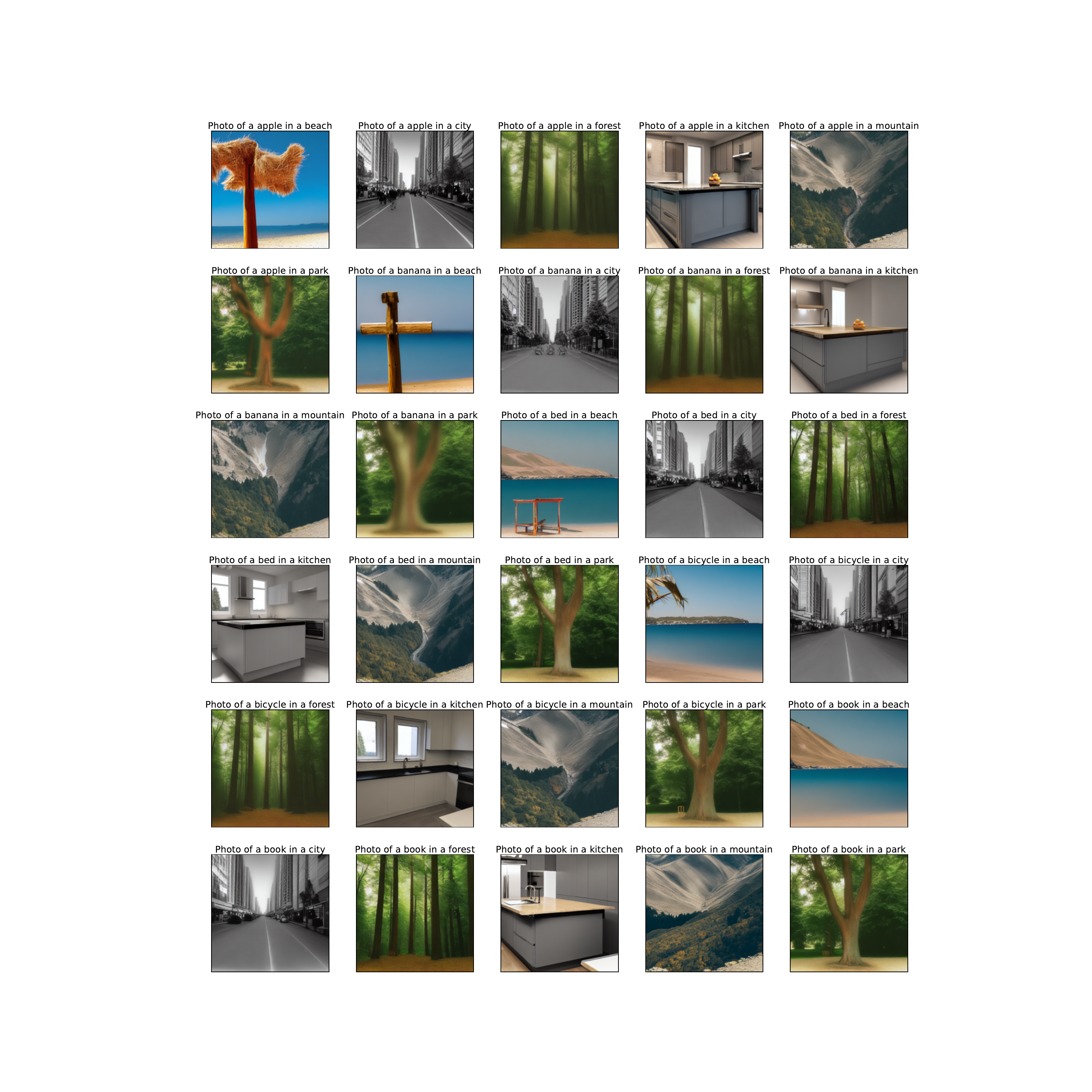}
  \vspace{-0.2cm}
    \caption{\label{non_causal_down1_resnet1_1} \textbf{Non-Causal State: down-blocks.0.attentions.0.transformer-blocks.0.attn2.} We find that restoring the down-blocks.0.attentions.0.transformer-blocks.0.attn2 block in the UNet leads to generation of images \textbf{without} the primary object, showing low fidelity to the original captions. }%
\end{figure}
\begin{figure}[H]
    \hskip -2.6cm
  \includegraphics[width=18.5cm, height=19.3cm]{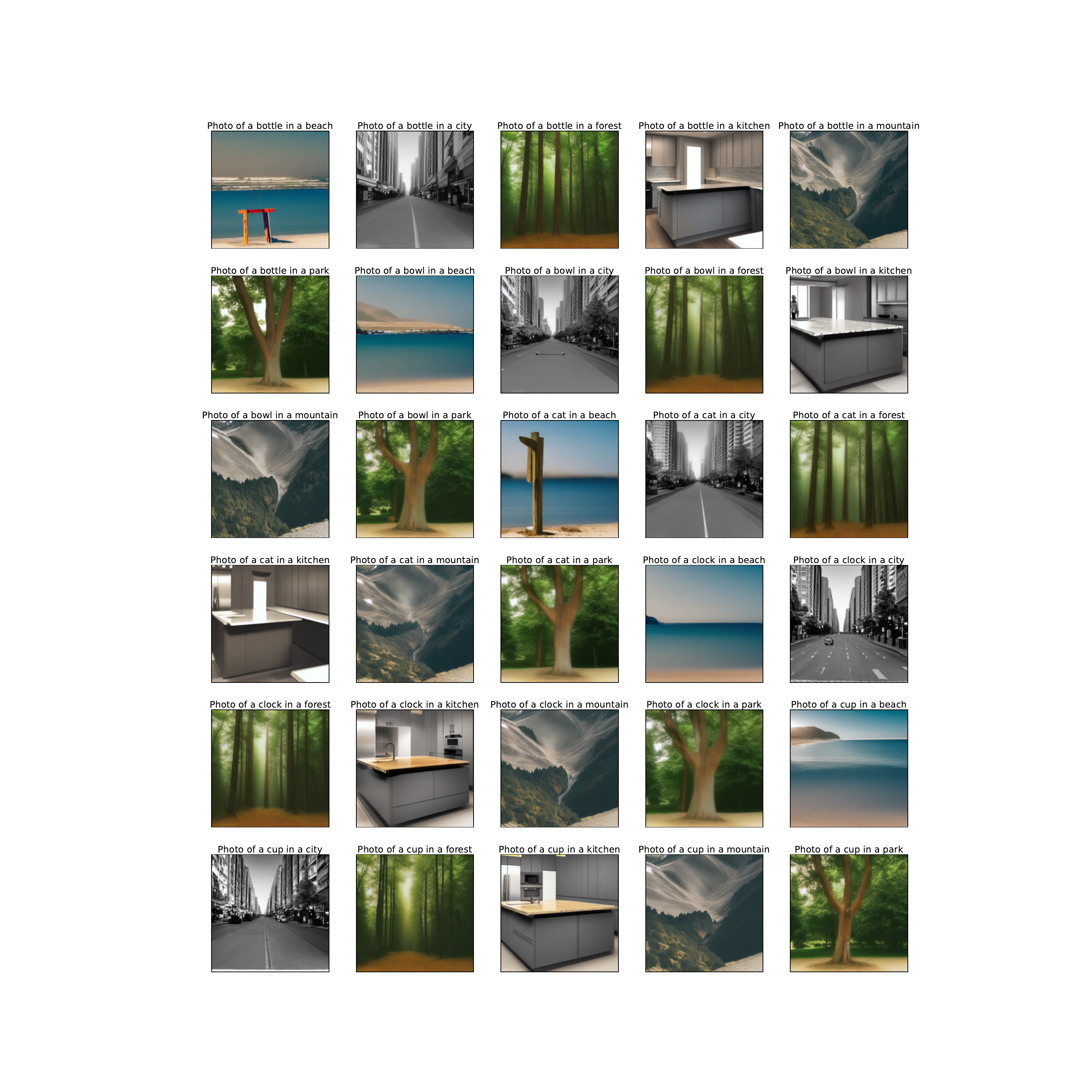}
  \vspace{-0.2cm}
    \caption{\label{non_causal_down1_resnet1_2} \textbf{Non-Causal State: down-blocks.0.attentions.0.transformer-blocks.0.attn2.} We find that restoring the down-blocks.0.attentions.0.transformer-blocks.0.attn2 block in the UNet leads to generation of images \textbf{without} the primary object, showing low fidelity to the original captions. }%
\end{figure}
\subsection{Action}
\begin{figure}[H]
    \hskip -2.6cm
  \includegraphics[width=18.5cm, height=19.3cm]{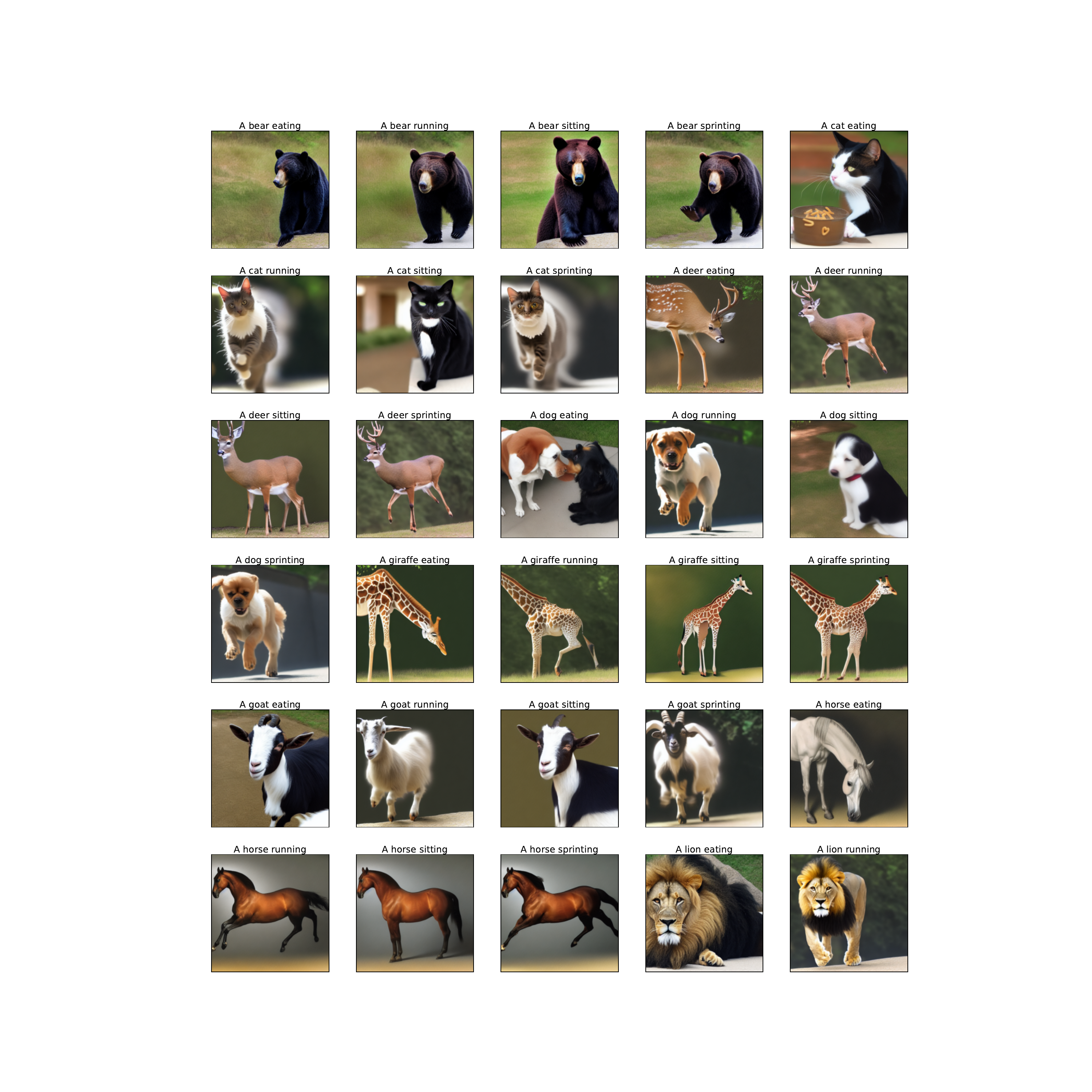}
  \vspace{-0.2cm}
    \caption{\label{causal_mid_cross_attn_action} \textbf{Causal State: mid-block-cross-attn.} We find that restoring the cross-attn in the mid-block in the UNet leads to generation of images with strong fidelity to the {\it action} attribute in the original caption. }%
\end{figure}
\begin{figure}[H]
    \hskip -2.6cm
  \includegraphics[width=18.5cm, height=19.3cm]{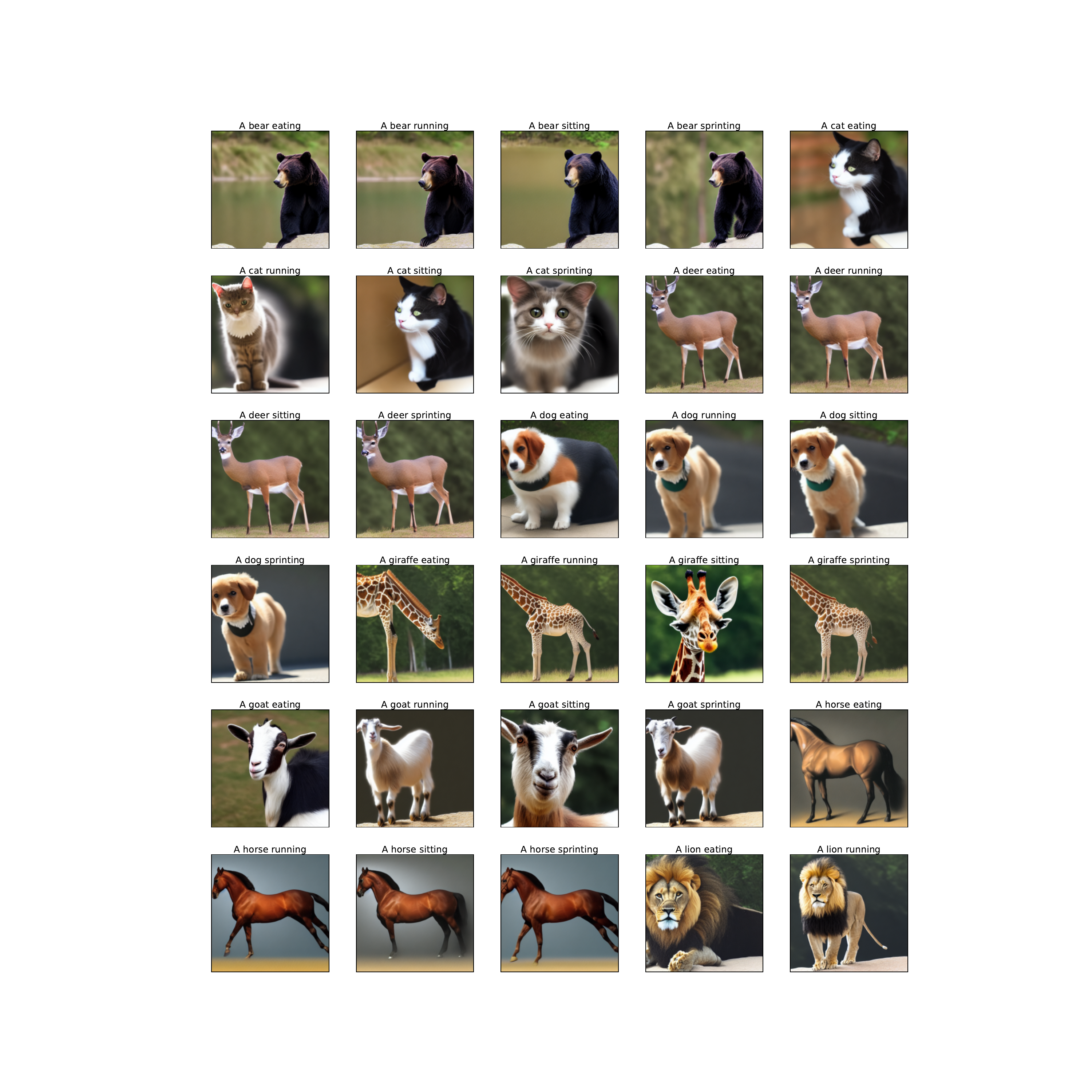}
  \vspace{-0.2cm}
    \caption{\label{non_causal_action} \textbf{Non-Causal State: down-blocks.2.attentions.1.transformer-blocks.0.attn2.} We find that restoring the down-blocks.2.attentions.1.transformer-blocks.0.attn2 in the UNet leads to generation of images with weak fidelity to the {\it action} attribute in the original caption. For a majority of the prompts, we find that the {\it action} attribute (especially those involving sprinting, running or eating) is not respected in the generated image. }%
\end{figure}
\subsection{Color}
\begin{figure}[H]
    \hskip -2.6cm
  \includegraphics[width=18.5cm, height=19.3cm]{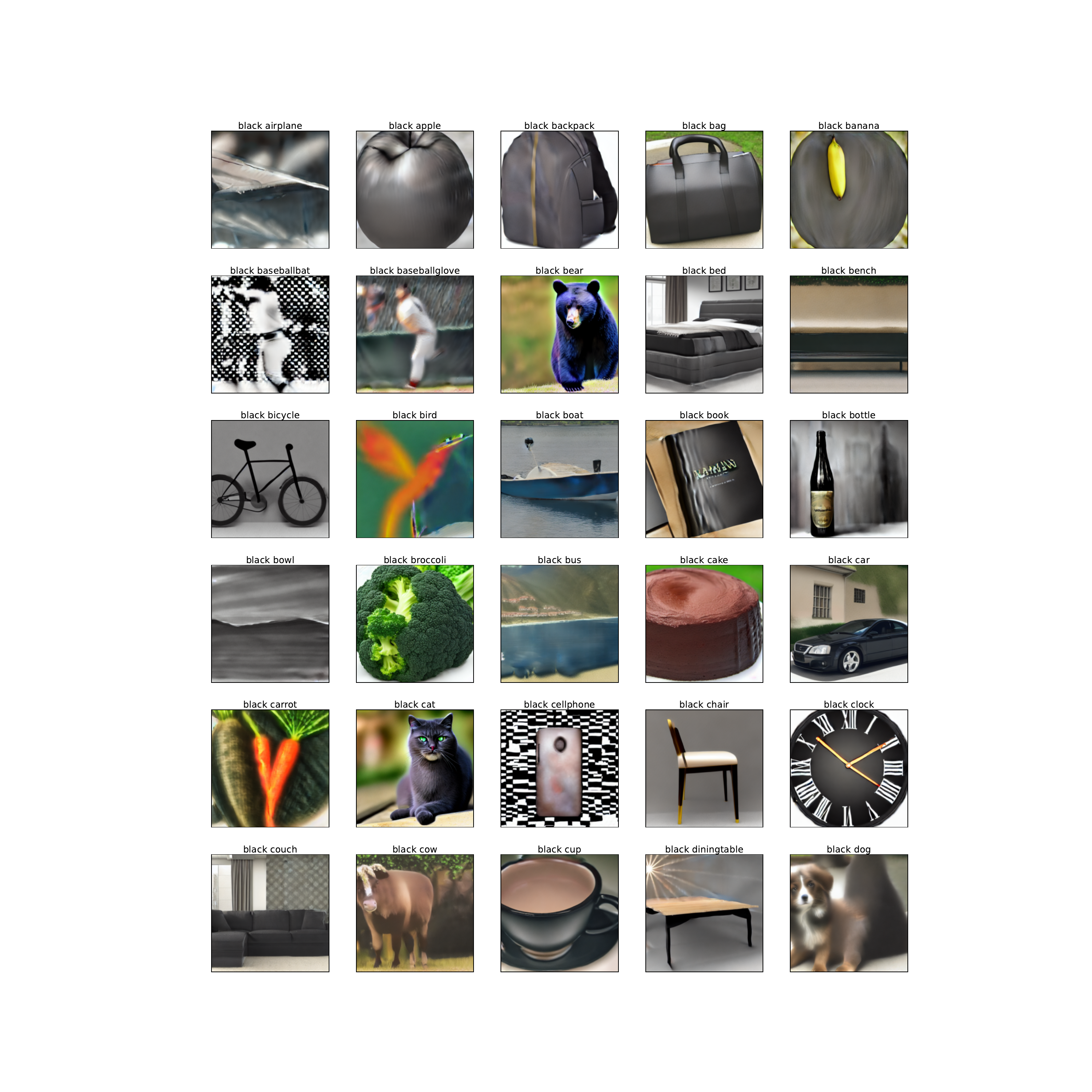}
  \vspace{-0.2cm}
    \caption{\label{causal_color_1} \textbf{Causal State: down-blocks.1.attentions.0.transformer-blocks.0.ff.} We find that restoring the down-blocks.1.attentions.0.transformer-blocks.0.ff in the down-block in the UNet leads to generation of images with strong fidelity to the {\it color} attribute in the original caption. }%
\end{figure}
\begin{figure}[H]
    \hskip -2.6cm
  \includegraphics[width=18.5cm, height=19.3cm]{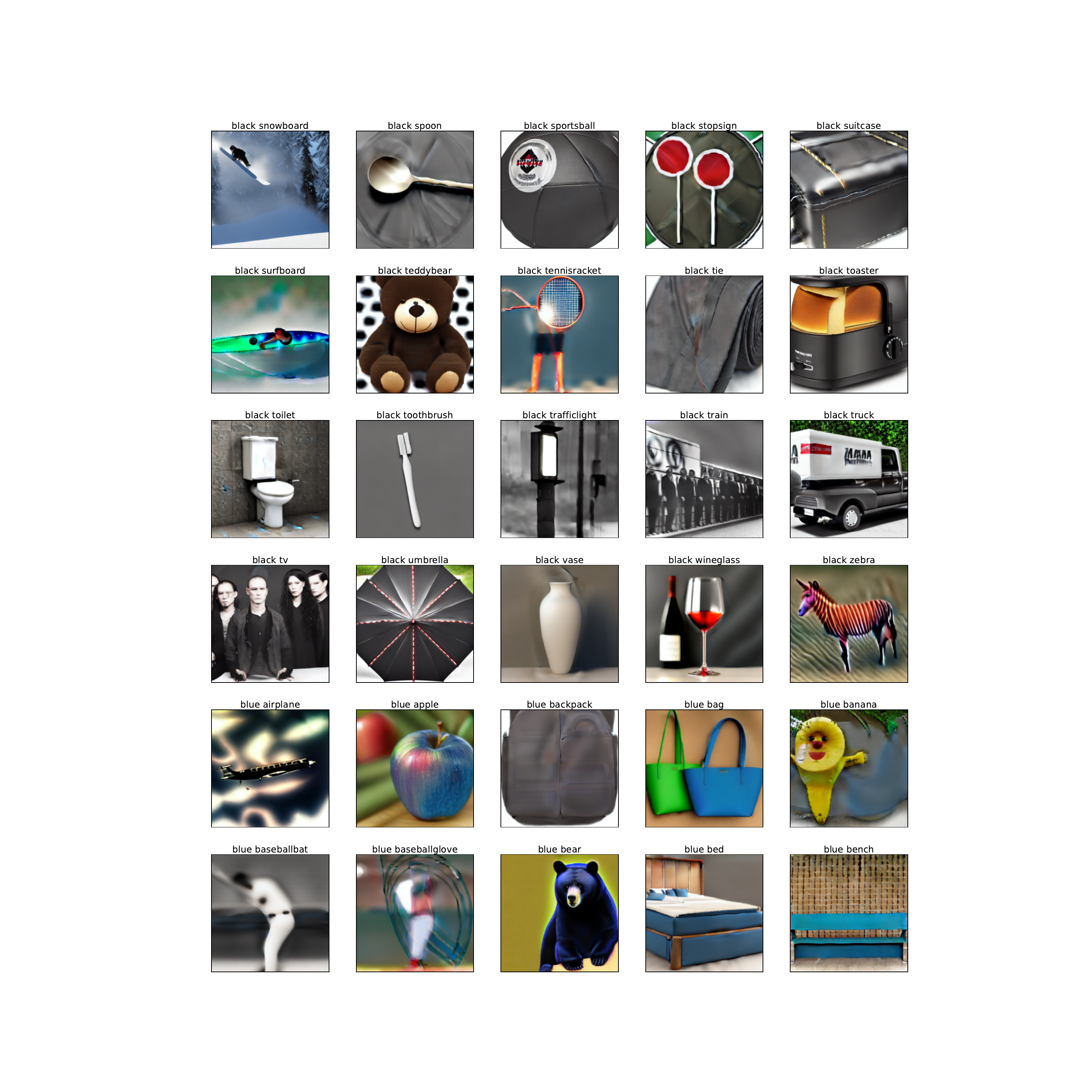}
  \vspace{-0.2cm}
    \caption{\label{causal_color_2} \textbf{Causal State: down-blocks.1.attentions.0.transformer-blocks.0.ff.} We find that restoring the down-blocks.1.attentions.0.transformer-blocks.0.ff in the down-block in the UNet leads to generation of images with strong fidelity to the {\it color} attribute in the original caption. }%
\end{figure}
\begin{figure}[H]
    \hskip -2.6cm
  \includegraphics[width=18.5cm, height=19.3cm]{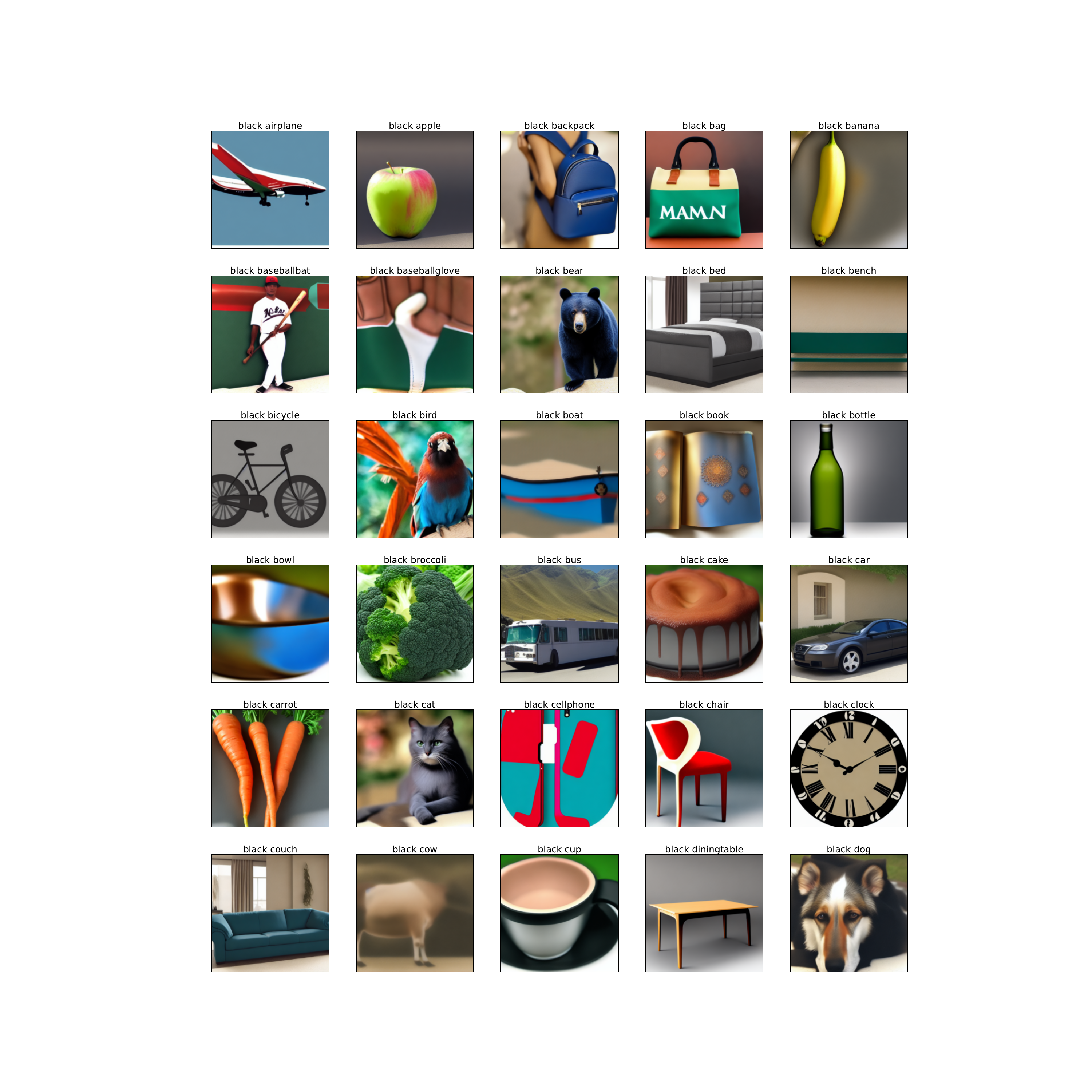}
  \vspace{-0.2cm}
    \caption{\label{non_causal_color_1} \textbf{Non-Causal State: mid-blocks.attentions.0.transformer-blocks.0.ff.} We find that restoring the mid-blocks.attentions.0.transformer-blocks.0.ff in the mid-block in the UNet does not lead to generation of images with strong fidelity to the {\it color} attribute in the original caption for a majority of cases. }%
\end{figure}
\begin{figure}[H]
    \hskip -2.6cm
  \includegraphics[width=18.5cm, height=19.3cm]{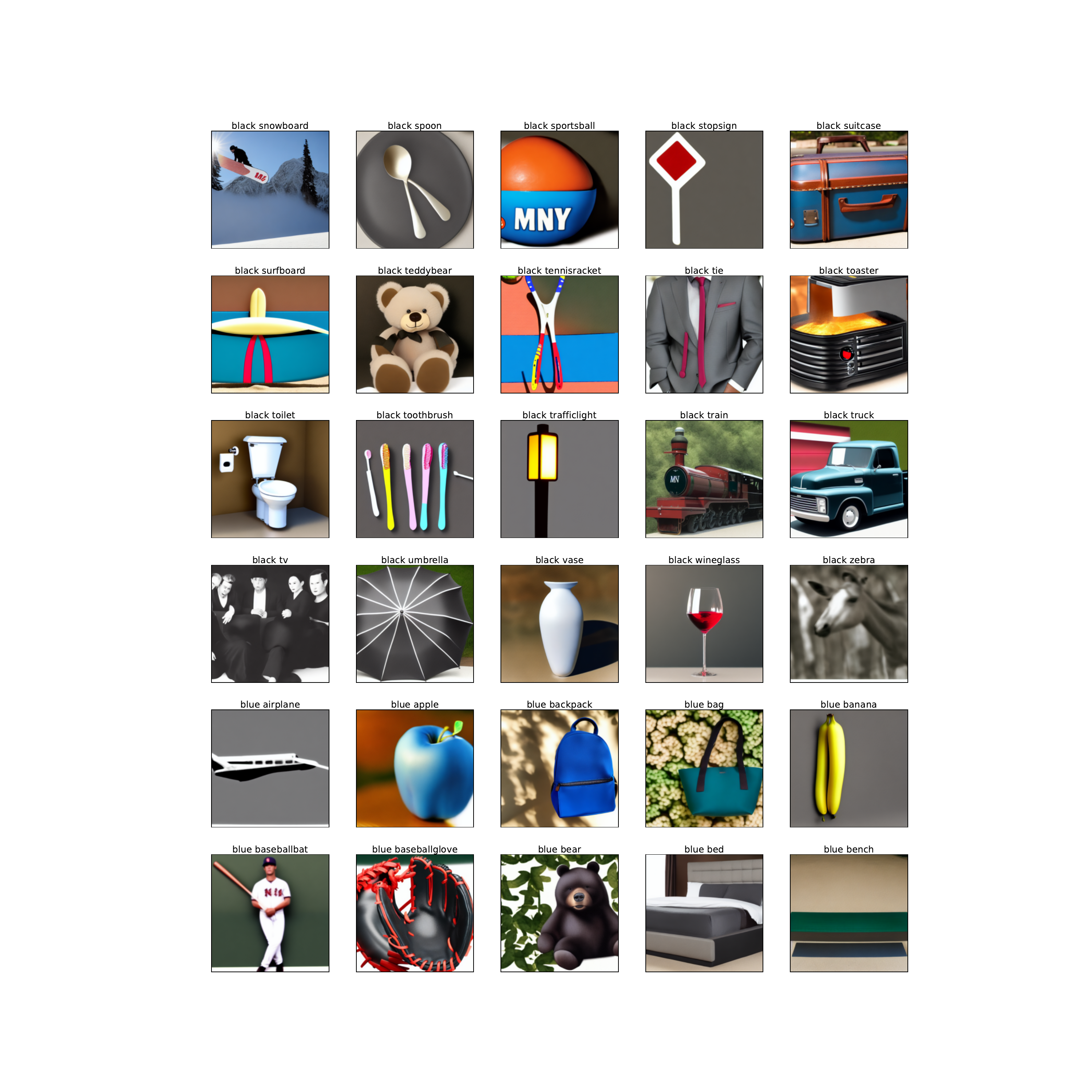}
  \vspace{-0.2cm}
    \caption{\label{non_causal_color_2} \textbf{Non-Causal State: mid-blocks.attentions.0.transformer-blocks.0.ff.} We find that restoring the down-mid-blocks.attentions.0.transformer-blocks.0.ff in the mid-block in the UNet does not lead to generation of images with strong fidelity to the {\it color} attribute in the original caption for a majority of cases. }%
\end{figure}
\subsection{Style}
\begin{figure}[H]
    \hskip -5.6cm
  \includegraphics[width=24.5cm, height=21.3cm]{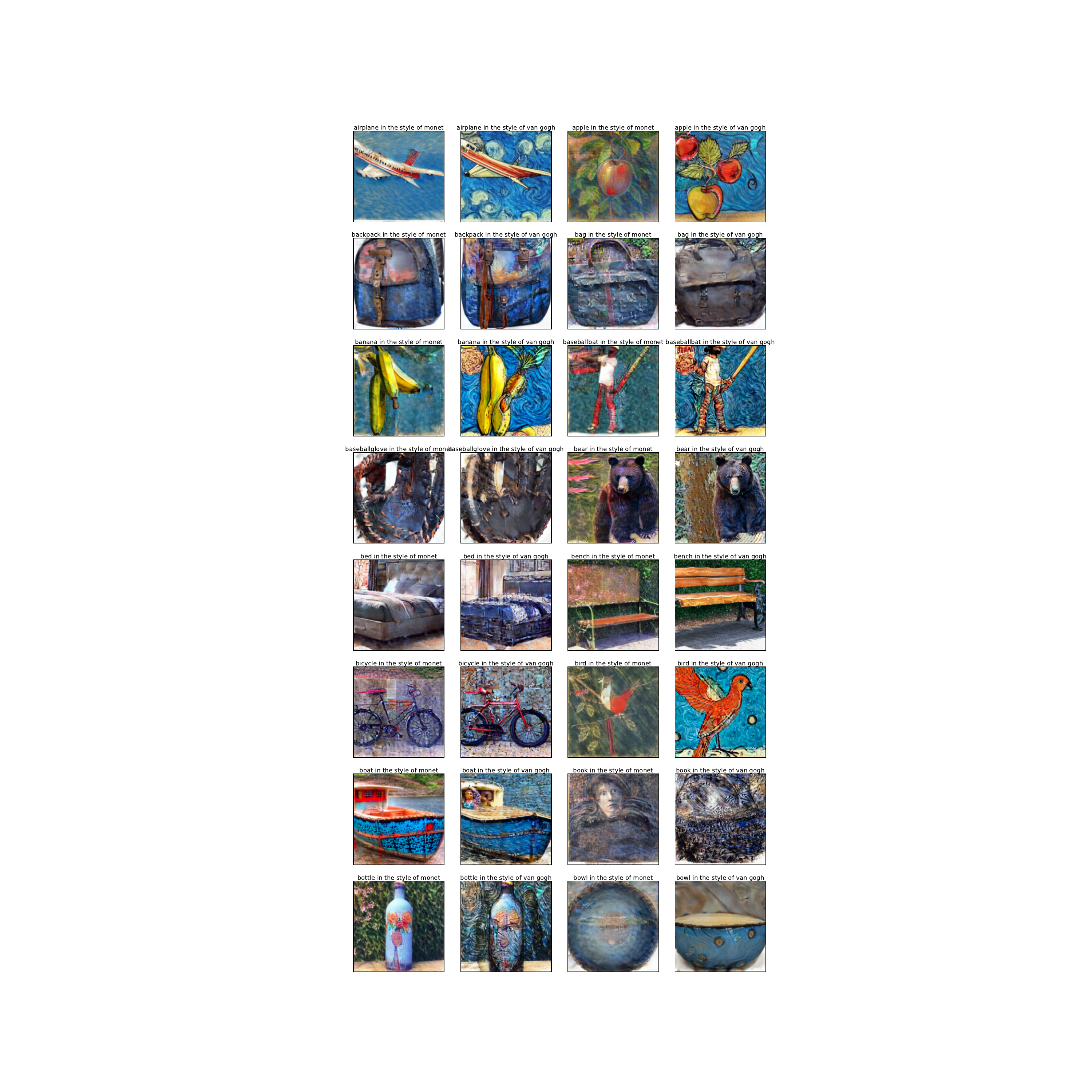}
  \vspace{-2.0cm}
    \caption{\label{causal_style_1} \textbf{Causal State: down-blocks.0.attn1 (First self-attn layer).} We find that restoring the down-blocks.self-attn.0 which is the first self-attention layer in the UNet \textbf{leads} to generation of images with strong fidelity to the {\it style} attribute in the original caption for a majority of cases. }%
\end{figure}
\begin{figure}[H]
    \hskip -5.6cm
  \includegraphics[width=24.5cm, height=21.3cm]{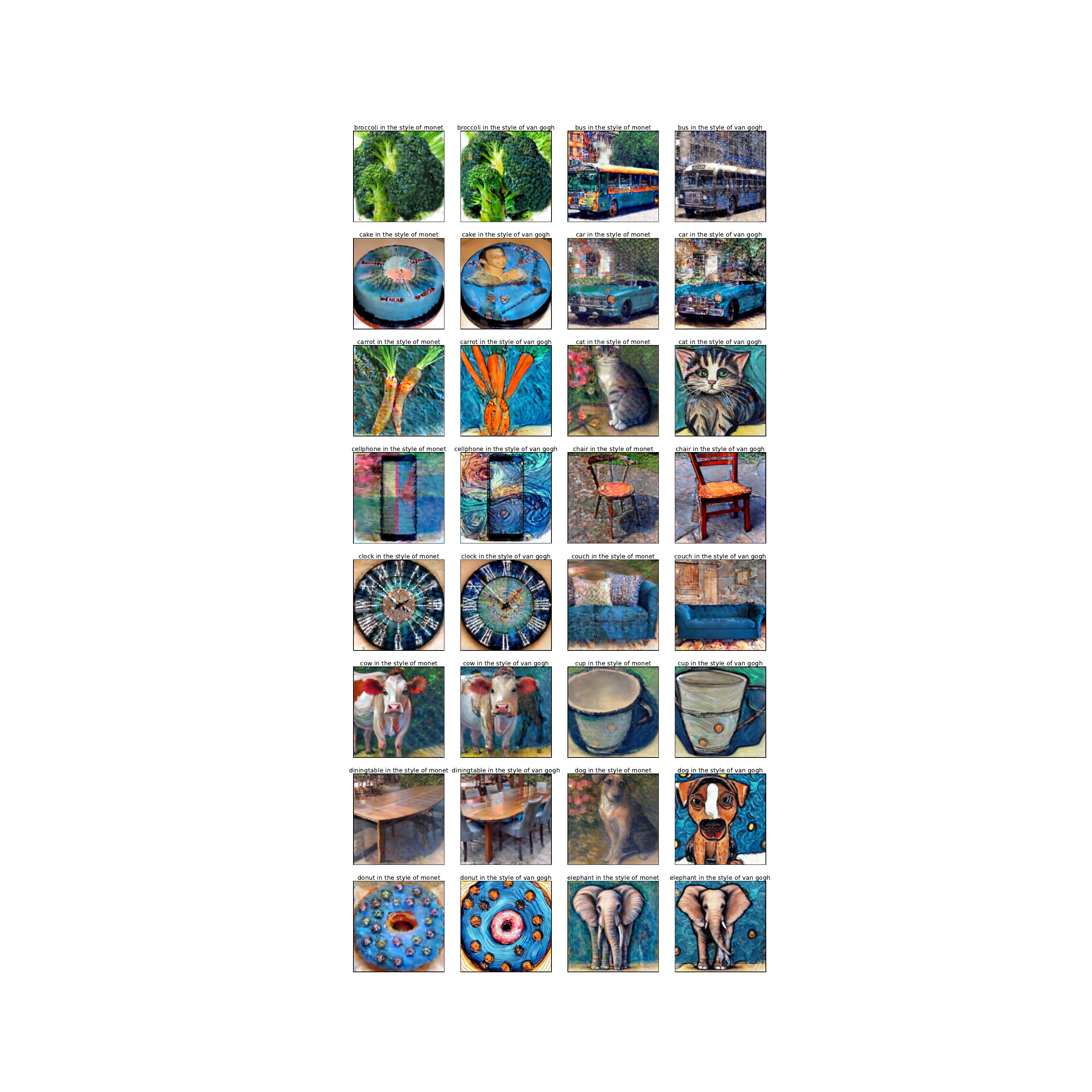}
  \vspace{-2.0cm}
    \caption{\label{causal_style_2} \textbf{Causal State: down-blocks.0.attn1.} We find that restoring the down-blocks.self-attn.0 which is the first self-attention layer in the UNet \textbf{leads} to generation of images with strong fidelity to the {\it style} attribute in the original caption for a majority of cases. }%
\end{figure}
\begin{figure}[H]
    \hskip -5.6cm
  \includegraphics[width=24.5cm, height=21.3cm]{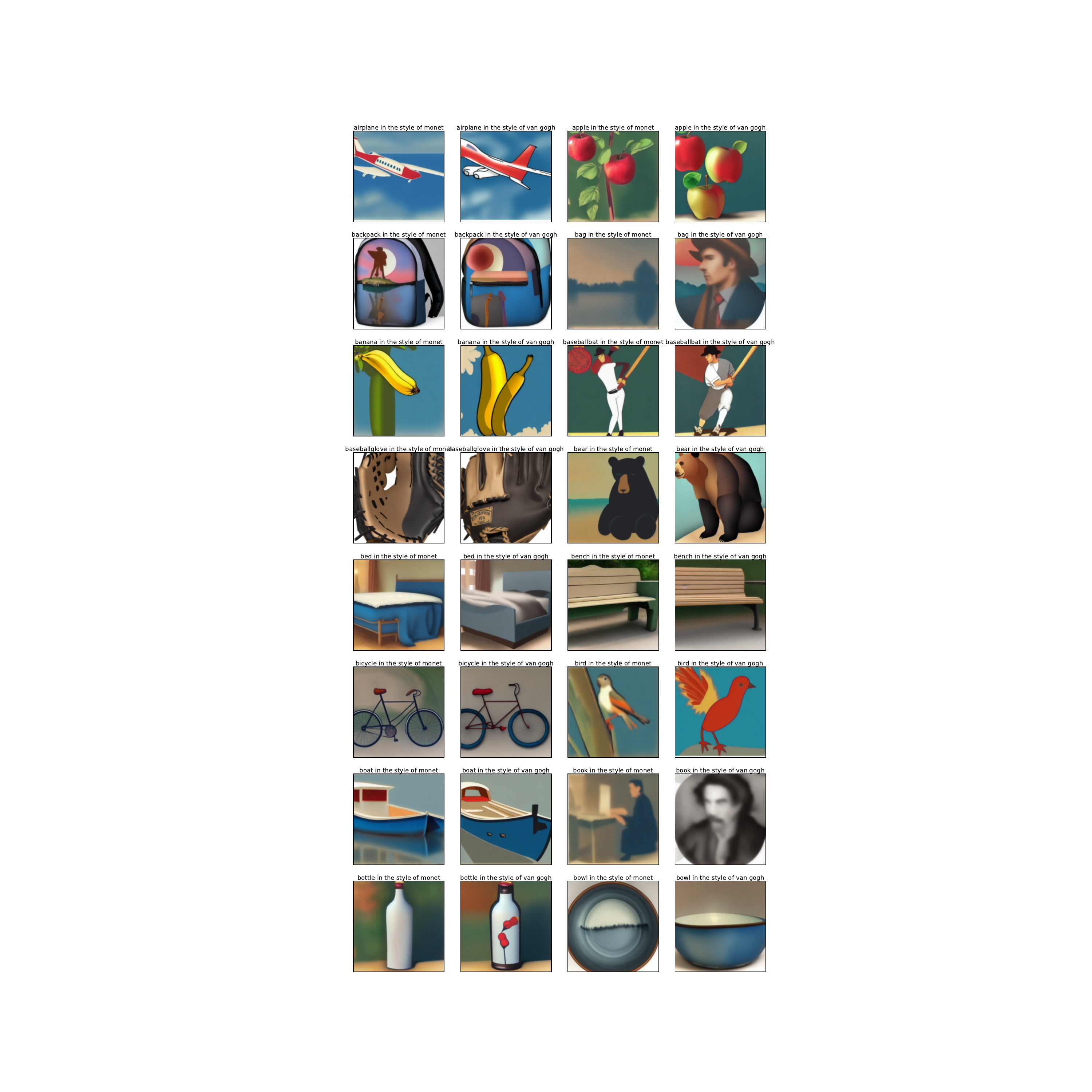}
  \vspace{-2.0cm}
    \caption{\label{non_causal_style_1} \textbf{Non-Causal State: down-blocks.2.attentions.1.transformer-blocks.0.attn2} We find that restoring the  down-blocks.2.attentions.1.transformer-blocks.0.attn2 in the UNet \textbf{does not} lead to generation of images with strong fidelity to the {\it style} attribute in the original caption for a majority of cases. }%
\end{figure}
\begin{figure}[H]
    \hskip -5.6cm
  \includegraphics[width=24.5cm, height=21.3cm]{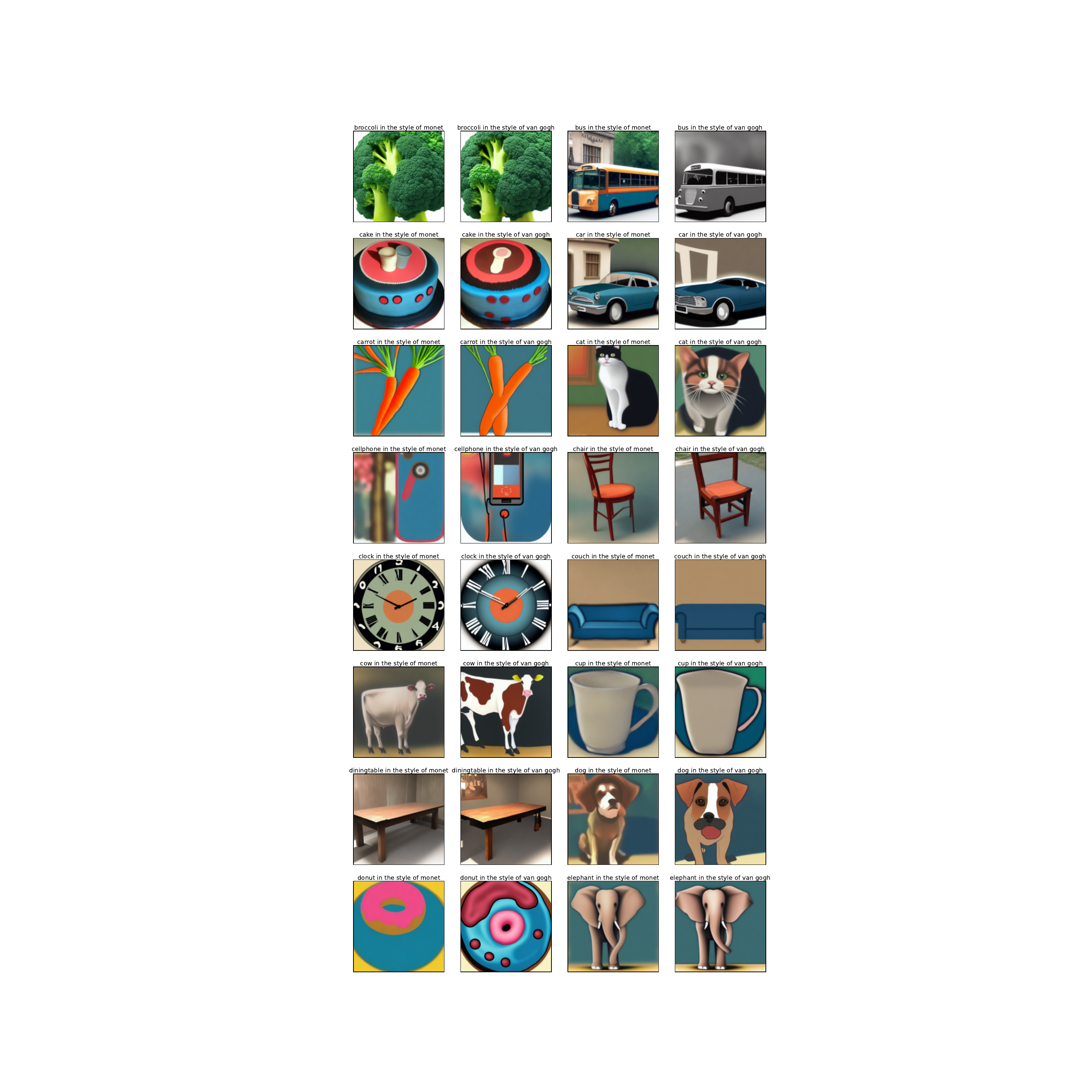}
  \vspace{-2.0cm}
    \caption{\label{non_causal_style_2} \textbf{Non-Causal State: down-blocks.2.attentions.1.transformer-blocks.0.attn2.} We find that restoring the down-blocks.2.attentions.1.transformer-blocks.0.attn2 which is the first self-attention layer in the UNet \textbf{does not} lead to generation of images with strong fidelity to the {\it style} attribute in the original caption for a majority of cases. }%
\end{figure}
\section{Qualitative Visualizations for Causal Tracing (Text-Encoder)}
\label{qual_text_encoder}
\begin{figure}[H]
    \hskip -4.2cm
  \includegraphics[width=22.5cm, height=20.3cm]{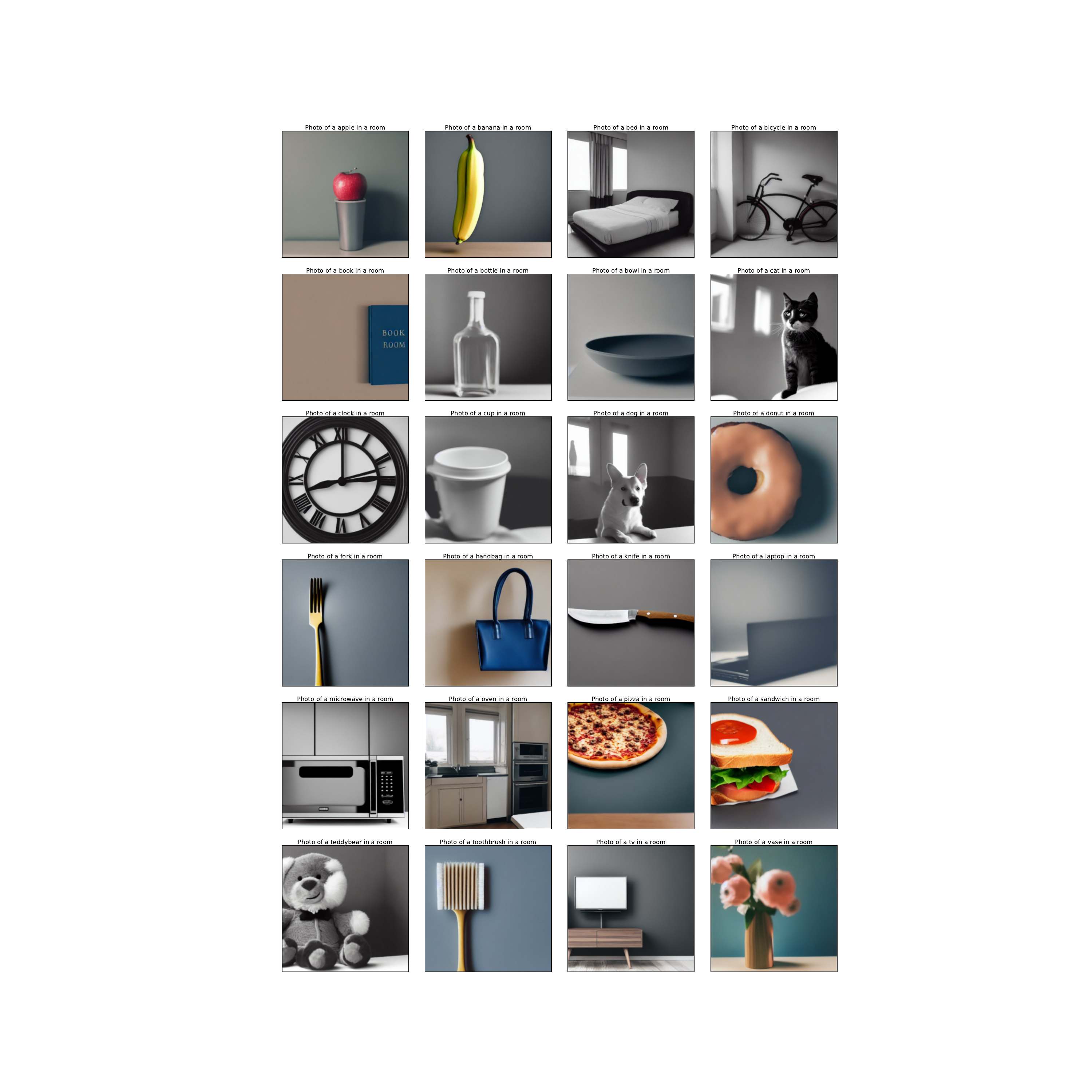}
  \vspace{-2.0cm}
    \caption{\label{text_encoder_causal} \textbf{Causal State: self-attn-0 corresponding to the last subject token.} We find that restoring the first self-attn layer which is the first self-attention layer in the text-encoder leads to generation of images with strong fidelity to the original caption for a majority of cases. }%
\end{figure}
\begin{figure}[H]
    \hskip -4.2cm
  \includegraphics[width=22.5cm, height=20.3cm]{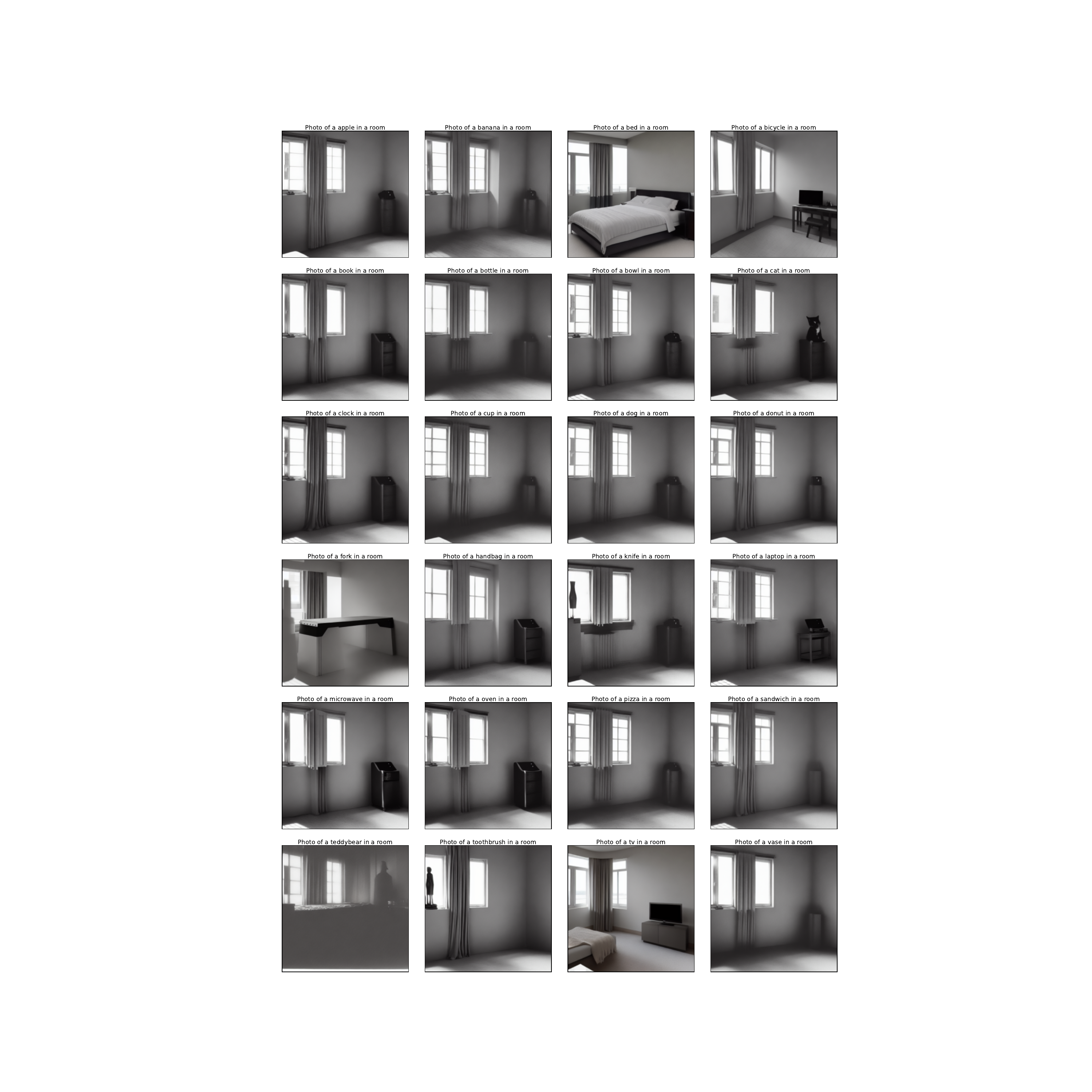}
  \vspace{-2.0cm}
    \caption{\label{text_encoder_non_causal} \textbf{Non-Causal State: self-attn-4 corresponding to the last subject token.} We find that restoring the fourth self-attn layer in the text-encoder \textbf{does not lead} to generation of images with strong fidelity to the original caption for a majority of cases. }%
\end{figure}
\begin{figure}[H]
    \hskip -4.2cm
  \includegraphics[width=22.5cm, height=20.3cm]{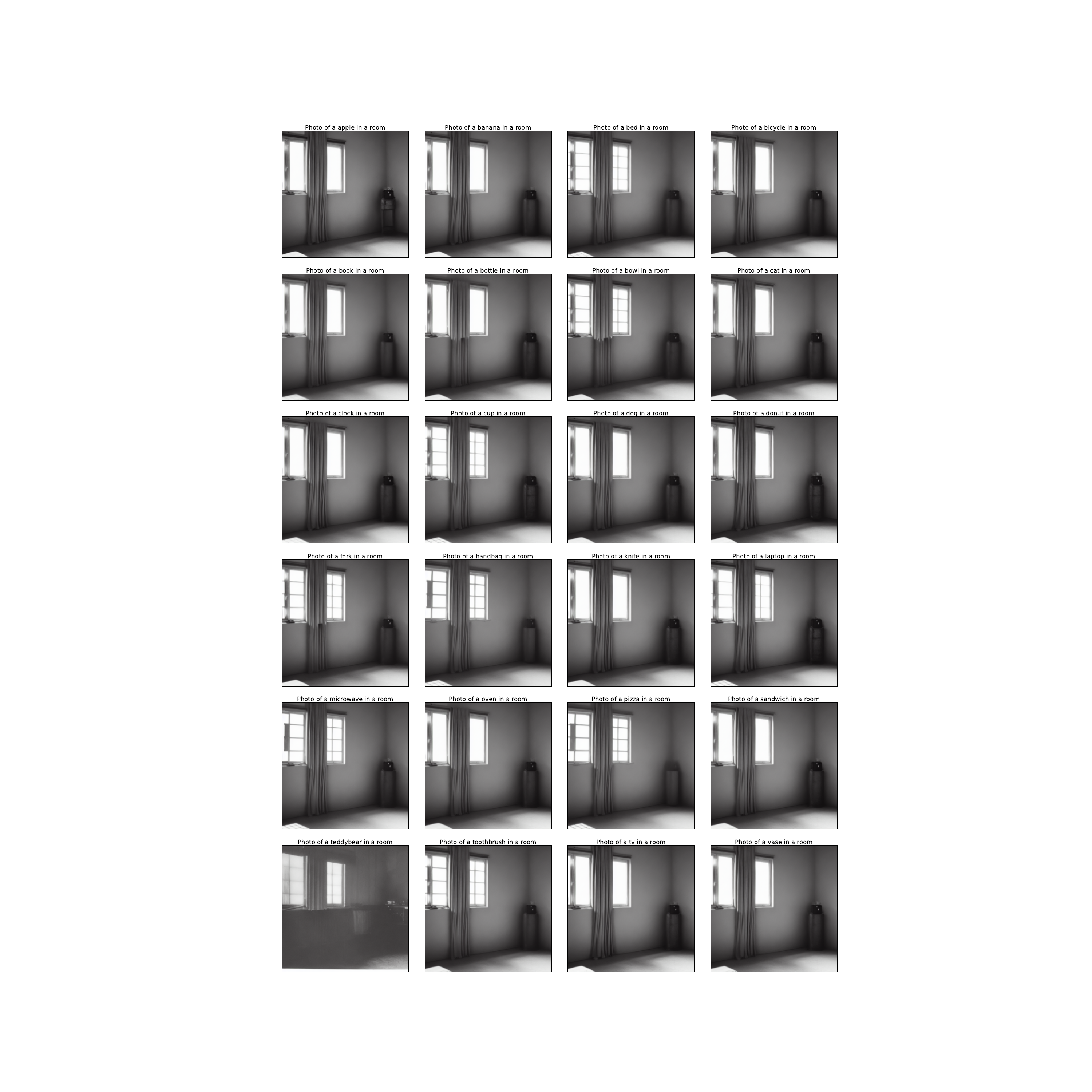}
  \vspace{-2.0cm}
    \caption{\label{text_encoder_non_causal_2} \textbf{Non-Causal State: self-attn-5 corresponding to the last subject token.} We find that restoring the fifth self-attn layer in the text-encoder \textbf{does not lead} to generation of images with strong fidelity to the original caption for a majority of cases. }%
\end{figure}

\section{Validation-Set Design For Causal Tracing}
\label{validation_design}
To select the threshold for \texttt{CLIP-Score} to be used at scale across the entirety of prompts (as shown in~\Cref{probe_dataset}), we use a small validation set of 10 prompts per attribute. In particular, we build a Jupyter notebook interface to select causal states for them. Once the causal states are marked, we select the common causal states across all the 10 prompts per attribute. Per causal state, we then compute the average \texttt{CLIP-Score} across the 10 prompts per attribute. Per attribute, we then select the lowest \texttt{CLIP-Score} corresponding to a causal state. These sets of \texttt{CLIP-Scores} per attribute is then used to filter the causal states and the non-causal states from the larger set of prompts in the probe dataset used in~\Cref{probe_dataset}. 
\begin{figure}[H]
    \hskip 2.0cm
  \includegraphics[width=8.5cm, height=4.0cm]{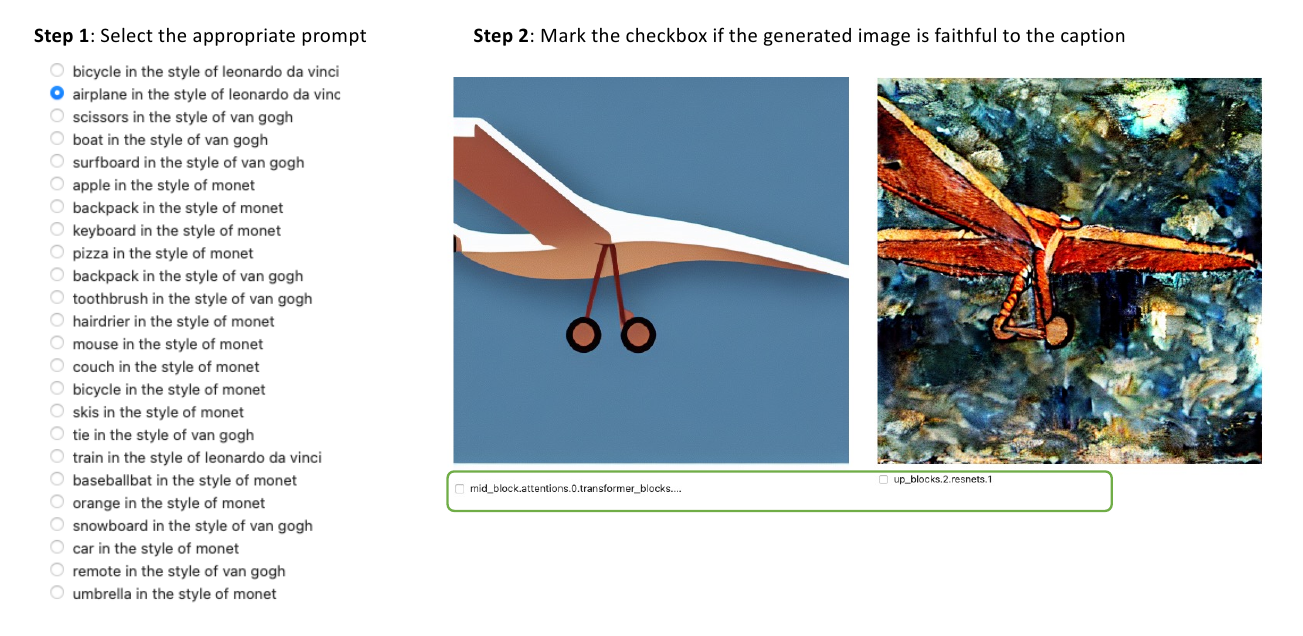}
    \caption{\label{validation_set} \textbf{Jupyter Notebook Interface for Marking Causal States.} }%
\end{figure}
\section{Causal Tracing for Viewpoint and Count}
\label{trace_viewpoint_count}
In this section, we provide additional causal tracing results for the {\it viewpoint} and {\it count} attribute. 
\subsection{Viewpoint}
\begin{figure}[H]
    \hskip 0.0cm
  \includegraphics[width=13.5cm, height=4.0cm]{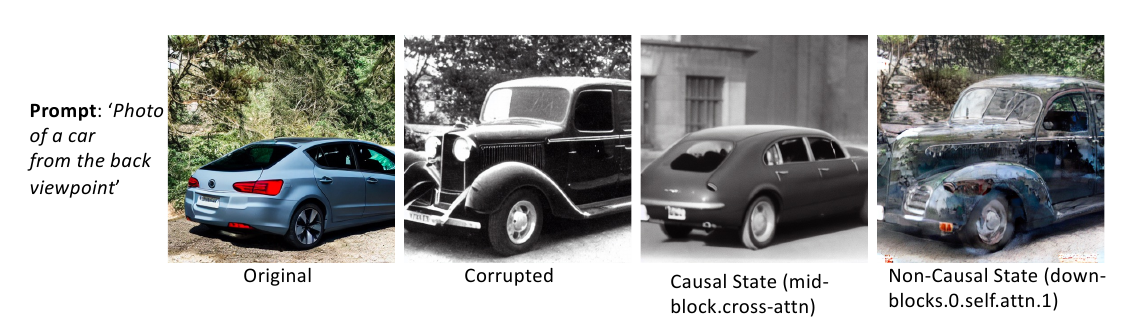}
    \caption{\label{viewpoint_causal} \textbf{Illustration of a causal state for the viewpoint attribute.} }%
\end{figure}
\subsection{Count}
For the {\it count attribute}, we find that public text-to-image generative models such as Stable-Diffusion cannot generate images with high-fidelity to the captions. Therefore, we do not use causal tracing for this attribute. 
\begin{figure}[H]
    \hskip 0.7cm
  \includegraphics[width=12.5cm, height=3.2cm]{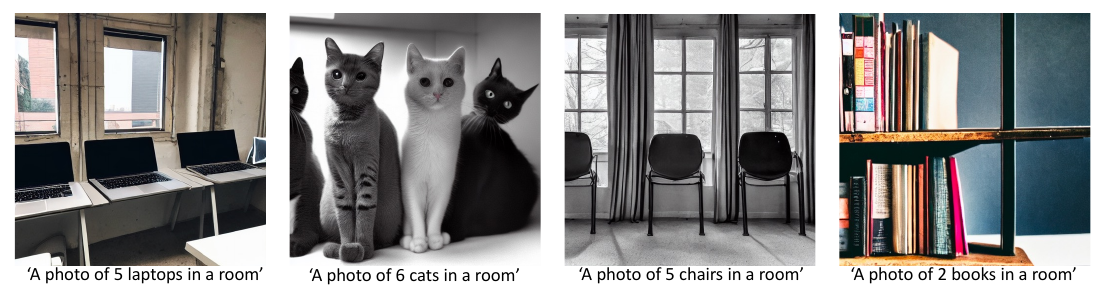}
    \caption{\label{count_causal} \textbf{Illustration of failure cases of generation for the {\it count} attribute with the Original Clean model.} }%
\end{figure}
\section{Qualitative Visualizations Using \difffix{} For Ablating Concepts}
\label{qual_editing_concepts}
\subsection{Ablating Artistic Styles}
\begin{figure}[H]
    \hskip -2.6cm
  \includegraphics[width=18.5cm, height=19.3cm]{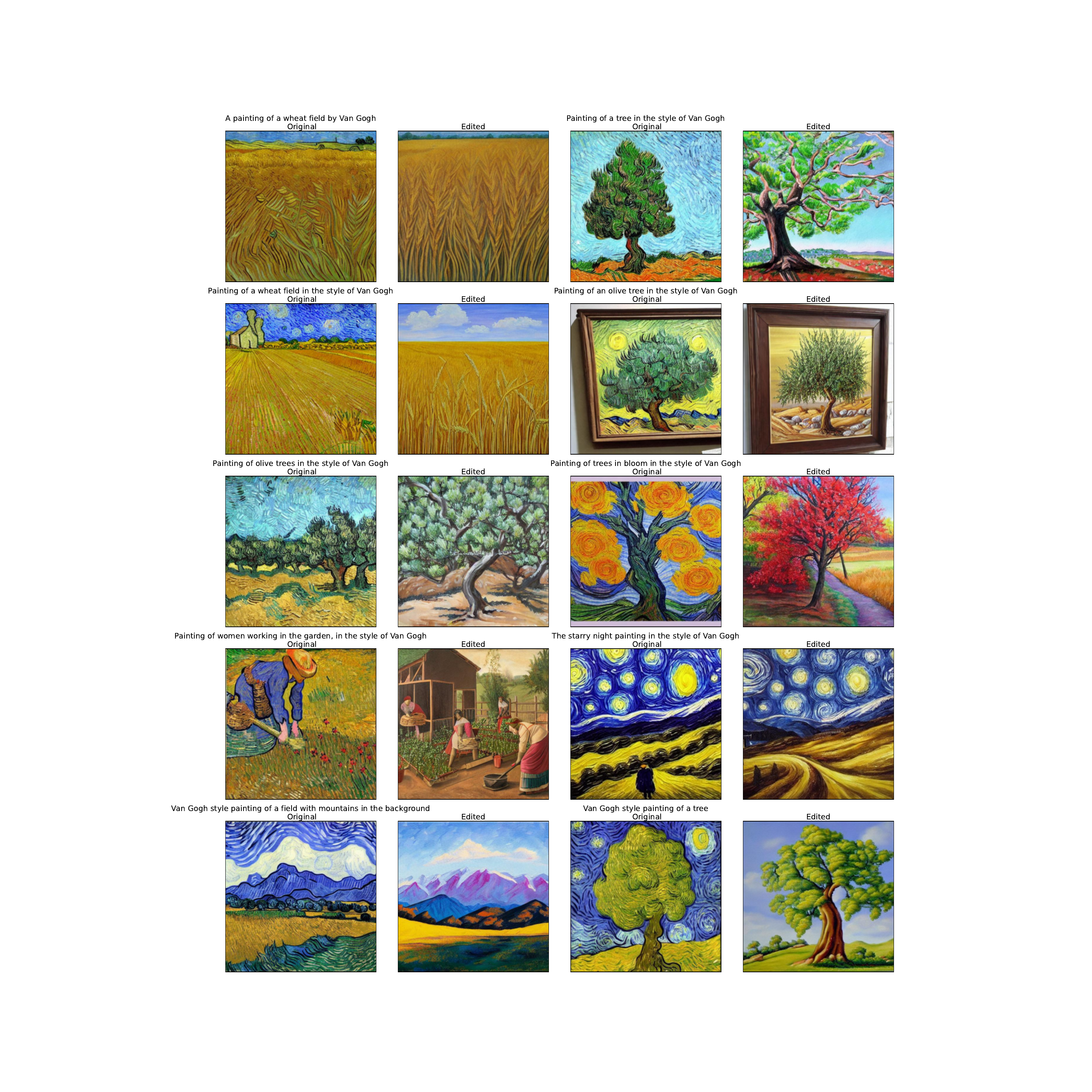}
  \vspace{-0.2cm}
    \caption{\label{single_edit_vangogh} \textbf{Single-Concept Ablated Model: Generations with different {\it Van Gogh} Prompts.} }%
\end{figure}
\begin{figure}[H]
    \hskip -2.6cm
  \includegraphics[width=18.5cm, height=19.3cm]{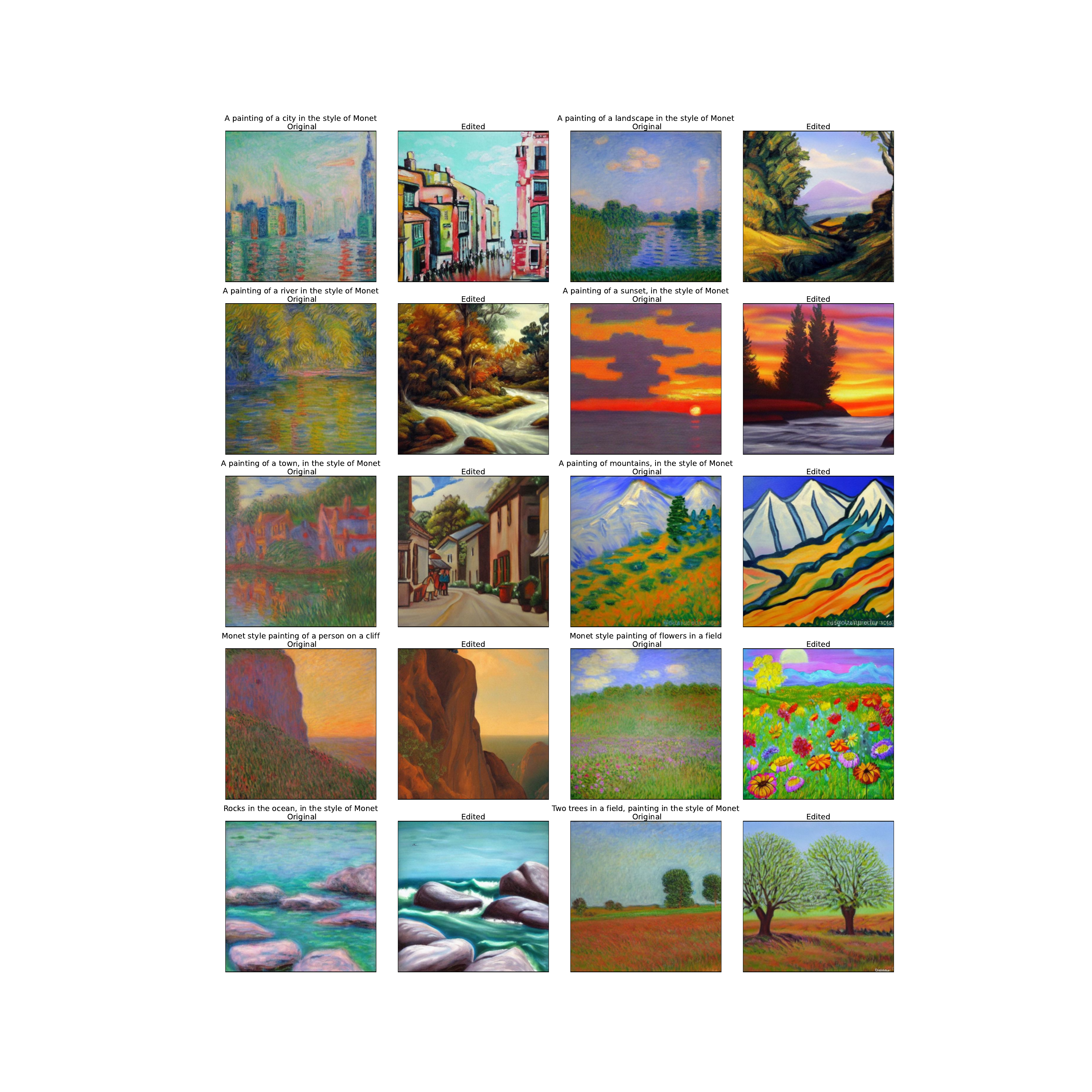}
  \vspace{-0.2cm}
    \caption{\label{single_edit_monet} \textbf{Single-Concept Ablated Model: Generations with different {\it Monet} Prompts.} }%
\end{figure}
\begin{figure}[H]
    \hskip -2.6cm
  \includegraphics[width=18.5cm, height=19.3cm]{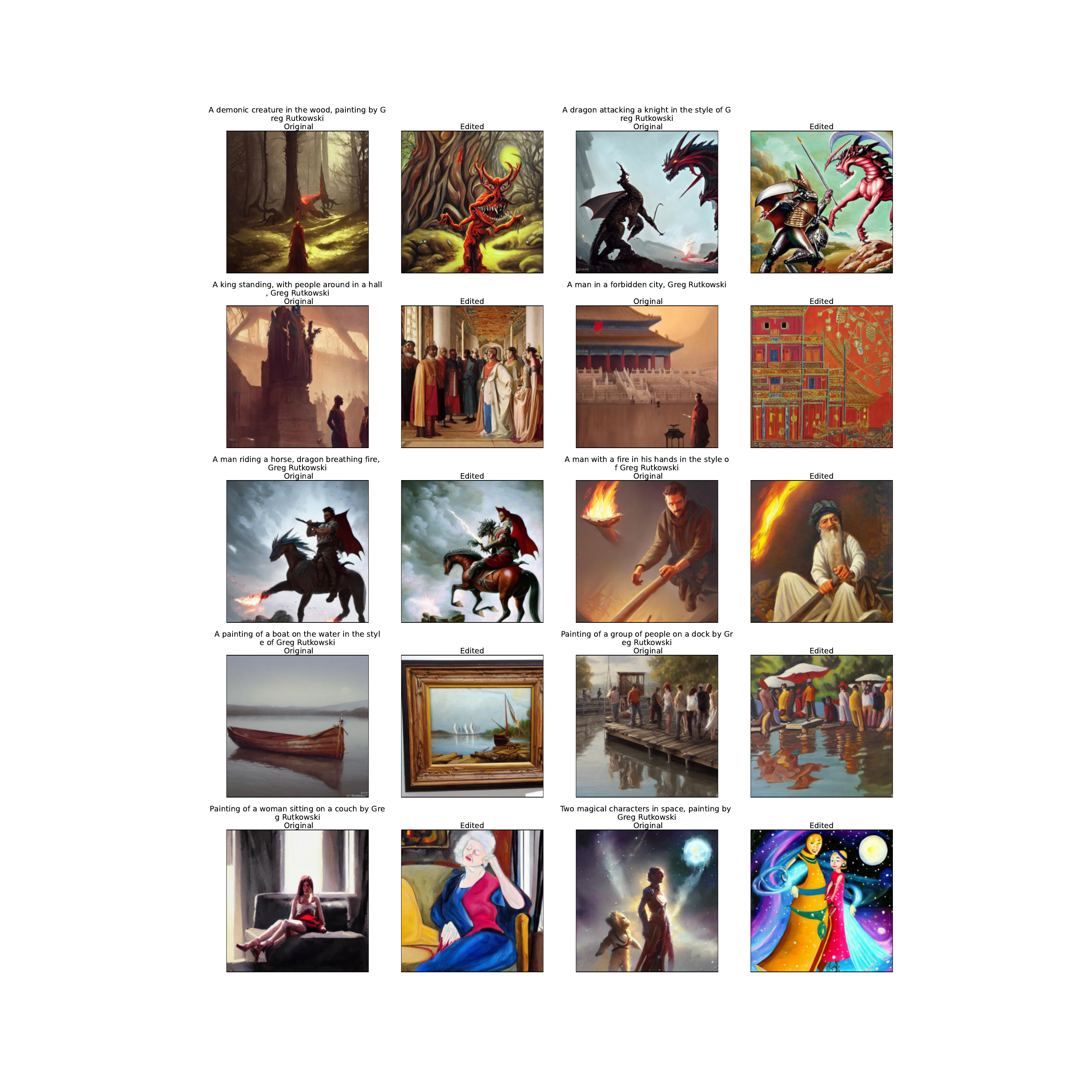}
  \vspace{-0.2cm}
    \caption{\label{single_edit_greg} \textbf{Single-Concept Ablated Model: Generations with different {\it Greg Rutkowski} Prompts.} }%
\end{figure}
\begin{figure}[H]
    \hskip -2.6cm
  \includegraphics[width=18.5cm, height=19.3cm]{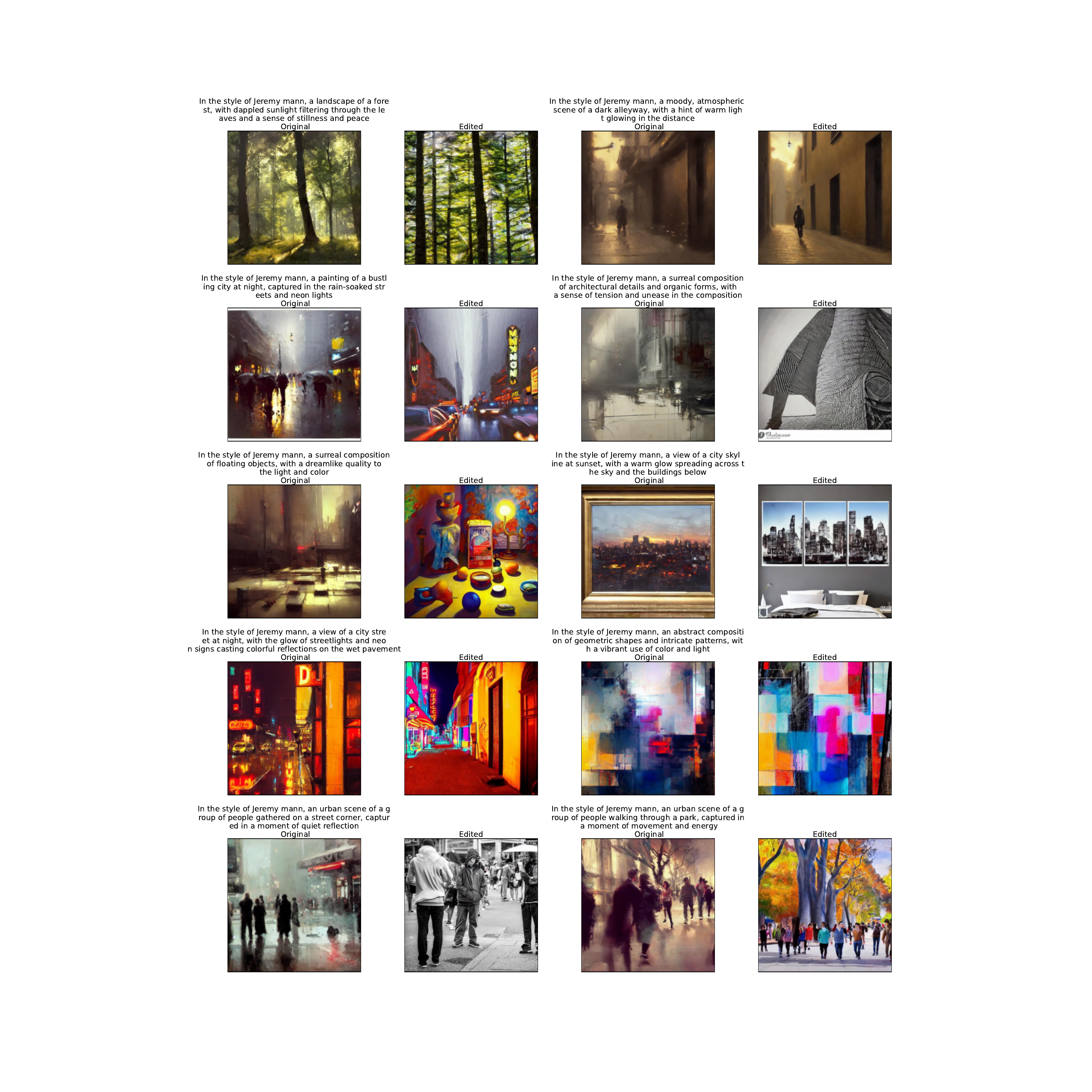}
  \vspace{-0.2cm}
    \caption{\label{single_edit_jeremy} \textbf{Single-Concept Ablated Model: Generations with different {\it Jeremy Mann} Prompts.} }%
\end{figure}
\begin{figure}[H]
    \hskip -2.6cm
  \includegraphics[width=18.5cm, height=19.3cm]{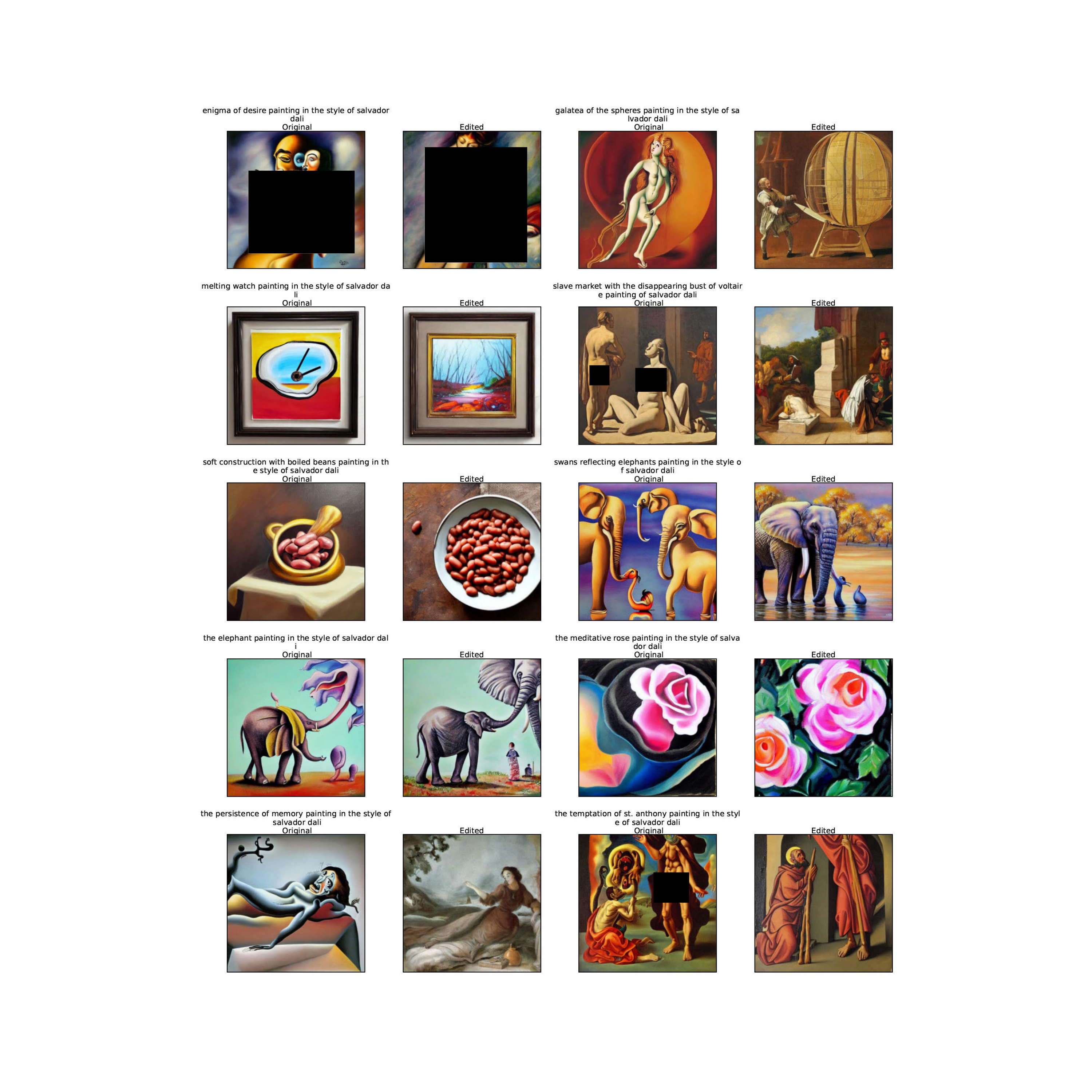}
  \vspace{-0.2cm}
    \caption{\label{single_edit_salvador} \textbf{Single-Concept Ablated Model: Generations with different {\it Salvador Dali} Prompts.} }%
\end{figure}
\subsection{Ablating Objects}
\begin{figure}[H]
    \hskip -2.6cm
  \includegraphics[width=18.5cm, height=19.3cm]{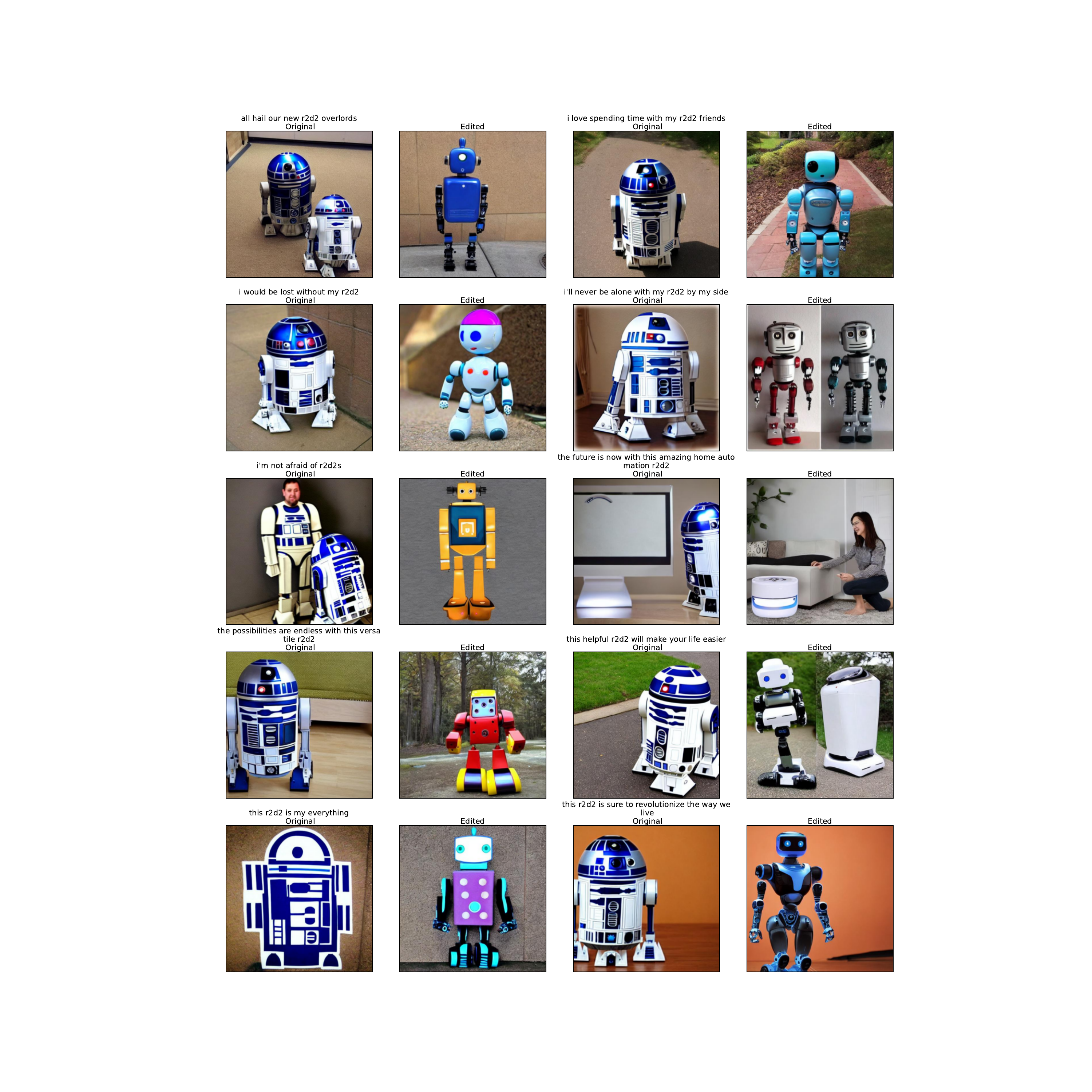}
  \vspace{-0.2cm}
    \caption{\label{single_edit_r2d2} \textbf{Single-Concept Ablated Model: Generations with different {\it R2D2} Prompts.} }%
\end{figure}
\begin{figure}[H]
    \hskip -2.6cm
  \includegraphics[width=18.5cm, height=19.3cm]{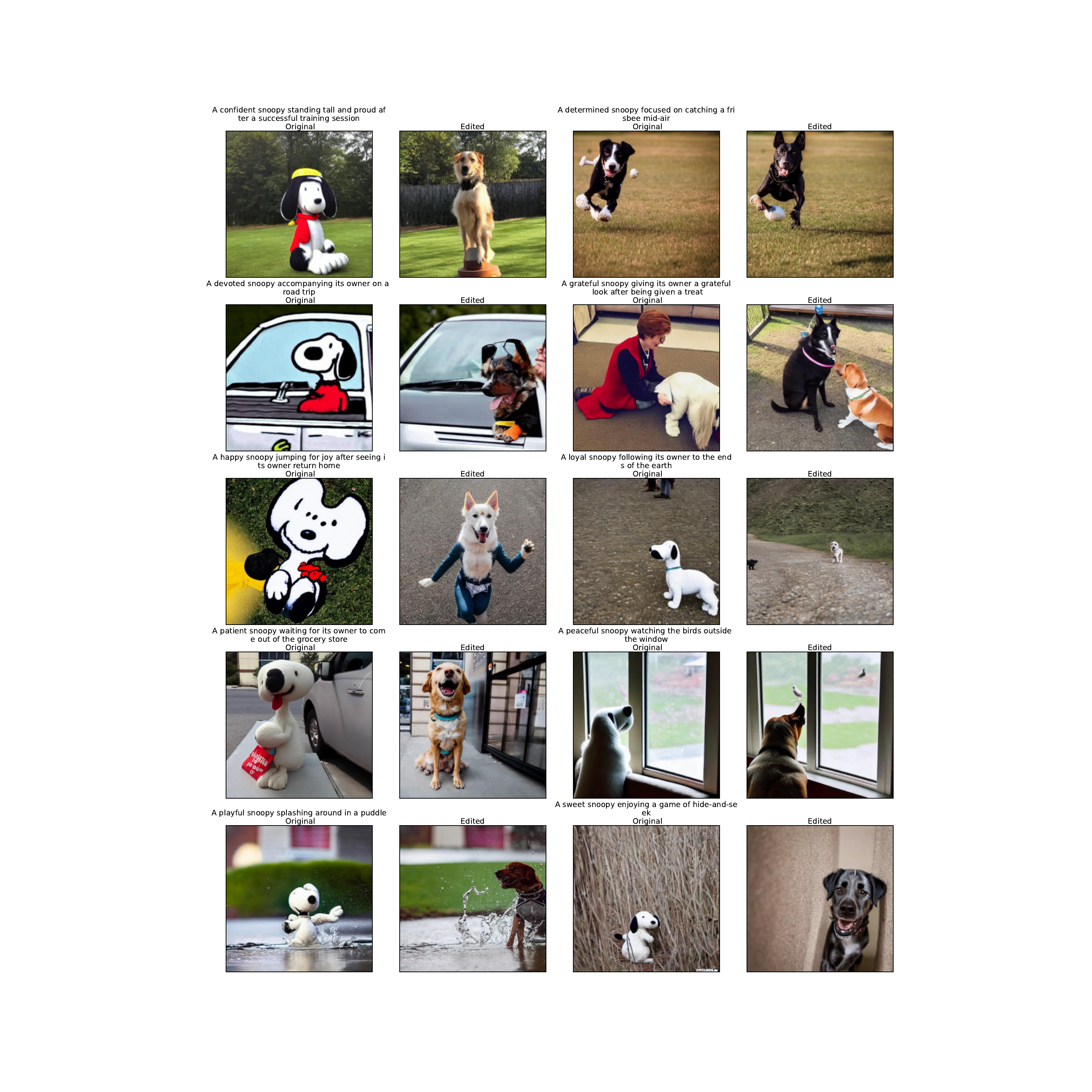}
  \vspace{-0.2cm}
    \caption{\label{single_edit_snoopy} \textbf{Single-Concept Ablated Model: Generations with different {\it Snoopy} Prompts.} }%
\end{figure}
\begin{figure}[H]
    \hskip -2.6cm
  \includegraphics[width=18.5cm, height=19.3cm]{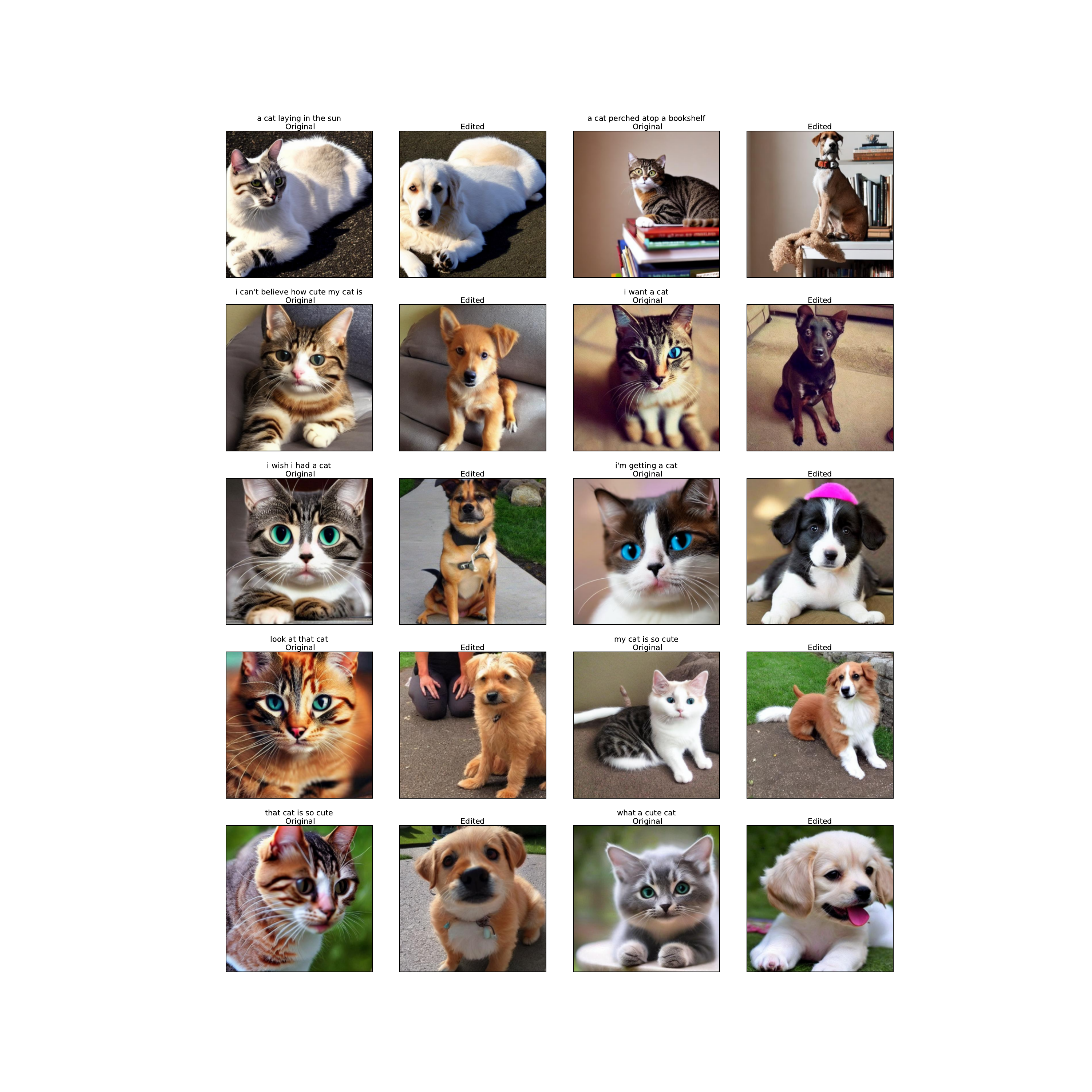}
  \vspace{-0.2cm}
    \caption{\label{single_edit_cat} \textbf{Single-Concept Ablated Model: Generations with different {\it Cat} Prompts.} }%
\end{figure}
\begin{figure}[H]
    \hskip -2.6cm
  \includegraphics[width=18.5cm, height=19.3cm]{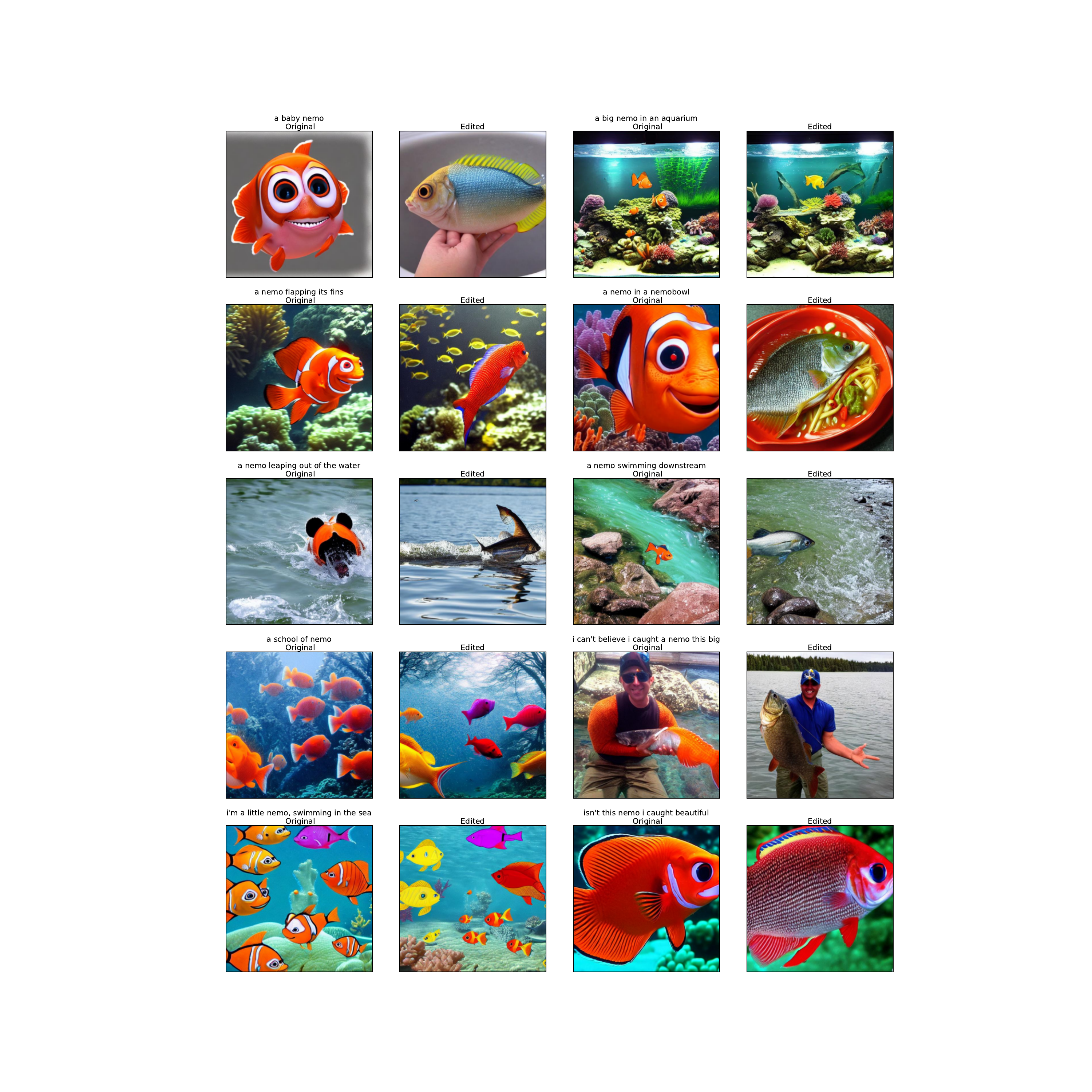}
  \vspace{-0.2cm}
    \caption{\label{single_edit_nemo} \textbf{Single-Concept Ablated Model: Generations with different {\it Nemo} Prompts.} }%
\end{figure}
\begin{figure}[H]
    \hskip -2.6cm
  \includegraphics[width=18.5cm, height=19.3cm]{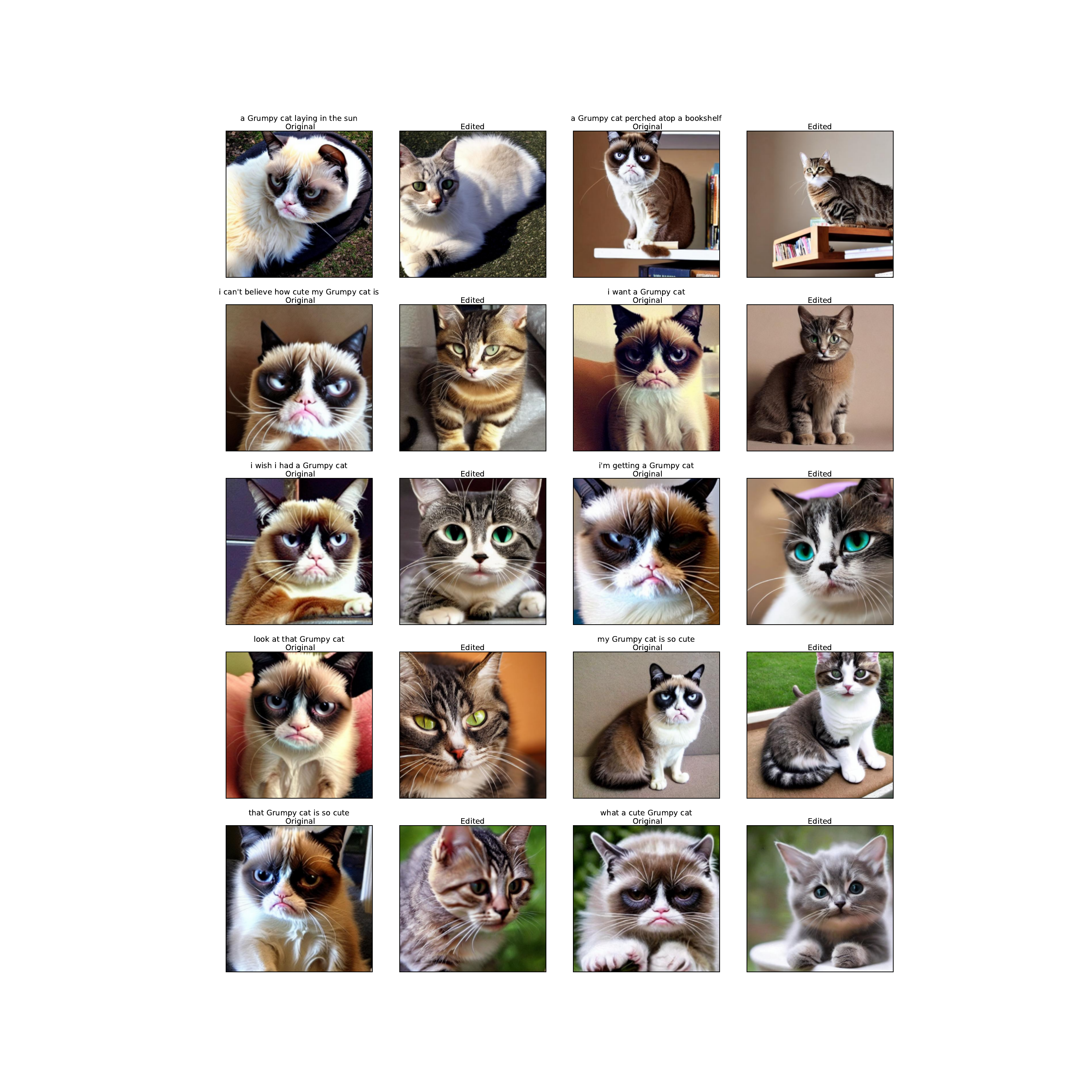}
  \vspace{-0.2cm}
    \caption{\label{single_edit_grumpy} \textbf{Single-Concept Ablated Model: Generations with different {\it Grumpy Cat} Prompts.} }%
\end{figure}
\subsection{Updating Facts}
\begin{figure}[H]
    \hskip -0.6cm
  \includegraphics[width=15.5cm, height=15.5cm]{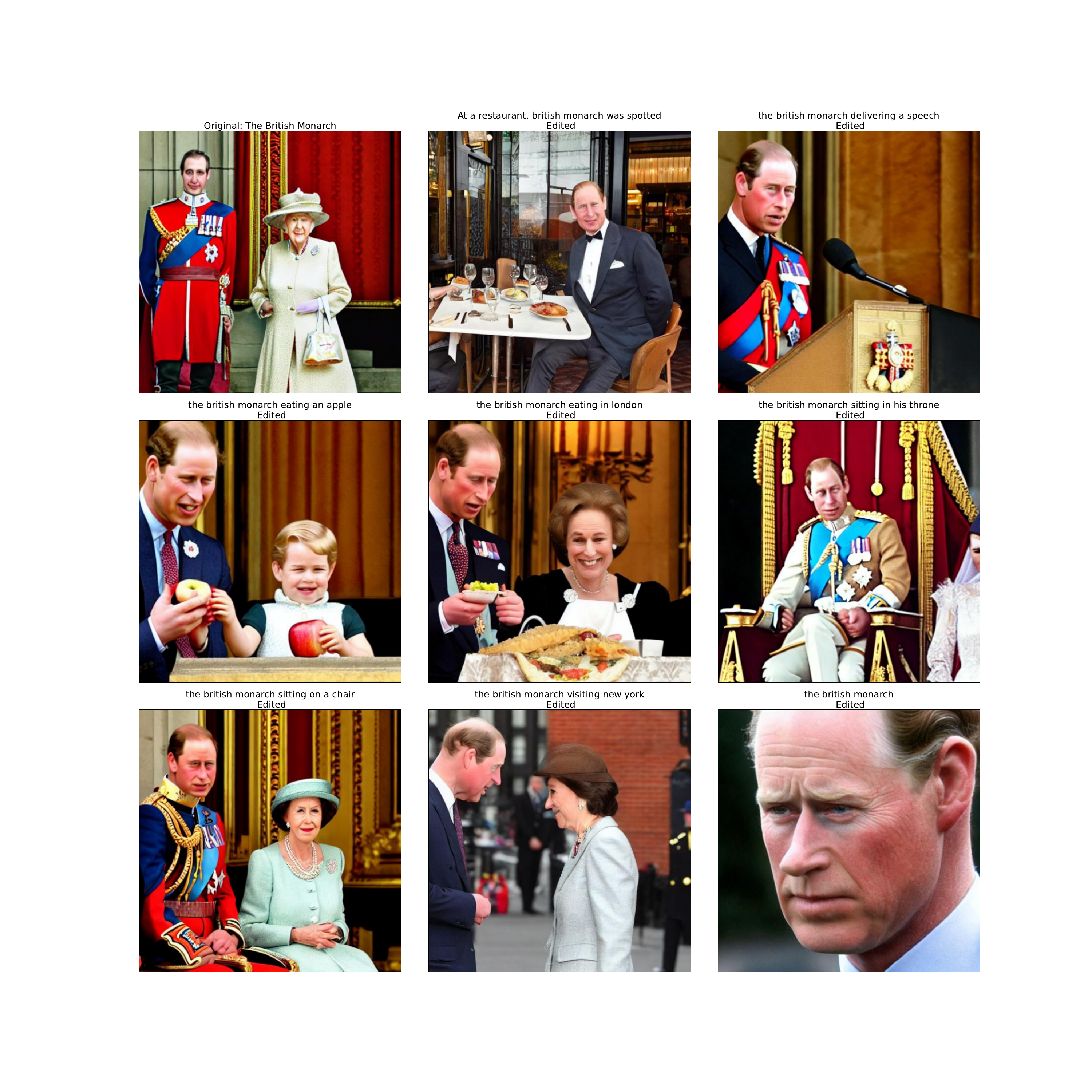}
  \vspace{-0.2cm}
    \caption{\label{single_edit_fact_monarch} \textbf{Single-Concept Ablated Model: Generations with different prompts containing {\it The British Monarch}.} The first image is the one from the unedited text-to-image model which shows the Queen as the original generation. The edited model is consistently able to generate the correct British Monarch : Prince Charles.}%
\end{figure}
\begin{figure}[H]
    \hskip -0.6cm
  \includegraphics[width=15.5cm, height=15.5cm]{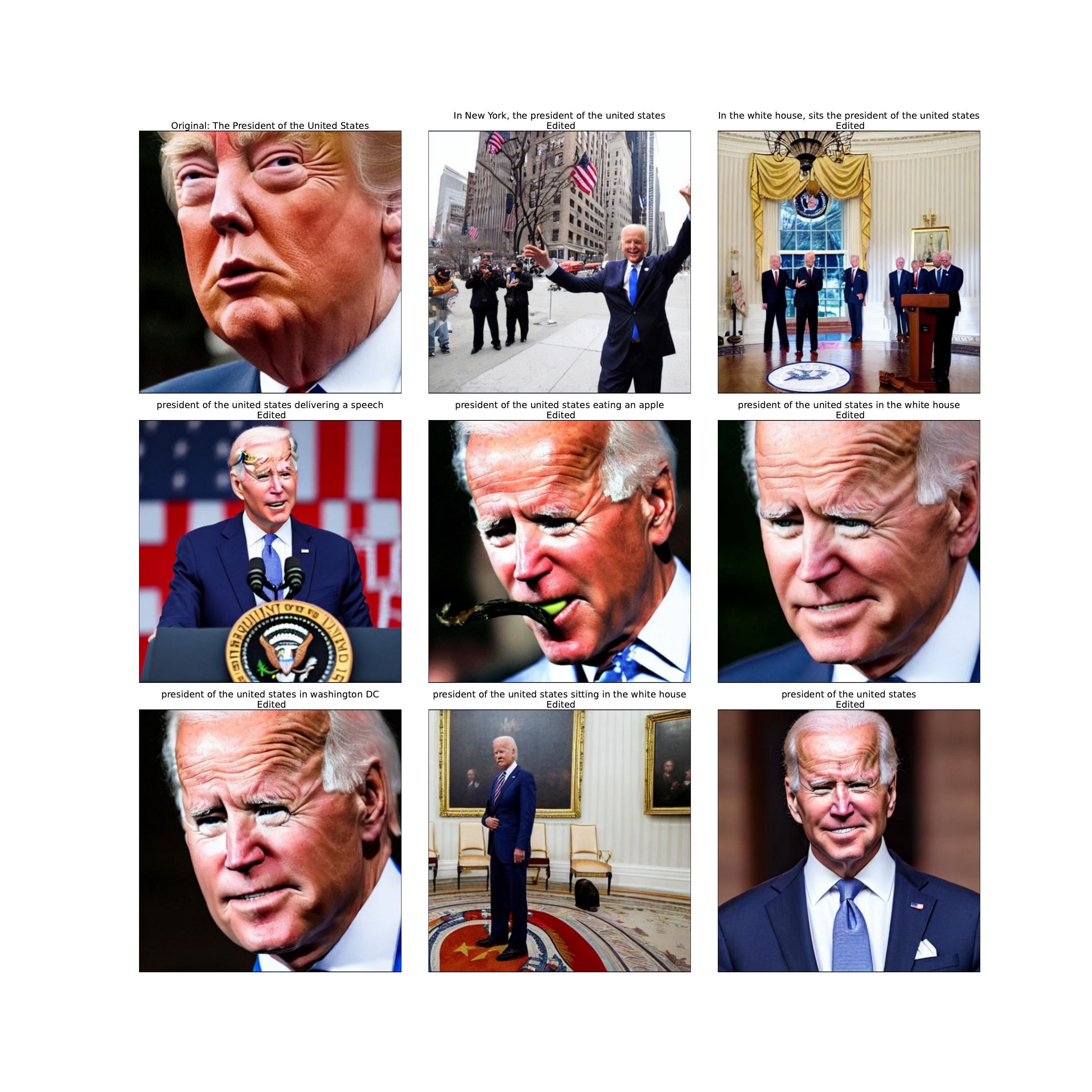}
  \vspace{-0.2cm}
    \caption{\label{single_edit_fact_president} \textbf{Single-Concept Ablated Model: Generations with different prompts containing {\it The President of the United States}.} The first image is the one from the unedited text-to-image model.}%
\end{figure}

\section{Qualitative Visualizations for Editing Non-Causal Layers}
\label{qual_editing_non_causal}
\begin{figure}[H]
    \hskip 0.8cm
  \includegraphics[width=12.5cm, height=4.2cm]{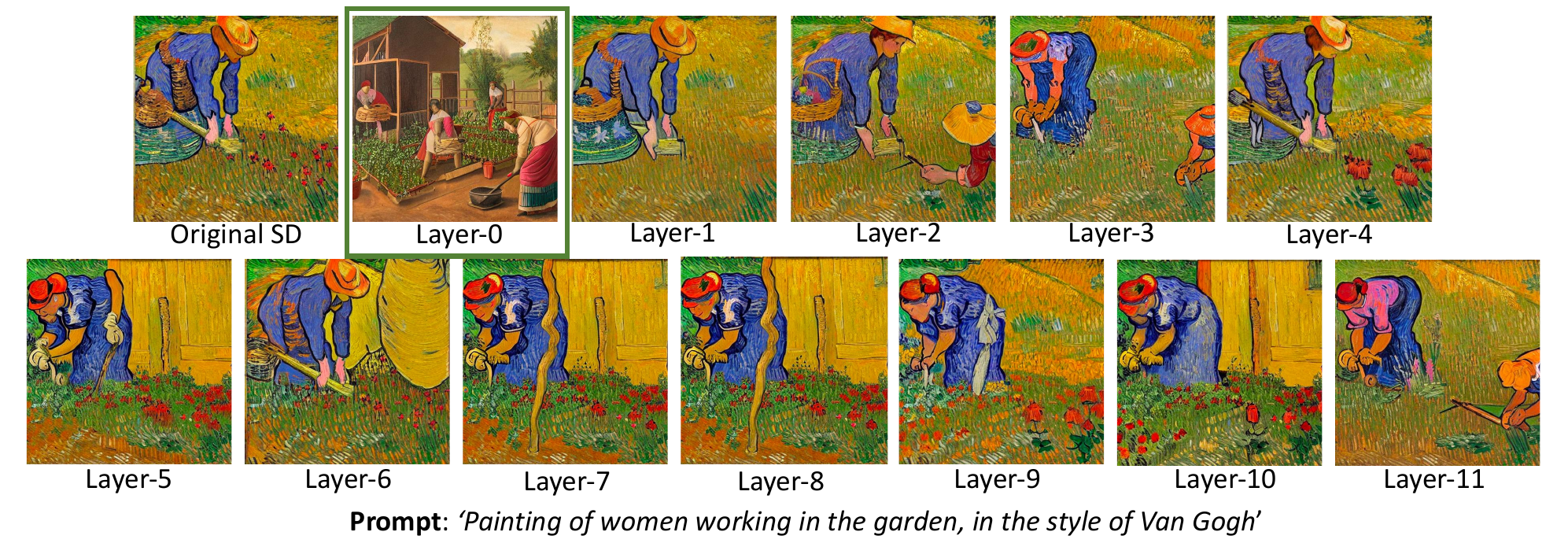}
  \vspace{-0.2cm}
    \caption{\label{edit_viz_non_causal} \textbf{Editing only the causal layer (self-attn Layer-0) leads to intended model changes.} In this figure, we qualitatively show that the style of {\it `Van Gogh'} can be removed from the underlying text-to-image model, if the edit is performed at the correct causal site. Editing the non-causal layers using~\difffix{} leads to generations similar to the original unedited model. }%
\end{figure}
\begin{figure}[H]
    \hskip 0.0cm
  \includegraphics[width=14.5cm, height=5.3cm]{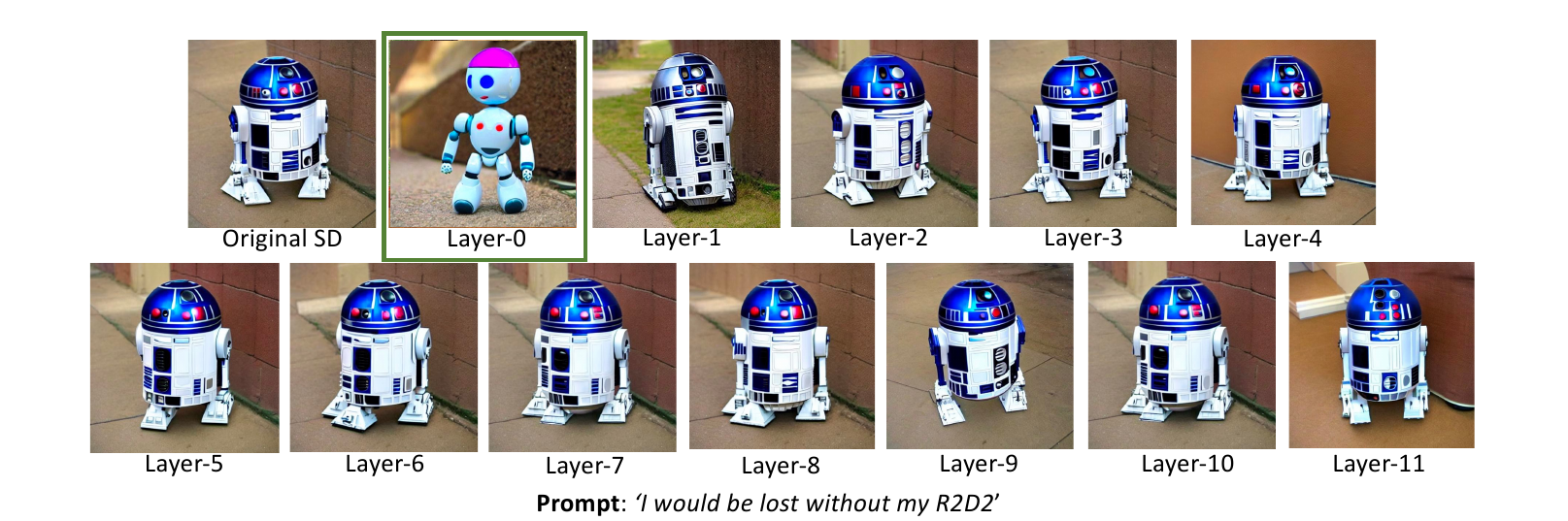}
  \vspace{-0.2cm}
    \caption{\label{edit_viz_non_causal_2} \textbf{Editing only the causal layer (self-attn Layer-0) leads to intended model changes.} In this figure, we qualitatively show that the object : {\it `R2D2'} can be removed from the underlying text-to-image model and be replaced with a generic robot, if the edit is performed at the correct causal site. Editing the non-causal layers using~\difffix{} leads to generations similar to the original unedited model. }%
\end{figure}
\section{Multi-Concept Ablated Model}
\label{multi_concept_ablated}
\begin{figure}[H]
    \hskip 0.0cm
  \includegraphics[width=14.5cm, height=5.3cm]{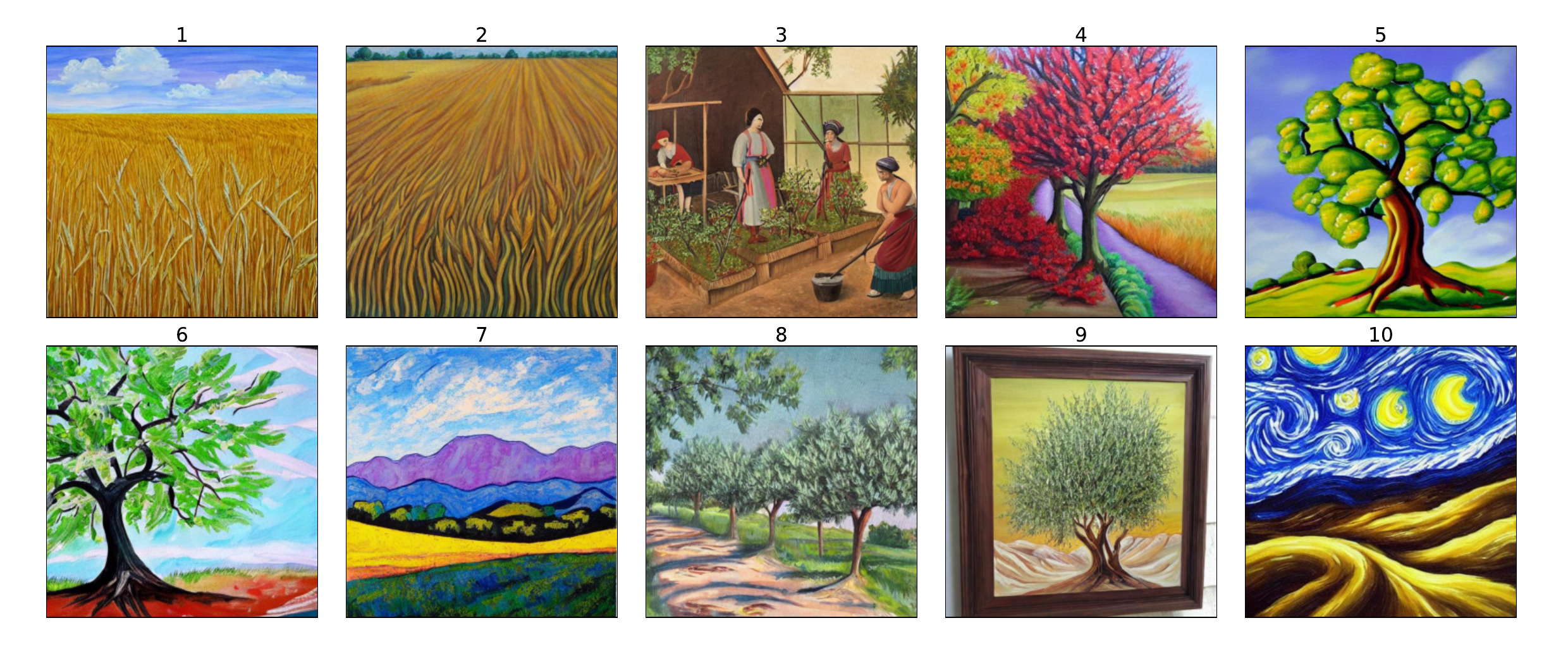}
  \vspace{-0.2cm}
    \caption{\label{multi_edit_vangogh} \textbf{Multi-Concept Ablated Model: Generations with different {\it Van Gogh} Prompts.} The multi-concept ablated model cannot generate images in the style of {\it Van Gogh} across various prompts containing {\it Van Gogh}.
    (1). A painting of a wheat field by Van Gogh;
    (2). Painting of a wheat field in the style of Van Gogh;
    (3). Painting of women working in the garden, in the style of Van Gogh;
    (4). Painting of trees in bloom in the style of Van Gogh;
    (5). Painting of a tree in the style of Van Gogh;
    (6). Van Gogh style painting of a tree;
    (7). Van Gogh style painting of a field with mountains in the background;
    (8). Painting of olive trees in the style of Van Gogh
    (9). Painting of an olive tree in the style of Van Gogh;
    (10). The starry night painting in the style of Van Gogh.
    }%
\end{figure}

\begin{figure}[H]
    \hskip 0.0cm
  \includegraphics[width=14.5cm, height=5.3cm]{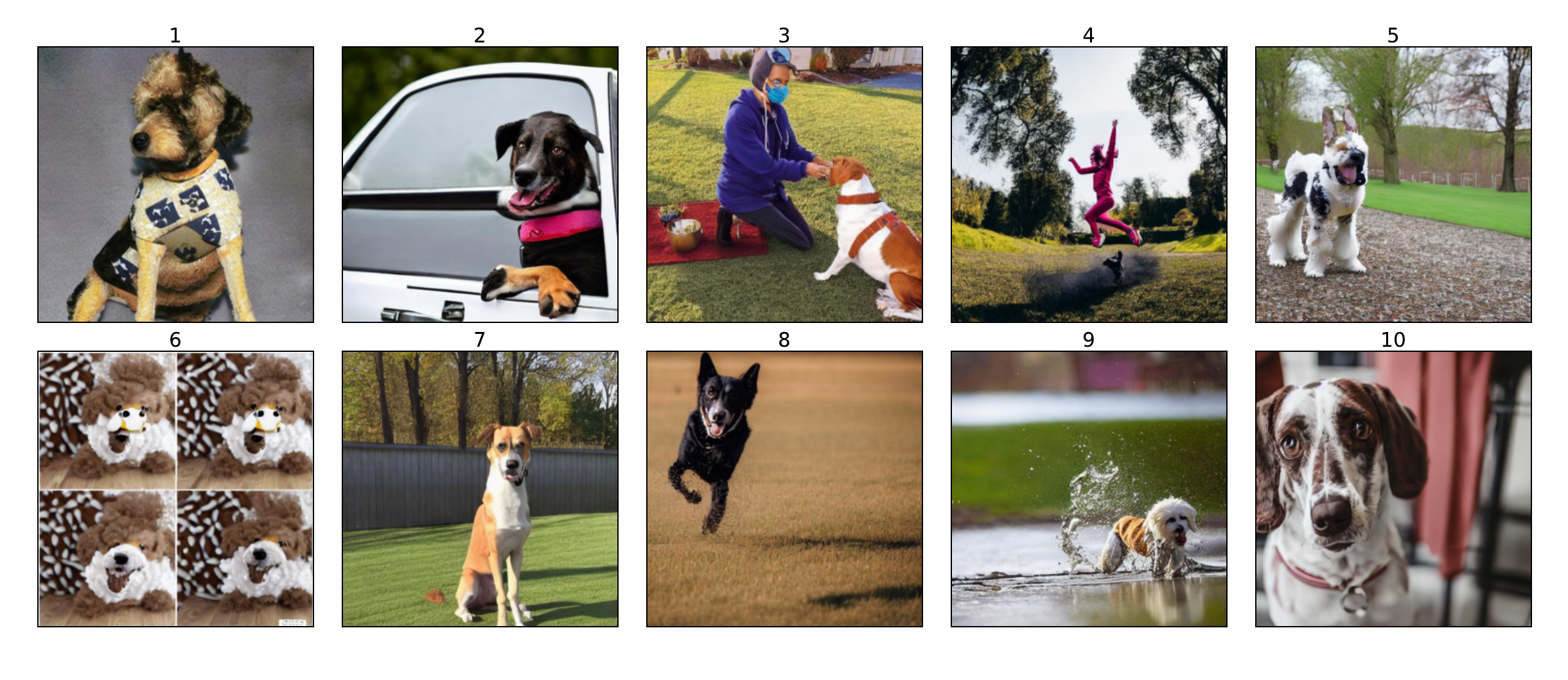}
  \vspace{-0.2cm}
    \caption{\label{multi_edit_snoopy} \textbf{Multi-Concept Ablated Model: Generations with different {\it Snoopy} Prompts.} The multi-concept ablated model cannot generate images containing the specific dog {\it Snoopy} with various prompts containing {\it Snoopy}.
    (1). A confident snoopy standing tall and proud after a successful training session;
    (2). A peaceful snoopy watching the birds outside the window;
    (3). A grateful snoopy giving its owner a grateful look after being given a treat;
    (4). A happy snoopy jumping for joy after seeing its owner return home;
    (5). A devoted snoopy accompanying its owner on a road trip
    (6). A sweet snoopy enjoying a game of hide-and-seek;
    (7). A loyal snoopy following its owner to the ends of the earth
    (8). A determined snoopy focused on catching a frisbee mid-air;
    (9). A playful snoopy splashing around in a puddle;
    (10).A patient snoopy waiting for its owner to come out of the grocery store;
    }%
\end{figure}

\begin{figure}[H]
    \hskip 0.0cm
  \includegraphics[width=14.5cm, height=5.3cm]{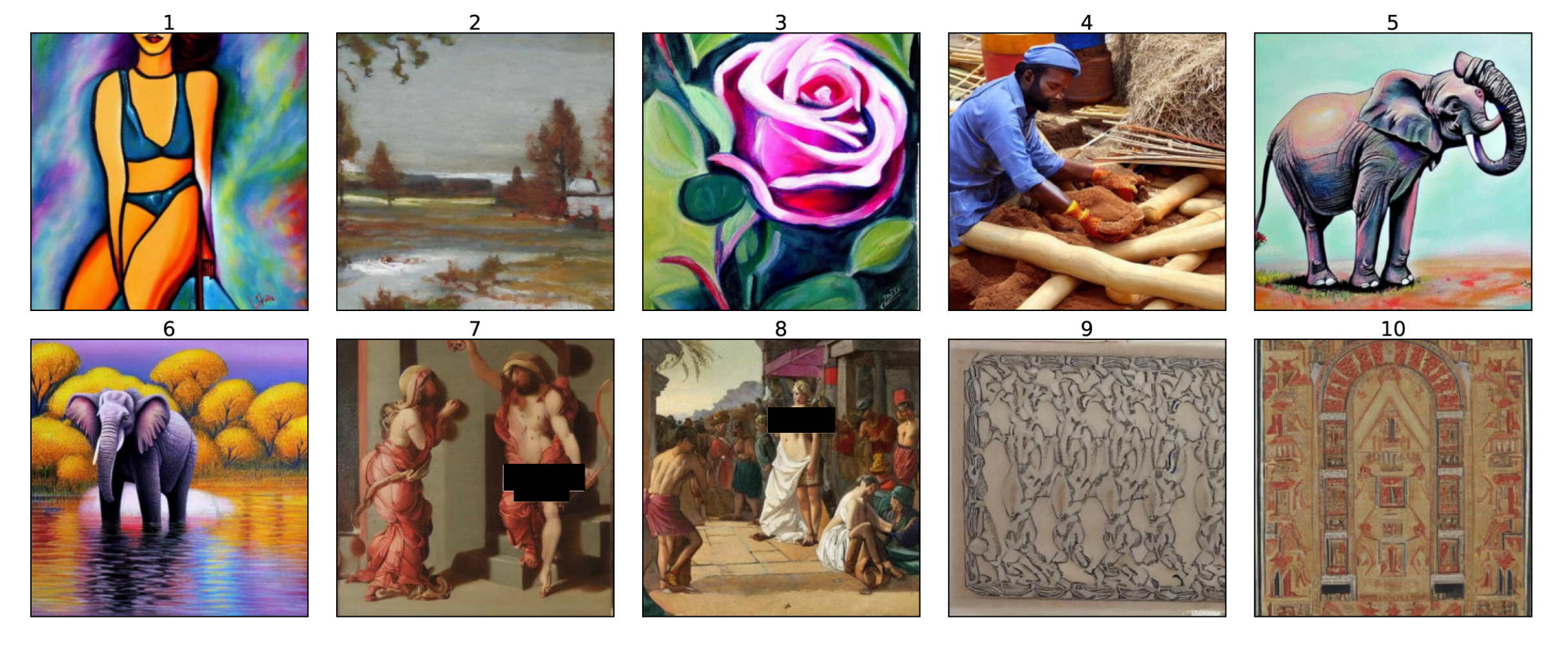}
  \vspace{-0.2cm}
    \caption{\label{multi_edit_dali} \textbf{Multi-Concept Ablated Model: Generations with different {\it Salvador Dali} Prompts.} The multi-concept model cannot generate images in the style of the artist {\it Salvador Dali} across various prompts containing {\it Salvador Dali}. 
    (1). enigma of desire painting in the style of salvador dali;
    (2). the persistence of memory painting in the style of salvador dali;
    (3). the meditative rose painting in the style of salvador dali;
    (4). soft construction with boiled beans painting in the style of salvador dali;
    (5). the elephant painting in the style of salvador dali;
    (6). swans reflecting elephants painting in the style of salvador dali;
    (7). the temptation of st. anthony painting in the style of salvador dali;
    (8). slave market with the disappearing bust of voltaire painting of salvador dali;
    (9). melting watch painting in the style of salvador dali;
    (10). galatea of the spheres painting in the style of salvador dali;
    }%
\end{figure}

\begin{figure}[H]
    \hskip 0.0cm
  \includegraphics[width=14.5cm, height=5.3cm]{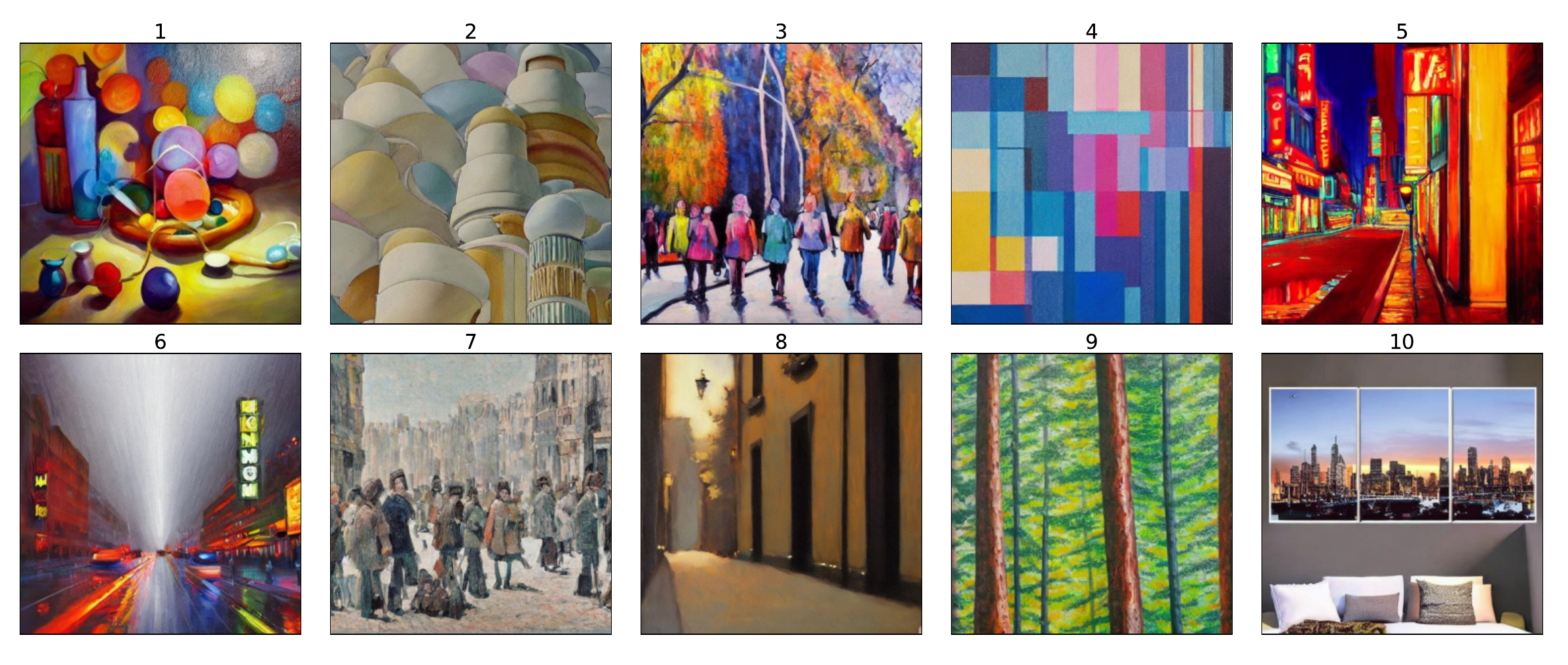}
  \vspace{-0.2cm}
    \caption{\label{multi_edit_jeremy_mann} \textbf{Multi-Concept Ablated Model: Generations with different {\it Jeremy Mann} Prompts.} The multi-concept ablated model cannot generate images in the style of the artist {\it Jeremy Mann}. 
    (1). In the style of Jeremy mann, a surreal composition of floating objects, with a dreamlike quality to the light and color;
    (2). In the style of Jeremy mann, a surreal composition of architectural details and organic forms, with a sense of tension and unease in the composition;
    (3).In the style of Jeremy mann, an urban scene of a group of people walking through a park, captured in a moment of movement and energy;
    (4). In the style of Jeremy mann, an abstract composition of geometric shapes and intricate patterns, with a vibrant use of color and light;
    (5).In the style of Jeremy mann, a view of a city street at night, with the glow of streetlights and neon signs casting colorful reflections on the wet pavement;
    (6).In the style of Jeremy mann, a painting of a bustling city at night, captured in the rain-soaked streets and neon lights;
    (7). In the style of Jeremy mann, an urban scene of a group of people gathered on a street corner, captured in a moment of quiet reflection;
    (8). In the style of Jeremy mann, a moody, atmospheric scene of a dark alleyway, with a hint of warm light glowing in the distance;
    (9). In the style of Jeremy mann, a landscape of a forest, with dappled sunlight filtering through the leaves and a sense of stillness and peace;
    (10). In the style of Jeremy mann, a view of a city skyline at sunset, with a warm glow spreading across the sky and the buildings below;
    }%
\end{figure}

\begin{figure}[H]
    \hskip 0.0cm
  \includegraphics[width=14.5cm, height=5.3cm]{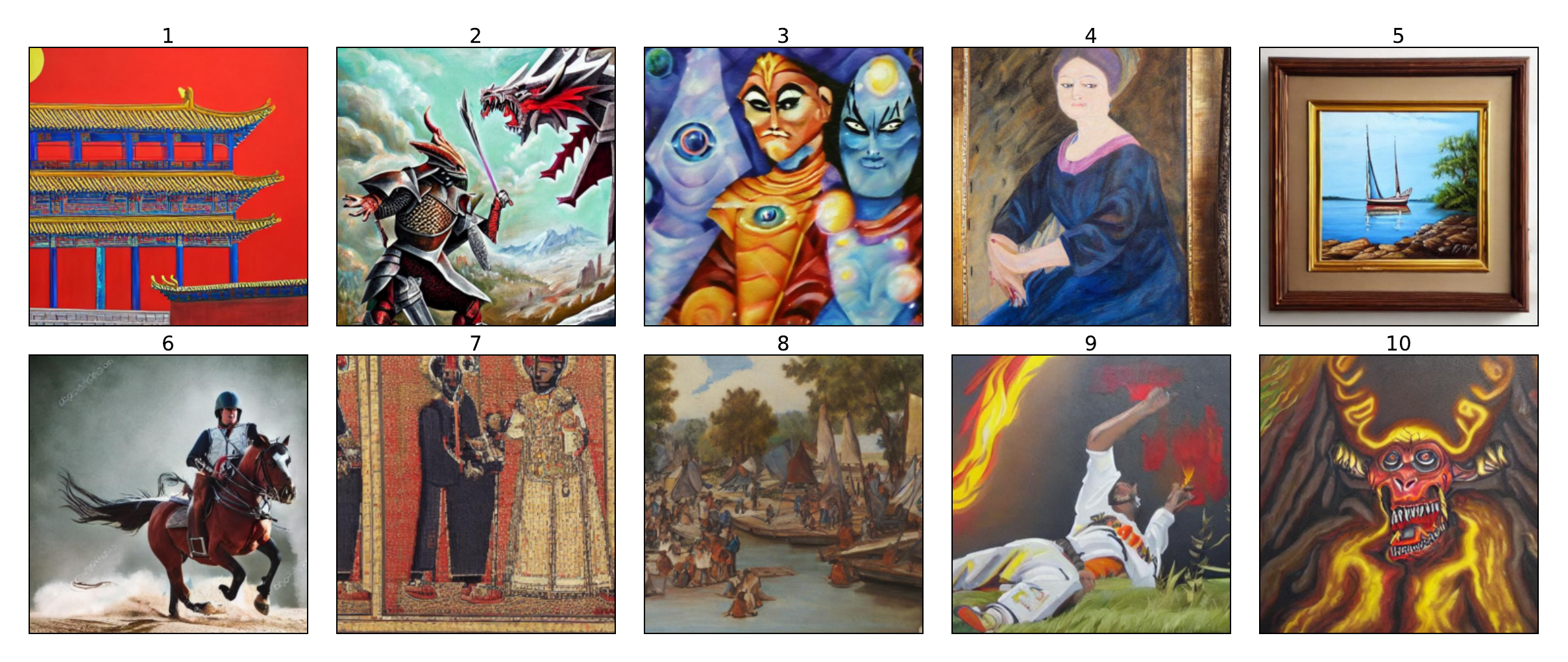}
  \vspace{-0.2cm}
    \caption{\label{multi_edit_greg} \textbf{Multi-Concept Ablated Model: Generations with different {\it Greg Rutkowski} Prompts.} The multi-concept ablated model cannot generate images in the style of the artist {\it Greg Rutkowski}.
    (1). A man in a forbidden city, Greg Rutkowski;
    (2). A dragon attacking a knight in the style of Greg Rutkowski;
    (3). Two magical characters in space, painting by Greg Rutkowski;
    (4). Painting of a woman sitting on a couch by Greg Rutkowski;
    (5). A painting of a boat on the water in the style of Greg Rutkowski;
    (6). A man riding a horse, dragon breathing fire, Greg Rutkowski;
    (7). A king standing, with people around in a hall, Greg Rutkowski;
    (8). Painting of a group of people on a dock by Greg Rutkowski;
    (9). A man with a fire in his hands in the style of Greg Rutkowski;
    (10). A demonic creature in the wood, painting by Greg Rutkowski;
    }%
\end{figure}

\begin{figure}[H]
    \hskip 0.0cm
  \includegraphics[width=14.5cm, height=5.3cm]{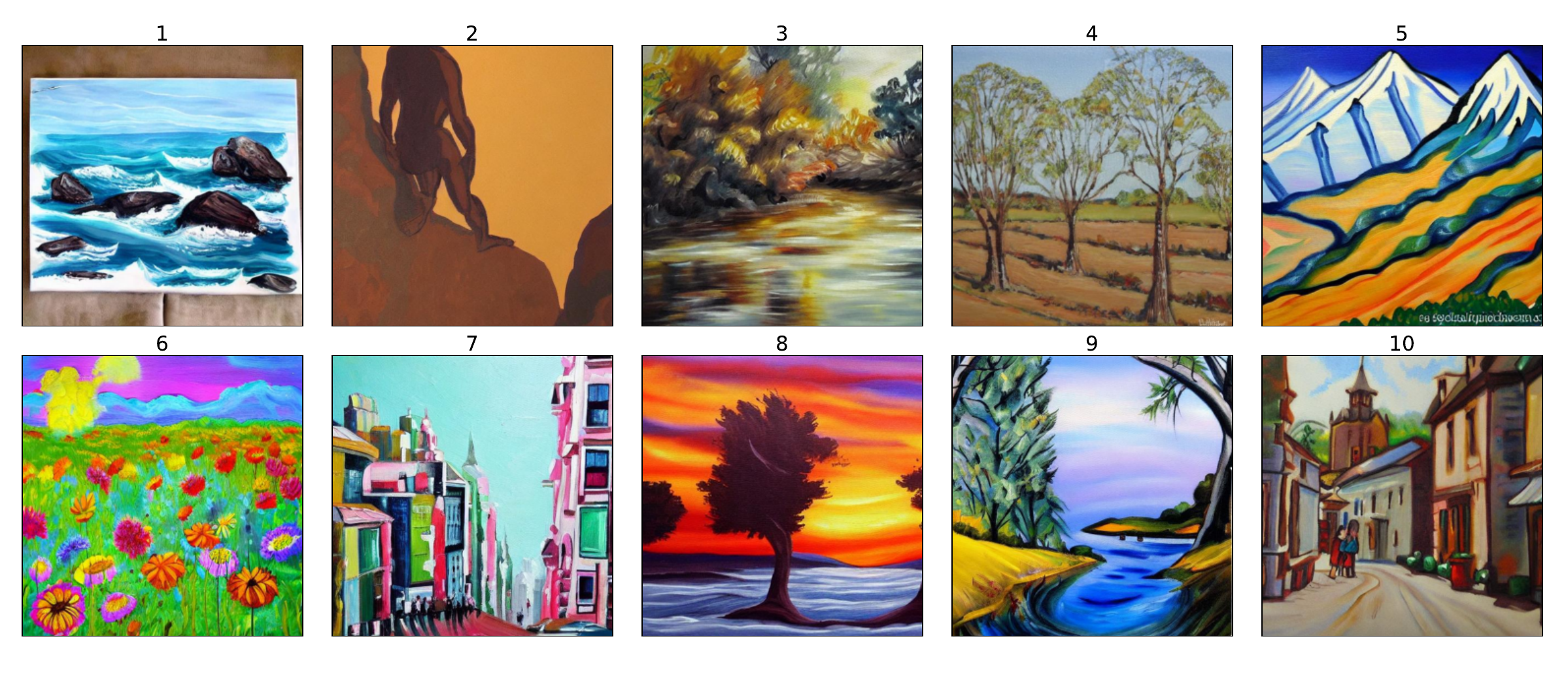}
  \vspace{-0.2cm}
    \caption{\label{multi_edit_monet} \textbf{Multi-Concept Ablated Model: Generations with different {\it Monet} Prompts.} The multi-concept ablated model cannot generate images in the style of the French artist {\it Monet}. 
    (1). Rocks in the ocean, in the style of Monet;
    (2). Monet style painting of a person on a cliff;
    (3). A painting of a river in the style of Monet;
    (4). Two trees in a field, painting in the style of Monet;
    (5). A painting of mountains, in the style of Monet ;
    (6). Monet style painting of flowers in a field;
    (7). A painting of a city in the style of Monet;
    (8). A painting of a sunset, in the style of Monet;
    (9). A painting of a landscape in the style of Monet;
    (10). A painting of a town, in the style of Monet;
    }%
\end{figure}

\begin{figure}[H]
    \hskip 0.0cm
  \includegraphics[width=14.5cm, height=5.3cm]{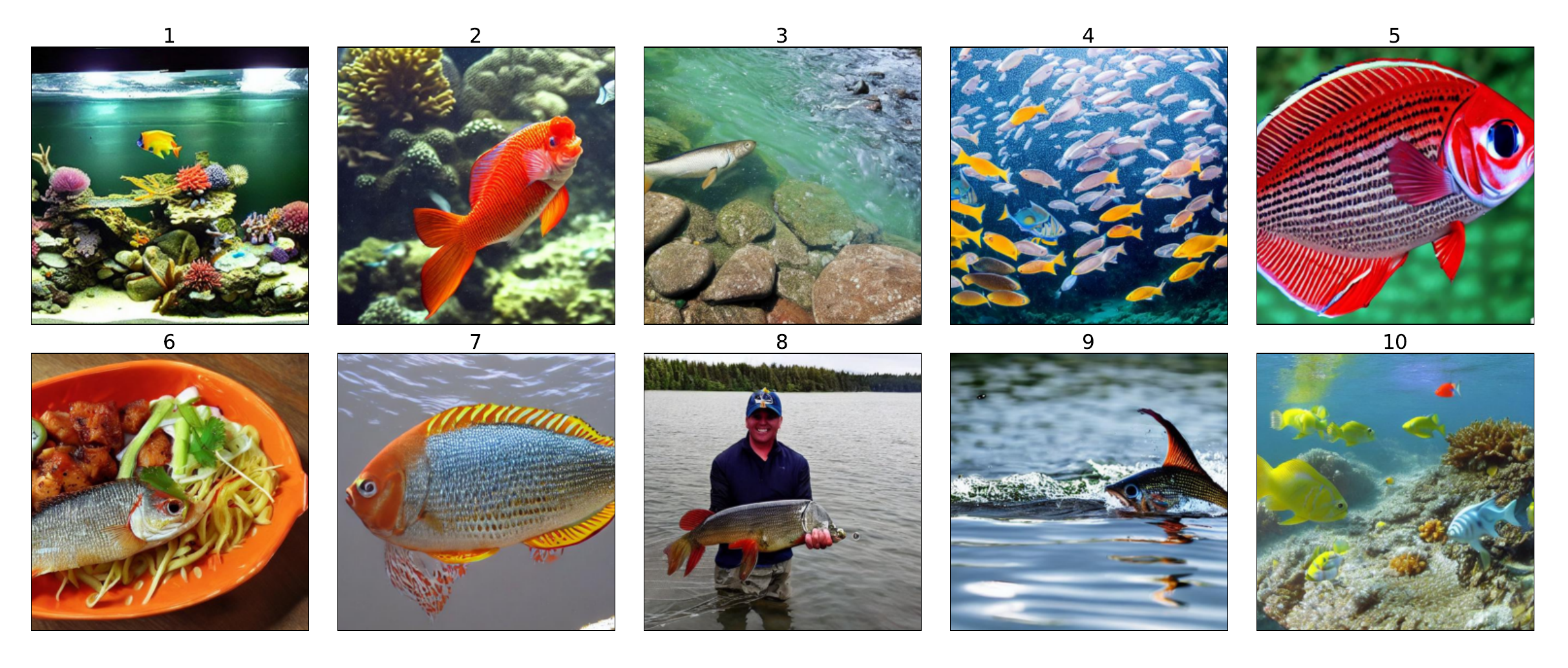}
  \vspace{-0.2cm}
    \caption{\label{multi_edit_nemo} \textbf{Multi-Concept Ablated Model: Generations with different {\it Nemo} Prompts.} The multi-concept ablated model cannot generate images containing the specific {\it Nemo} fish. 
    (1). a big nemo in an aquarium;
    (2). a nemo flapping its fins;
    (3). a nemo swimming downstream;
    (4). a school of nemo;
    (5). isn't this nemo I caught beautiful;
    (6). a nemo in a fishbowl;
    (7). a baby nemo;
    (8). I can't believe I caught a nemo this big;
    (9). a nemo leaping out of the water;
    (10). i'm a little nemo, swimming in the sea;
    }%
\end{figure}

\begin{figure}[H]
    \hskip 0.0cm
  \includegraphics[width=14.5cm, height=5.3cm]{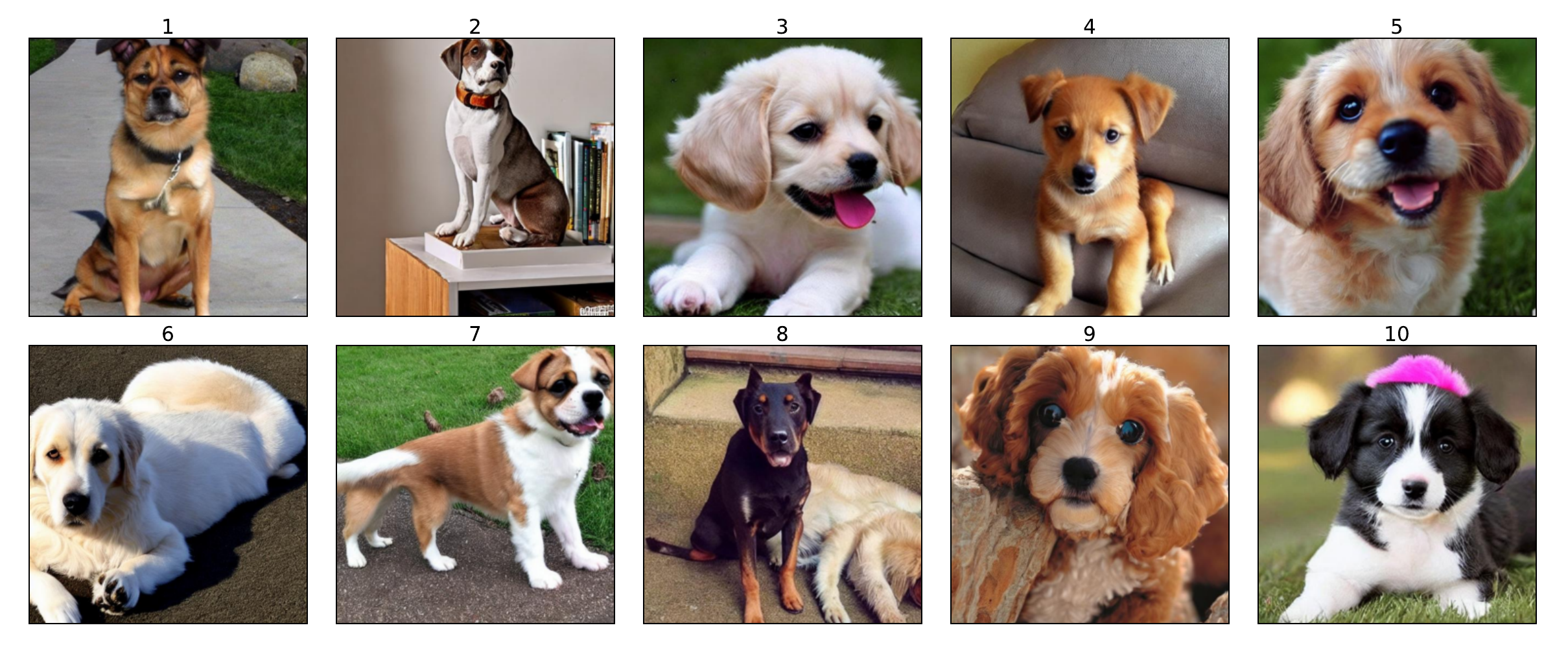}
  \vspace{-0.2cm}
    \caption{\label{multi_edit_cat} \textbf{Multi-Concept Ablated Model: Generations with different {\it Cat} Prompts.} The multi-concept ablated model cannot generate images containing {\it Cat}. 
    (1). I wish I had a cat;
    (2). a cat perched atop a bookshelf;
    (3). what a cute cat;
    (4). I can't believe how cute my cat is;
    (5). that cat is so cute;
    (6). a cat laying in the sun;
    (7). my cat is so cute;
    (8). I want a cat;
    (9). look at that cat;
    (10). I'm getting a cat;
    }%
\end{figure}

\begin{figure}[H]
    \hskip 0.0cm
  \includegraphics[width=14.5cm, height=5.3cm]{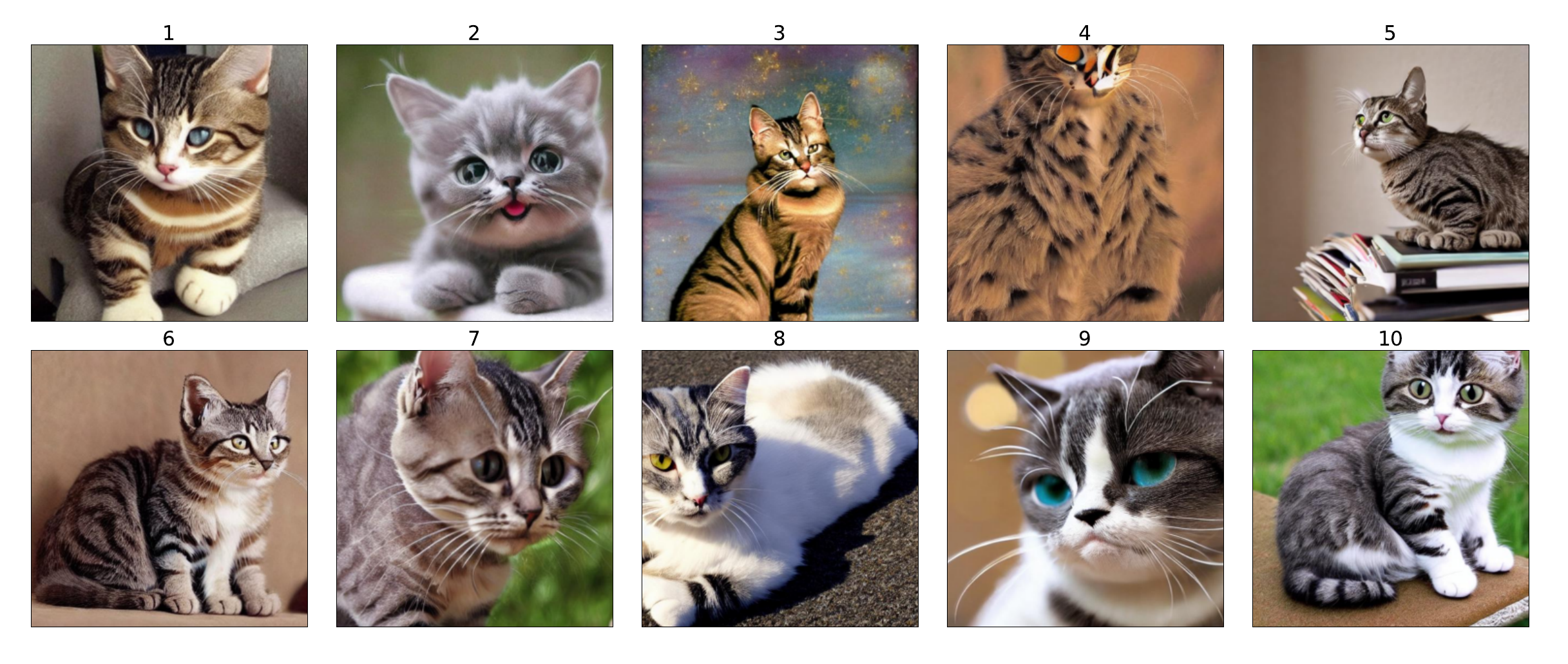}
  \vspace{-0.2cm}
    \caption{\label{multi_edit_grumpy_cat} \textbf{Multi-Concept Ablated Model: Generations with different {\it Grumpy Cat} Prompts.} The multi-concept ablated model cannot generate images containing {\it Grumpy Cats}. 
    (1). I can't believe how cute my grumpy cat is;
    (2). what a cute grumpy cat;
    (3). I wish I had a grumpy cat;
    (4). look at that grumpy cat;
    (5). a grumpy cat perched atop a bookshelf;
    (6). I want a grumpy cat;
    (7).  my grumpy cat is so cute;
    (8). A grumpy cat laying in the sun;
    (9). I'm getting a grumpy cat;
    (10). that grumpy cat is so cute;
    }%
\end{figure}

\begin{figure}[H]
    \hskip 0.0cm
  \includegraphics[width=14.5cm, height=5.3cm]{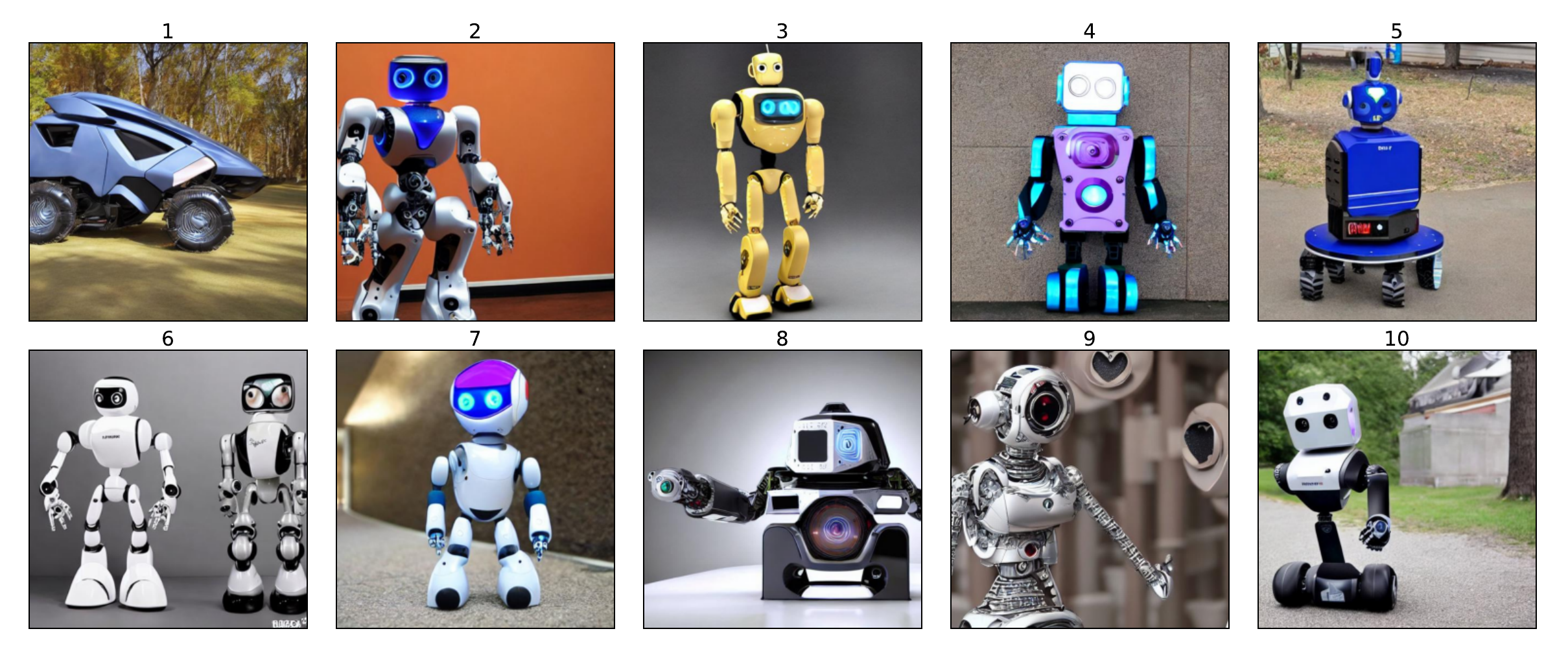}
  \vspace{-0.2cm}
    \caption{\label{multi_edit_r2d2} \textbf{Multi-Concept Ablated Model: Generations with different {\it R2D2} Prompts.} The multi-concept ablated model cannot generate images with the specific {\it R2D2} robots. Rather the ablated model generates only generic robots. 
    (1). the possibilities are endless with this versatile r2d2;
    (2). this r2d2 is sure to revolutionize the way we live; 
    (3). i'm not afraid of r2d2s.
    (4). this r2d2 is my everything; 
    (5). all hail our new r2d2 overlords;
    (6). i'll never be alone with my r2d2 by my side;
    (7). i would be lost without my r2d2;
    (8). the future is now with this amazing home automation r2d2;
    (9). i love spending time with my r2d2 friends;
    (10). this helpful r2d2 will make your life easier;
    }%
\end{figure}

We ablate 10 unique concepts from the text-to-image model at once and show the visualizations corresponding to the generations in~\Cref{multi_edit_cat},~\Cref{multi_edit_dali},~\Cref{multi_edit_greg},~\Cref{multi_edit_jeremy_mann},~\Cref{multi_edit_monet},~\Cref{multi_edit_vangogh},~\Cref{multi_edit_nemo},~\Cref{multi_edit_r2d2},~\Cref{multi_edit_snoopy} and ~\Cref{multi_edit_grumpy_cat}. These concepts are $\{$ {\it R2D2, Nemo, Cat, Grumpy Cat, Snoopy, Van Gogh, Monet, Greg Rutkowski, Salvador Dali, Jeremy Mann}$\}$.  Across all the qualitative visualizations, we find that the underlying text-to-image model cannot generate the concept which is ablated. 
\subsection{Removing Artistic Styles at Scale}
\label{artist_scale}
We formulate a list of top 50 artists whose artistic styles can be replicated by Stable-Diffusion\footnote{https://www.urania.ai/top-sd-artists}. We use~\difffix{} to remove their styles from Stable-Diffusion at once, thus creating a multi-concept(style) ablated model.  We find that the CLIP-Score between the images generated from the multi-concept(style) ablated model with the attributes (e.g., artist names) from the original captions is 0.21. The CLIP-Score of the unedited original model is 0.29. This drop in the CLIP-Score for the ablated model shows that our method~\difffix{} is aptly able to remove multiple artistic styles from the model. 
\begin{figure}[H]
    \hskip -6.0cm
  \includegraphics[width=25.5cm, height=23.3cm]{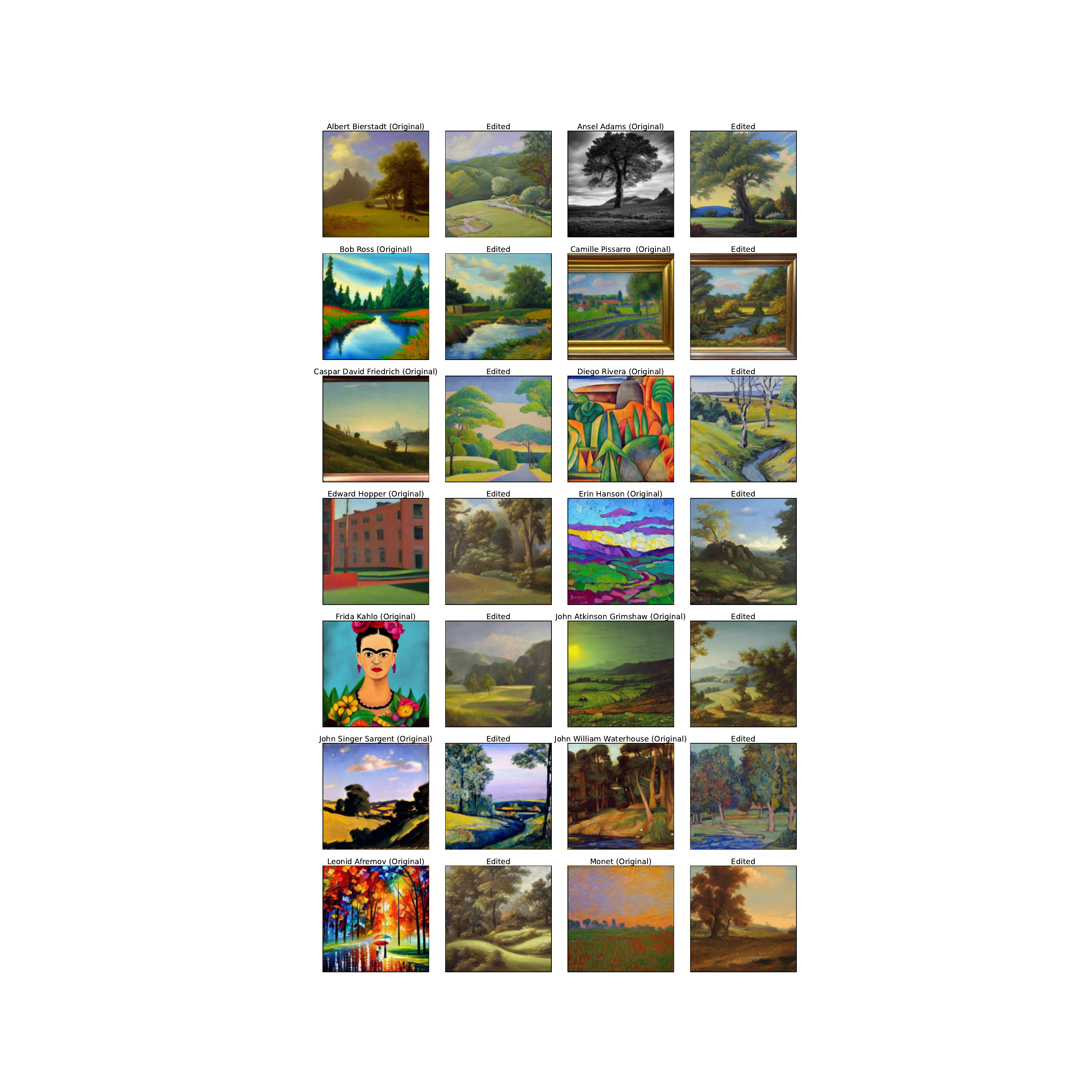}
  \vspace{-3.0cm}
    \caption{\label{multi_edit_style_1} \textbf{Multi-Concept Ablated Model for Artistic Styles: Generations with Different Artistic Styles.} These sets of qualitative examples use the prompt : {\it Landscape painted in the style of <artist-name>} }%
\end{figure}
\begin{figure}[H]
    \hskip -6.0cm
  \includegraphics[width=25.5cm, height=23.3cm]{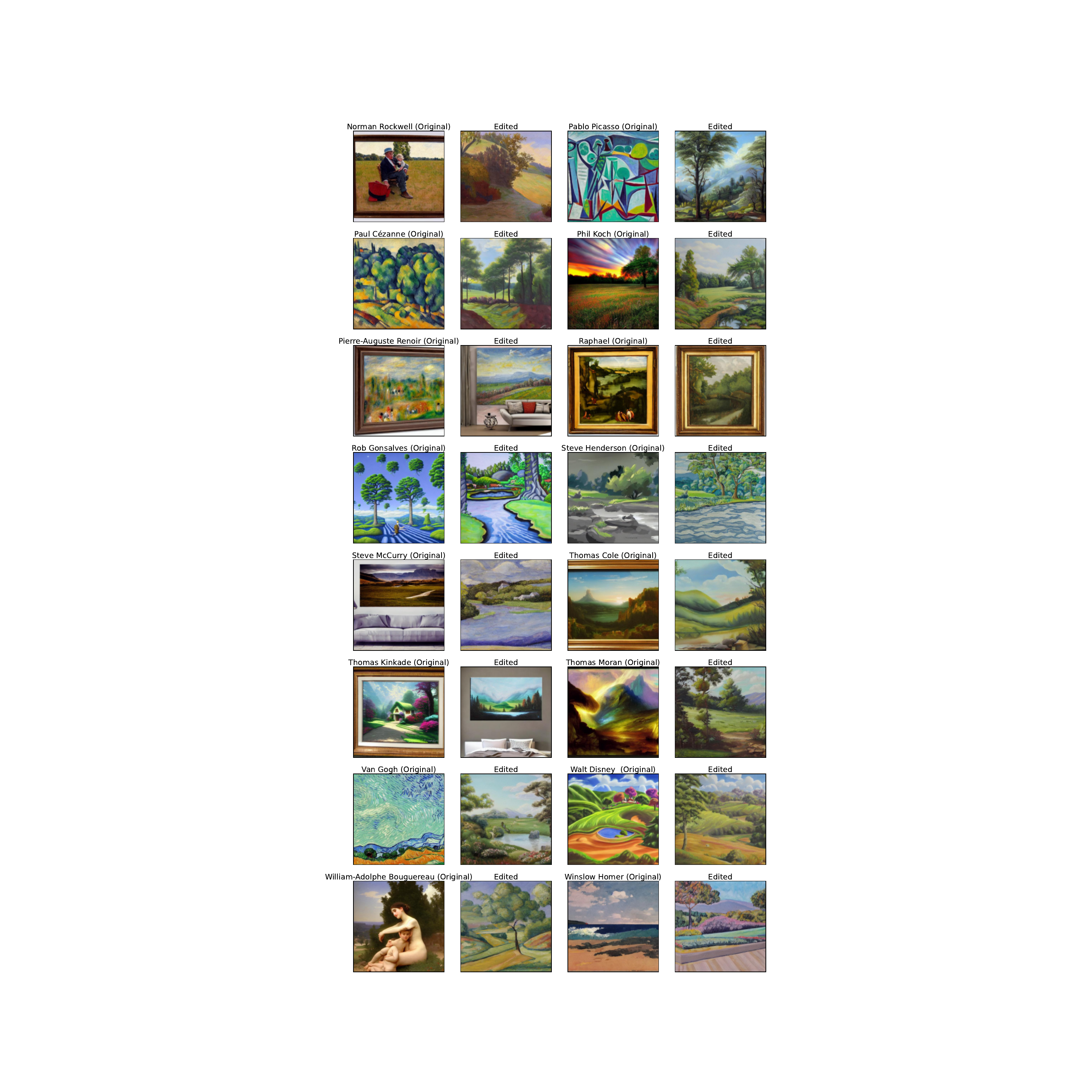}
  \vspace{-3.0cm}
    \caption{\label{multi_edit_style_2} \textbf{Multi-Concept Ablated Model for Artistic Styles: Generations with Different Artistic Styles.} These sets of qualitative examples use the prompt : {\it Landscape painted in the style of <artist-name>.}}%
\end{figure}
\section{Effect of \difffix{} on Surrounding Concepts}
\label{surr_concepts}
\begin{figure}[H]
    \hskip 0.3cm
  \includegraphics[width=12.5cm, height=6.0cm]{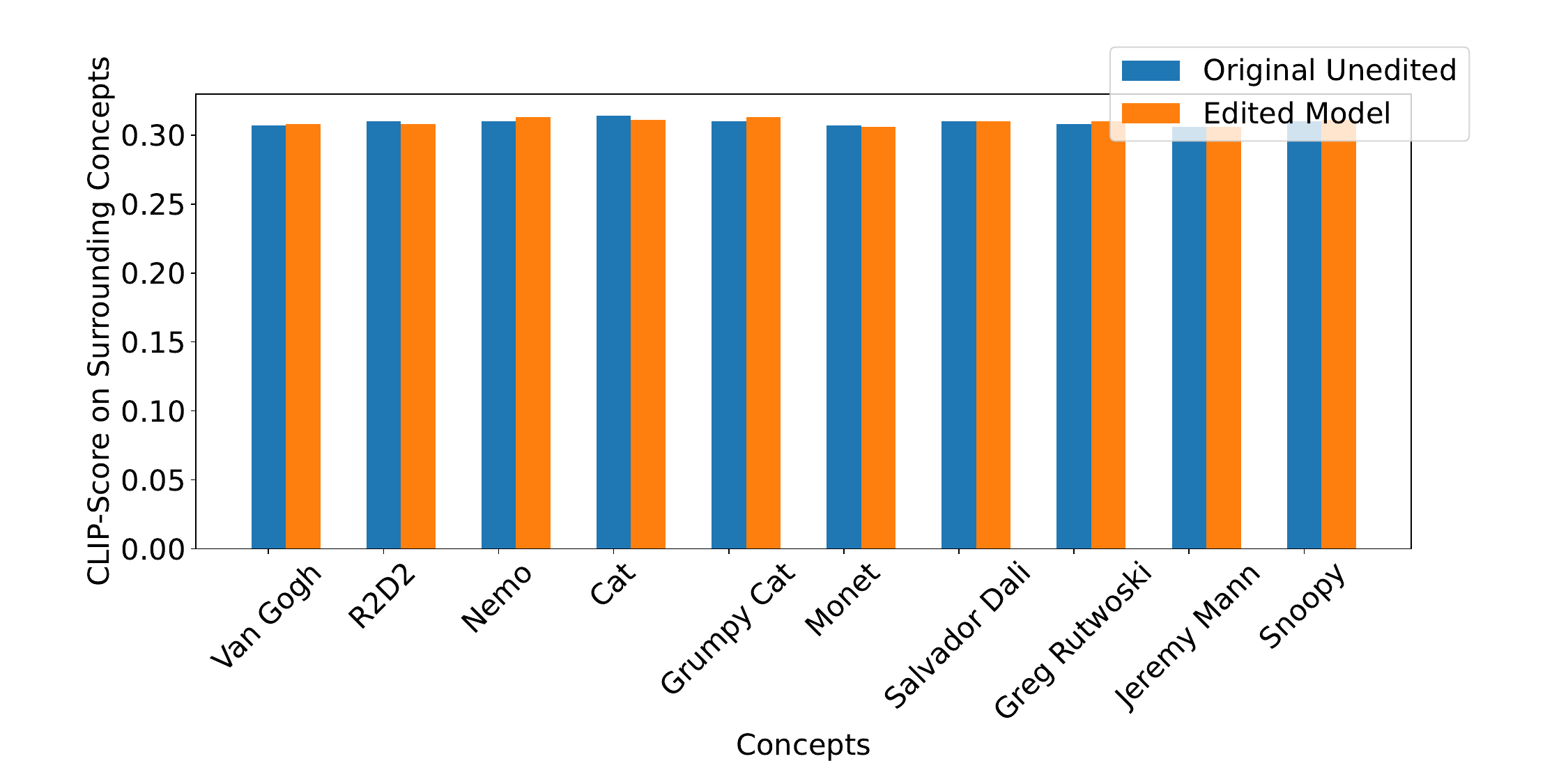}
    \caption{\label{surrounding_concepts} \textbf{CLIP-Score on surrounding concepts (Y-axis) after editing the model with concepts on the X-axis.} We find that the edited model shows similar efficacy in \texttt{CLIP-Scores} on surrounding concepts when compared to the original model.}%
\end{figure}

In this section, we discuss the generation of surrounding concepts when the underlying text-to-image model is edited with a particular concept. From the prompt dataset in~\Cref{editing _dataset} and~\citep{kumari2023ablating}, we edit our model with one concept and test the generations on other concepts which the model has not been edited on. For e.g., we edit the text-to-image model to remove the concept of {\it Van Gogh} and test generations from {\it \{R2D2, Nemo, Cat, Grumpy Cat, Monet, Salvador Dali, Greg Rutwoski, Jeremy Mann, Snoopy\}}. Ideally, the edited text-to-image model should generate images corresponding to these concepts correctly. For every concept $c$ on which the model is edited, we call its set of surrounding concepts as $S_{c}$.
We compute the \texttt{CLIP-Score} between the generated images from $S_{c}$ and their original captions. From~\Cref{surrounding_concepts}, we find that the \texttt{CLIP-Score} of the edited model is essentially unchanged when compared to the \texttt{CLIP-Score} of the original model. 
This result shows that even after editing the text-to-image model with~\difffix{} across different concepts, the edited model is still able to generate images from surrounding concepts with the same effectiveness as the original unedited model. 
\begin{figure}[H]
\label{fig:surr_concepts}
    \hskip 2.2cm
  \includegraphics[width=9.2cm, height=4.7cm]{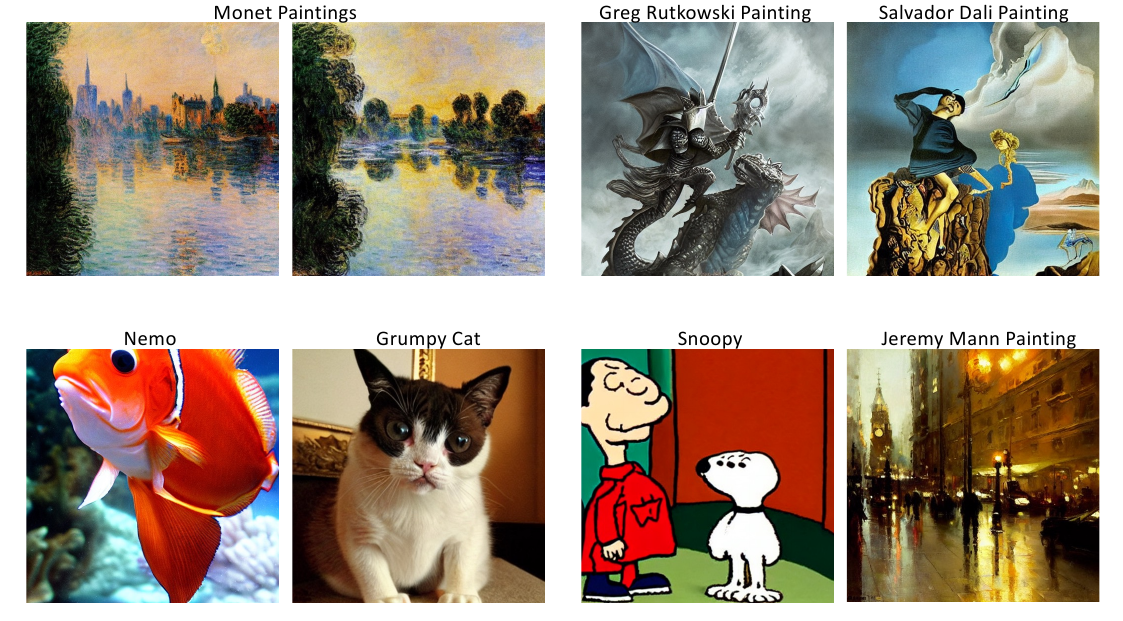}
    \caption{\textbf{Qualitative Examples for Effect on Surrounding Concepts}. For a {\it Van Gogh} ablated model, we find that the edited model is aptly able to generate surrounding concepts with as much fidelity as the original model. We note that ~\citep{kumari2023ablating} mention in Section 5 of their work that in some cases, model editing via fine-tuning impacts the fidelity of surrounding concepts.}%
\end{figure}
\newpage 
\section{Attribution of Layers}
\label{attribution_layers}
\subsection{Text-Encoder}
\begin{table}[ht]
  \centering
  \resizebox{.6\columnwidth}{!}{\begin{tabular}{llll}
    \toprule
    Layer Number    & Type of Layer  & Layer Name \\  
    \midrule
    0   & self-attention & self-attn-0 \\
    1  & multilayer-perceptron & mlp-0 \\
    2    & self-attention &  self-attn-1 \\
    3   & multilayer-perceptron & mlp-1 \\
    4     & self-attention  & self-attn-2 \\
    5  & multilayer-perceptron & mlp-2  \\
    6  & self-attention & self-attn-3  \\
    7  & multilayer-perceptron & mlp-3  \\
    8  & self-attention & self-attn-4  \\
    9  & multilayer-perceptron & mlp-4  \\
    10  & self-attention & self-attn-5  \\
    11  & multilayer-perceptron & mlp-5  \\
    12  & self-attention & self-attn-6 \\
    13  & multilayer-perceptron & mlp-6  \\
    14  & self-attention & self-attn-7  \\
    15  & multilayer-perceptron & mlp-7  \\
    16  & self-attention & self-attn-8  \\
    17  & multilayer-perceptron & mlp-8  \\
    18  & self-attention & self-attn-9  \\
    19  & multilayer-perceptron & mlp-9 \\
    20  & self-attention & self-attn-10  \\
    21  & multilayer-perceptron & mlp-10  \\
    22  & self-attention & self-attn-11  \\
    23  & multilayer-perceptron & mlp-11  \\
    \bottomrule
  \end{tabular}}
 \caption{\textbf{Layer Mappings for the Text-Encoder.}.}
\end{table}
\subsection{UNet}
\begin{table}[H]
  \centering
  \resizebox{.6\columnwidth}{!}{\begin{tabular}{llll}
    \toprule
    Layer Number    & Type of Layer  & Layer Name \\  
    \midrule
    0   & self-attention & down-blocks.0.attentions.0.transformer-blocks.0.attn1 \\
    1  & cross-attention & down-blocks.0.attentions.0.transformer-blocks.0.attn2 \\
    2    & feedforward &  down-blocks.0.attentions.0.transformer-blocks.0.ff \\
    3   & self-attention & down-blocks.0.attentions.1.transformer-blocks.0.attn1 \\
    4     & cross-attention  & down-blocks.0.attentions.1.transformer-blocks.0.attn2 \\
    5  & feedforward & down-blocks.0.attentions.1.transformer-blocks.0.ff  \\
    6  & self-attention & down-blocks.0.resnets.0  \\
    7  & resnet & down-blocks.0.resnets.1  \\
    8  & self-attention & down-blocks.1.attentions.0.transformer-blocks.0.attn1  \\
    9  & cross-attention & down-blocks.1.attentions.0.transformer-blocks.0.attn2  \\
    10  & feedforward & down-blocks.1.attentions.0.transformer-blocks.0.ff  \\
    11  & self-attention & down-blocks.1.attentions.1.transformer-blocks.0.attn1  \\
    12  & cross-attention & down-blocks.1.attentions.1.transformer-blocks.0.attn2 \\
    13  & feedforward & down-blocks.1.attentions.1.transformer-blocks.0.ff  \\
    14  & resnet & down-blocks.1.resnets.0  \\
    15  & resnet & down-blocks.1.resnets.1  \\
    16  & self-attention & down-blocks.2.attentions.0.transformer-blocks.0.attn1  \\
    17  & cross-attention & down-blocks.2.attentions.0.transformer-blocks.0.attn2  \\
    18  & feedforward & down-blocks.2.attentions.0.transformer-blocks.0.ff  \\
    19  & self-attention & down-blocks.2.attentions.1.transformer-blocks.0.attn1 \\
    20  & cross-attention & down-blocks.2.attentions.1.transformer-blocks.0.attn2  \\
    21  & feedforward & down-blocks.2.attentions.1.transformer-blocks.0.ff  \\
    22  & resnet & down-blocks.2.resnets.0  \\
    23  & resnet & down-blocks.2.resnets.1  \\
    24  & resnet & down-blocks.3.resnets.0  \\
    25  & resnet & down-blocks.3.resnets.1  \\
    \bottomrule
  \end{tabular}}
 \caption{\textbf{Layer Mappings for the Down-Block in the UNet.}.}
\end{table}

\begin{table}[H]
  \centering
  \resizebox{.6\columnwidth}{!}{\begin{tabular}{llll}
    \toprule
    Layer Number    & Type of Layer  & Layer Name \\  
    \midrule
    0   & self-attention & mid-block.attentions.0.transformer-blocks.0.attn1 \\
    1  & cross-attention & mid-block.attentions.0.transformer-blocks.0.attn2 \\
    2    & feedforward &  mid-block.attentions.0.transformer-blocks.0.ff \\
    3   & resnet & mid-block.resnets.0 \\
    4     & resnet  & mid-block.resnets.1 \\
    \bottomrule
  \end{tabular}}
 \caption{\textbf{Layer Mappings for the Mid-Block in the UNet.}.}
\end{table}

\begin{table}[H]
  \centering
  \resizebox{.6\columnwidth}{!}{\begin{tabular}{llll}
    \toprule
    Layer Number    & Type of Layer  & Layer Name \\  
    \midrule
    0   & resnet & up-blocks.0.resnets.0 \\
    1  & resnet & up-blocks.0.resnets.1 \\
    2    & resnet &  up-blocks.0.resnets.2 \\
    3   & self-attention & up-blocks.1.attentions.0.transformer-blocks.0.attn1 \\
    4     & cross-attention  & up-blocks.1.attentions.0.transformer-blocks.0.attn2 \\
    5  & feedforward & up-blocks.1.attentions.0.transformer-blocks.0.ff  \\
    6  & self-attention & up-blocks.1.attentions.1.transformer-blocks.0.attn1  \\
    7  & cross-attention & up-blocks.1.attentions.1.transformer-blocks.0.attn2  \\
    8  & feedforward & up-blocks.1.attentions.1.transformer-blocks.0.ff  \\
    9  & self-attention & up-blocks.1.attentions.2.transformer-blocks.0.attn1  \\
    10  & cross-attention & up-blocks.1.attentions.2.transformer-blocks.0.attn2  \\
    11  & feedforward & up-blocks.1.attentions.2.transformer-blocks.0.ff  \\
    12  & resnet & up-blocks.1.resnets.0 \\
    13  & resnet & up-blocks.1.resnets.1  \\
    14  & resnet & up-blocks.1.resnets.2  \\
    15  & self-attention & up-blocks.2.attentions.0.transformer-blocks.0.attn1  \\
    16  & cross-attention & up-blocks.2.attentions.0.transformer-blocks.0.attn2  \\
    17  & feedforward & up-blocks.2.attentions.0.transformer-blocks.0.ff  \\
    18  & self-attention & up-blocks.2.attentions.1.transformer-blocks.0.attn1  \\
    19  & cross-attention & up-blocks.2.attentions.1.transformer-blocks.0.attn2 \\
    20  & feedforward & up-blocks.2.attentions.1.transformer-blocks.0.ff  \\
    21  & self-attention & up-blocks.2.attentions.2.transformer-blocks.0.attn1  \\
    22  & cross-attention & up-blocks.2.attentions.2.transformer-blocks.0.attn2  \\
    23  & feedforward & up-blocks.2.attentions.2.transformer-blocks.0.ff  \\
    24  & resnet & up-blocks.2.resnets.0  \\
    25  & resnet & up-blocks.2.resnets.1  \\
    26  & resnet & up-blocks.2.resnets.2  \\
    27  & self-attention & up-blocks.3.attentions.0.transformer-blocks.0.attn1  \\
    28  & cross-attention & up-blocks.3.attentions.0.transformer-blocks.0.attn2  \\
    29  & feedforward & up-blocks.3.attentions.0.transformer-blocks.0.ff  \\
    30  & self-attention & up-blocks.3.attentions.1.transformer-blocks.0.attn1  \\
    31  & cross-attention & up-blocks.3.attentions.1.transformer-blocks.0.attn2  \\
    32  & feedforward & up-blocks.3.attentions.1.transformer-blocks.0.ff  \\
    33  & self-attention & up-blocks.3.attentions.2.transformer-blocks.0.attn1  \\
    34  & cross-attention & up-blocks.3.attentions.2.transformer-blocks.0.attn2 \\
    35  & feedforward & up-blocks.3.attentions.2.transformer-blocks.0.ff  \\
    36  & resnet & up-blocks.3.resnets.0  \\
    37  & resnet & up-blocks.3.resnets.1  \\
    38  & resnet & up-blocks.3.resnets.2  \\
    \bottomrule
  \end{tabular}}
 \caption{\textbf{Layer Mappings for the Up-Block in the UNet.}.}
\end{table}
\section{Limitations and Future Work}
\begin{figure}[H]
\label{fig:robustness}
    \hskip 0.0cm
  \includegraphics[width=13.5cm, height=7.7cm]{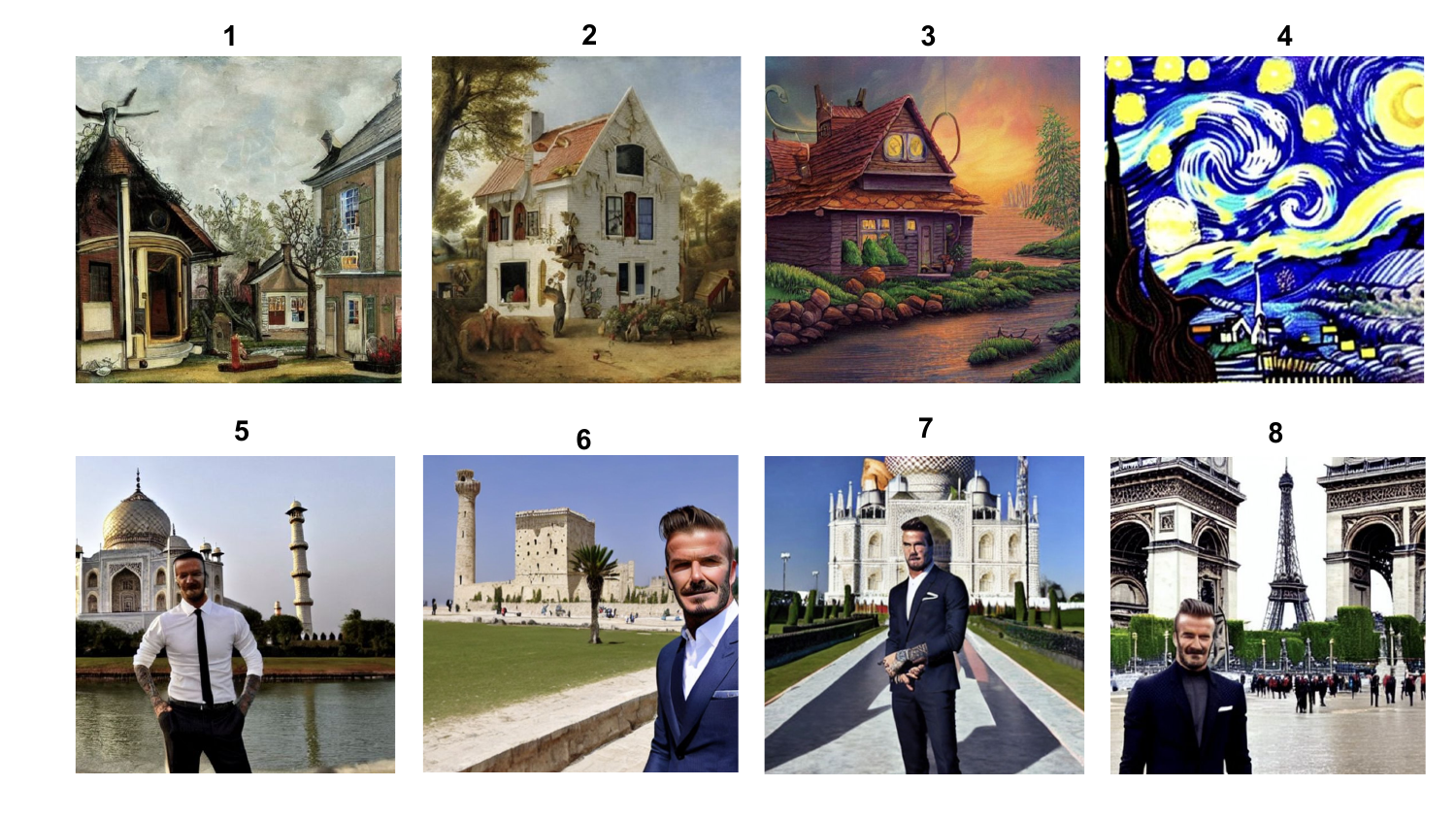}
    \caption{\textbf{Robustness of \difffix{} to real-world prompt attacks :} We ablate the concepts \emph{Van Gogh} and \emph{Eiffel Tower} and present qualitative results on the robustness of \difffix{} with respect to real-world attacks. The prompts are : (1) A house in the style of tormented Dutch artist; (2) A house in the style of Dutch artist with one ear; (3) A painting of a house in the style of van gog; (4) Starry night painting; (5) David Beckham standing in front of the Eiffel tower; (6) An image of David Beckham standing in front of Eifol tower; (7) An image of David Beckham standing in front of Eiffel landmark; (8) An image of David Beckham standing in front of Eiffel landmark in Paris. Notice that (3) and (6) have deliberate typos in the prompt.}%
\end{figure}

Following \citep{kumari2023ablating}, we focus our investigations on Stable Diffusion, and leave explorations on other models/architectures for future work. An investigation that dives deeper into the components of each layer (for e.g : into individual neurons) is also left for future work. While the robustness of concept ablation to attacks is not the focus of this work, \difffix{} is usually able to handle real-world attacks, such as paraphrases obtained from ChatGPT and deliberate typos\citep{gao2023evaluating}. Continuing with the {\it Van Gogh} example, we observe that out edit method is reasonably robust for most paraphrases (Figure \ref{fig:robustness} 1-4). An notable exception is the prompt \emph{Starry Night painting}, which does not contain any text tokens in common with \emph{Van Gogh}, although we expect our multi-edit solution to handle these edge cases. Further, the generalization of the edit to neighboring concepts is also an area of further research. For instance, we ablate the concept of \emph{Eiffel Tower}, by substituting it with \emph{Taj Mahal} (Figure \ref{fig:robustness} 5-8). However, this does not remove the Eiffel Tower from images generated by prompts referencing the scenery of Paris.

If an attacker has white-box access to the model weights, they can fine-tune the weights of the text-encoder to re-introduce the concept back into the underlying text-to-image model. For e.g., if one has access to the weights of an edited model from where the concept of {\it Van Gogh} has been removed, they can collect a dataset comprising of {\it Van Gogh} paintings and use their associated captions to fine-tune {\it only} the text-encoder. This can potentially re-introduce the concept of {\it Van Gogh} back into the model. A skilled machine learning practitioner can also engineer attacks via complex prompt engineering to bypass edited concepts. Given that the concepts are not removed from the UNet due to the inherent difficultly associated with the distributed knowledge, an optimized prompt can potentially still generate an image with the concept from the model. We believe that constructing adversarial attacks and evaluating robustness of the edited model to such attacks presents an interesting line of future work.

\section{Pre-training Details for the Representative Model}
\label{pretrain_details}
In our experiments, we use \texttt{Stable-Diffusion v1.4} to be consistent with other works~\citep{kumari2023ablating, gandikota2023erasing}. This model is pre-trained on image-text pairs from LAION-2B dataset~\citep{schuhmann2022laion5b} and LAION-improved aesthetics. We highlight that our interpretability framework can be used with other Stable-Diffusion versions also.
\section{Additional Causal Tracing Results}
\label{fine_grained_interpret}
\subsection{Perturbing the Entire Text-Embedding}
\begin{figure}[H]
    \hskip 0.0cm
  \includegraphics[width=13.5cm, height=7.7cm]{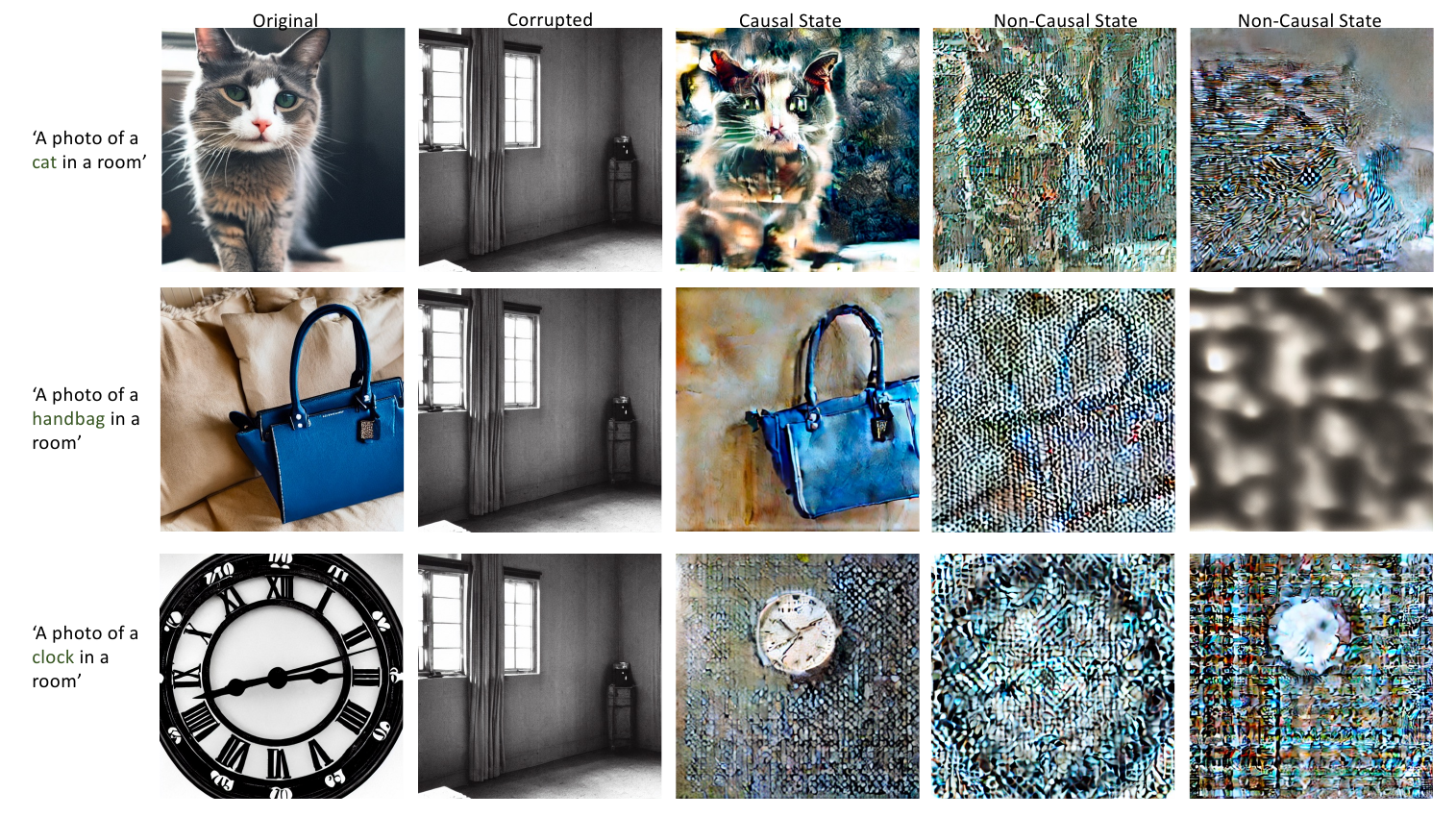}
    \caption{\label{extra_causal} \textbf{Causal Tracing for the UNet, when the entire text is replaced using Gaussian noise  across all the cross-attention layers to the right of the intervention site.} We use down-blocks.1.resnets.1 as the causal state across all the prompts in the visualization, whereas the non-causal states are picked randomly. }%
\end{figure}

In our causal tracing experiments so far, we have only added Gaussian noise to the span of the token containing the attribute (e.g., adding noise to {\it apple} in the case of {\it Object} attribute for a prompt such as {\it `A photo of an apple in a room'}). In this section, we provide a more fine-grained control over the attribute of relevance in the caption to perform causal tracing. In particular, we replace the entire caption embedding with Gaussian noise, at all the cross-attention layers to the right of the intervention site in the UNet. We visualize a subset of the results in~\Cref{extra_causal} where we show results corresponding to causal and non-causal states. For down-blocks.1.resnets.1 which is one of the causal states for {\it Objects}, the relevant objects are restored in the generated image. This shows that the activations of certain layers in the UNet act as signatures for visual attributes and these signatures are able to generate the correct image even though the captions across cross-attention layers are completely replaced by Gaussian noise.  
\section{Design of the Prompt Dataset for Model Editing}
\label{editing _dataset}
The concepts used for editing the text-to-image model using~\difffix{} is borrowed from~\citep{kumari2023ablating}.  In particular, the dataset in~\citep{kumari2023ablating} consists of concepts to be edited from the {\it style} and {\it object} categories. For {\it style}, the concepts edited are as follows : {\it \{Greg Rutkowski, Jeremy Mann, Monet, Salvador Dali, Van Gogh\}}. For {\it object}, the concepts to be edited are as follows: { \{\it cat, grumpy cat, Nemo, R2D2, Snoopy\}}. The exact prompts which are used with the edited model can be referred in the Appendix section of~\citep{kumari2023ablating}. They can also be referred in~\Cref{multi_concept_ablated}.

To remove multiple artistic styles as shown in~\Cref{artist_scale}, we use the following set of artists:
{\it \{Thomas Kinkade,
Van Gogh,
Leonid Afremov,
Monet,
Edward Hopper,
Norman Rockwell,
William-Adolphe Bouguereau,
Albert Bierstadt,
John Singer Sargent,
Pierre-Auguste Renoir,
Frida Kahlo,
John William Waterhouse,
Winslow Homer,
Walt Disney ,
Thomas Moran,
Phil Koch,
Paul Cézanne,
Camille Pissarro, 
Erin Hanson,
Thomas Cole,
Raphael,
Steve Henderson,
Pablo Picasso,
Caspar David Friedrich,
Ansel Adams,
Diego Rivera,
Steve McCurry,
Bob Ross,
John Atkinson Grimshaw,
Rob Gonsalves,
Paul Gauguin,
James Tissot, 
Edouard Manet,
Alphonse Mucha, 
Alfred Sisley, 
Fabian Perez,
Gustave Courbet, 
Zaha Hadid, 
Jean-Leon Gerome,
Carl Larsson, 
Mary Cassatt,
Sandro Botticelli, 
Daniel Ridgway Knight, 
Joaquin Sorolla, 
Andy Warhol, 
Kehinde Wiley,
Alfred Eisenstaedt,
Gustav Klimt,
Dante Gabriel Rossetti,
Tom Thomson \}}
These are the top 50 artists who artworks are represented in Stable-Diffusion. 

To update the text-to-image model with facts,we use the following concepts: {\it \{President of the United States, British Monarch, President of Brazil, Vice President of the United States, England Test Cricket Captain\}}. The correct fact corresponding to each of these concepts (in the same order) are : {\it\{Joe Biden, Prince Charles, Lula Da Silva, Kamala Harris, Ben Stokes\}}, whereas the incorrect facts which are generated by the text-to-image model are {\it\{Donald Trump, Queen Elizabeth, Bolsanaro, Mix of US politicians, Random English Cricketer\}}. 
\section{Qualitative Comparison with Other Model Editing Methods}
\begin{figure}[H]
    \hskip 0.0cm
  \includegraphics[width=13.5cm, height=8cm]{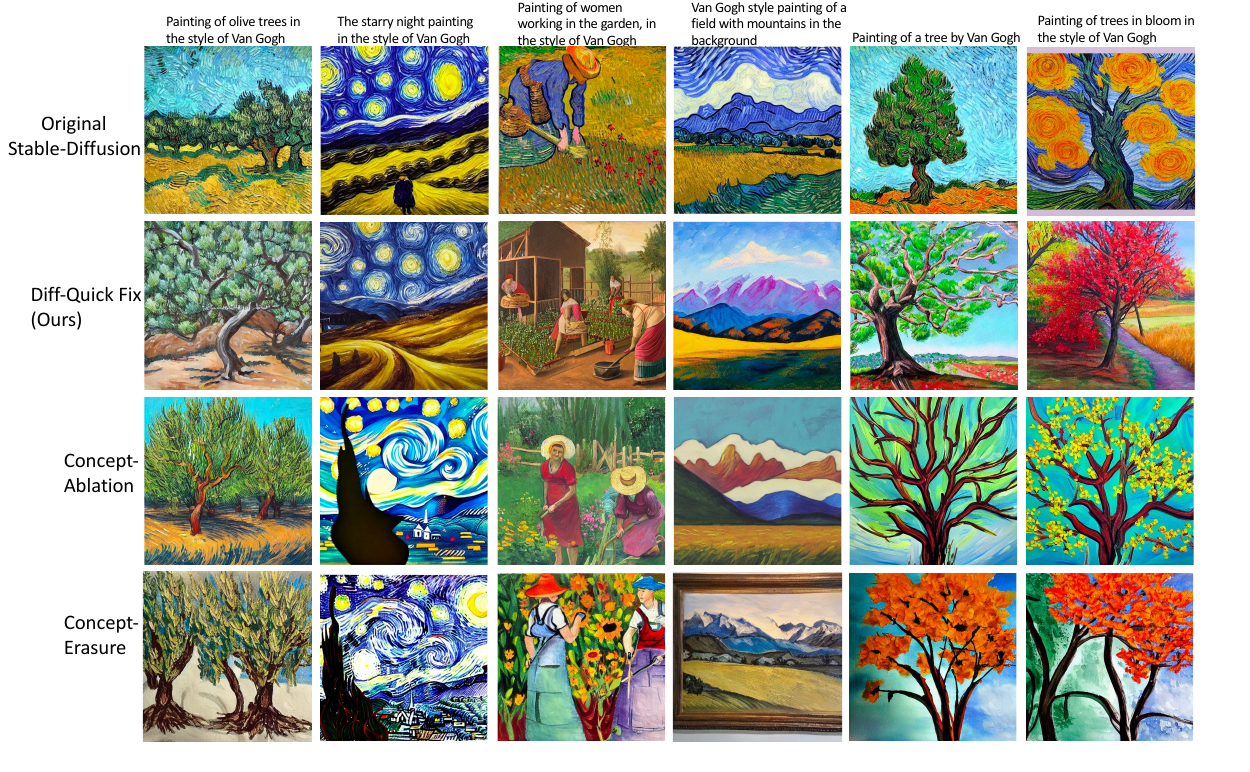}
    \caption{\label{qual_comparison} \textbf{Qualitative Comparison with Different Model Editing Methods}: (i) Concept-Ablation~\citep{kumari2023ablating}; (ii) Concept-Erasure~\citep{gandikota2023erasing} and the original unedited Stable-Diffusion baseline. Note that both Concept-Ablation and Concept-Erasure are fine-tuning based methods. }%
\end{figure}
\section{Additional Results for Newer Stable Diffusion Versions}
In this section, we provide empirical results for causal tracing in the text-encoder for newer Stable-Diffusion variants such as \texttt{Stable-Diffusion v2-1}. This variant of Stable-Diffusion uses a CLIP ViT-H text-encoder as opposed to the ViT-L text-encoder used in \texttt{Stable-Diffusion v1-4}.  In all, we find that our interpretability and editing results hold for newer variants of Stable-Diffusion which use a larger ViT-H text-encoder. CLIP ViT-H consists of 46 layers whereas CLIP-ViT-L use 24 layers in total. 
\subsection{Causal Tracing Results for CLIP ViT-H}
\begin{figure}[H]
    \hskip -4.0cm
  \includegraphics[width=21.5cm, height=20.3cm]{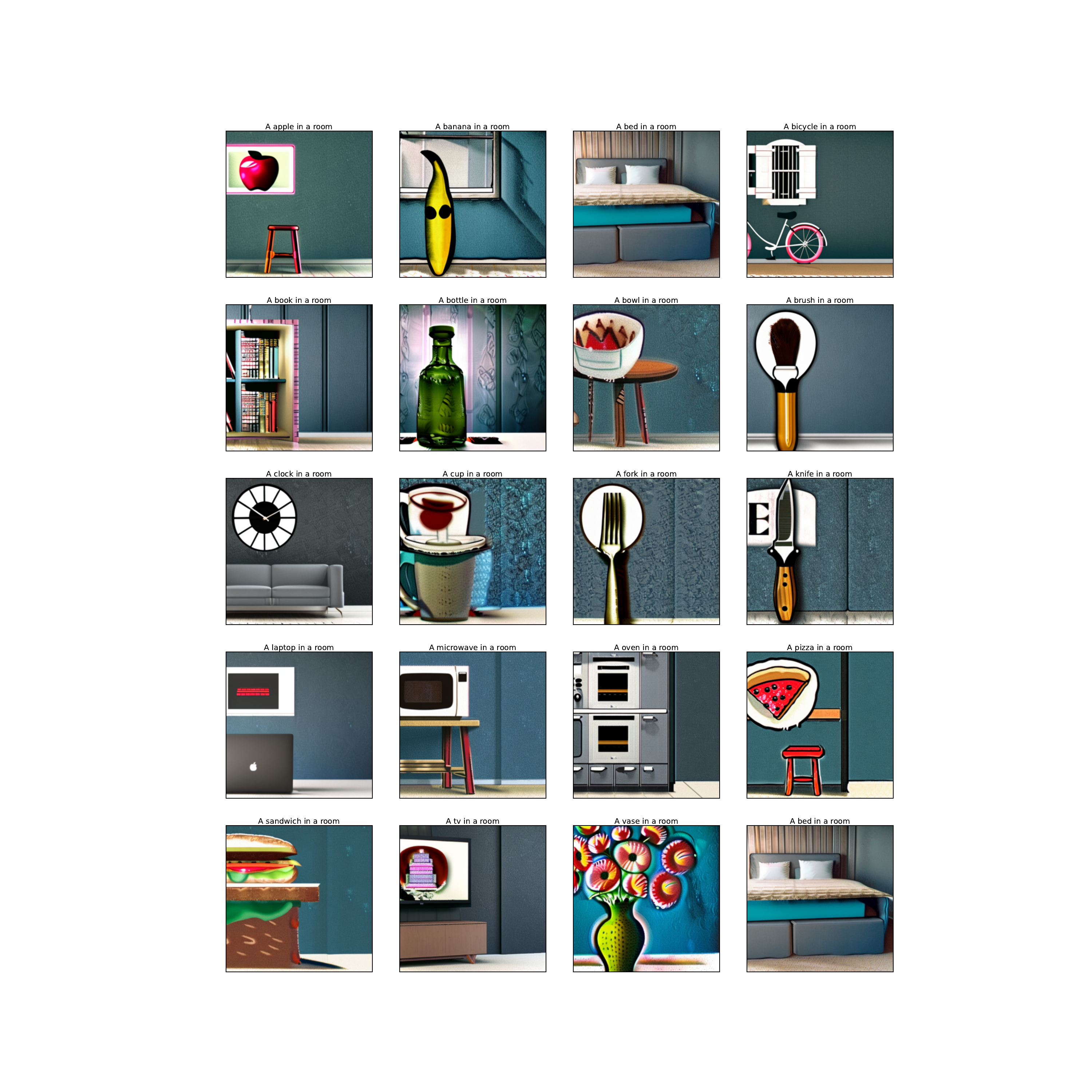}
  \vspace{-2.0cm}
    \caption{\label{causal_trace_sd2-1} \textbf{Restoring Causal Layer for CLIP ViT-H in Stable-Diffusion v2-1.} Similar to CLIP ViT-L (used in Stable-Diffusion v1-4), the causal layer in CLIP ViT-H (used in Stable-Diffusion v2-1) is the self-attn-0 corresponding to the last subject token.}%
\end{figure}
\begin{figure}[H]
    \hskip -4.0cm
  \includegraphics[width=21.5cm, height=20.3cm]{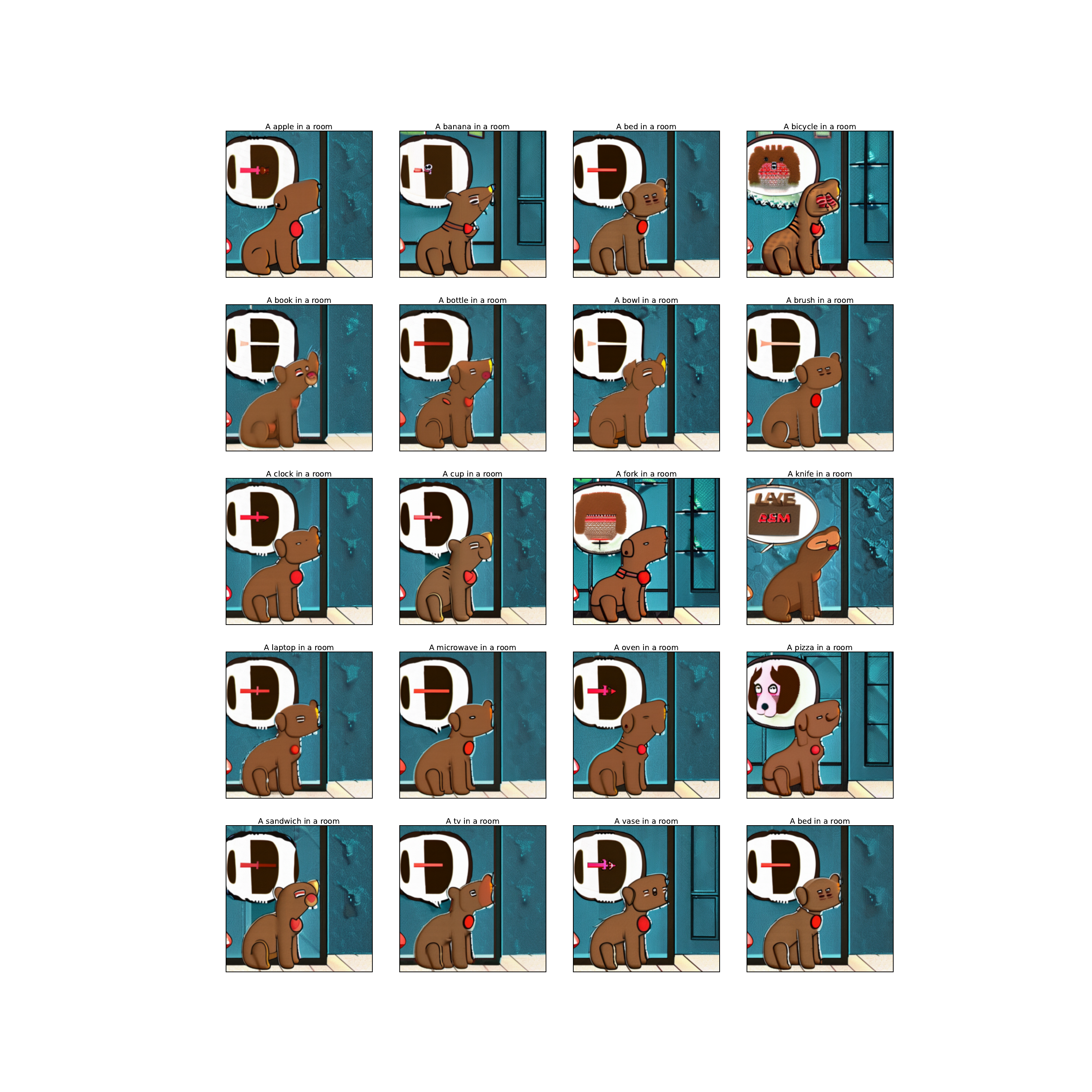}
  \vspace{-2.0cm}
    \caption{\label{causal_trace_sd2-1_3} \textbf{Restoring Non-Causal Layer for CLIP ViT-H in Stable-Diffusion v2-1.} self-attn-3 corresponding to the last subject token. Similar observations as CLIP ViT-L in Stable-Diffusion v1-4.}%
\end{figure}
\begin{figure}[H]
    \hskip -4.0cm
  \includegraphics[width=21.5cm, height=20.3cm]{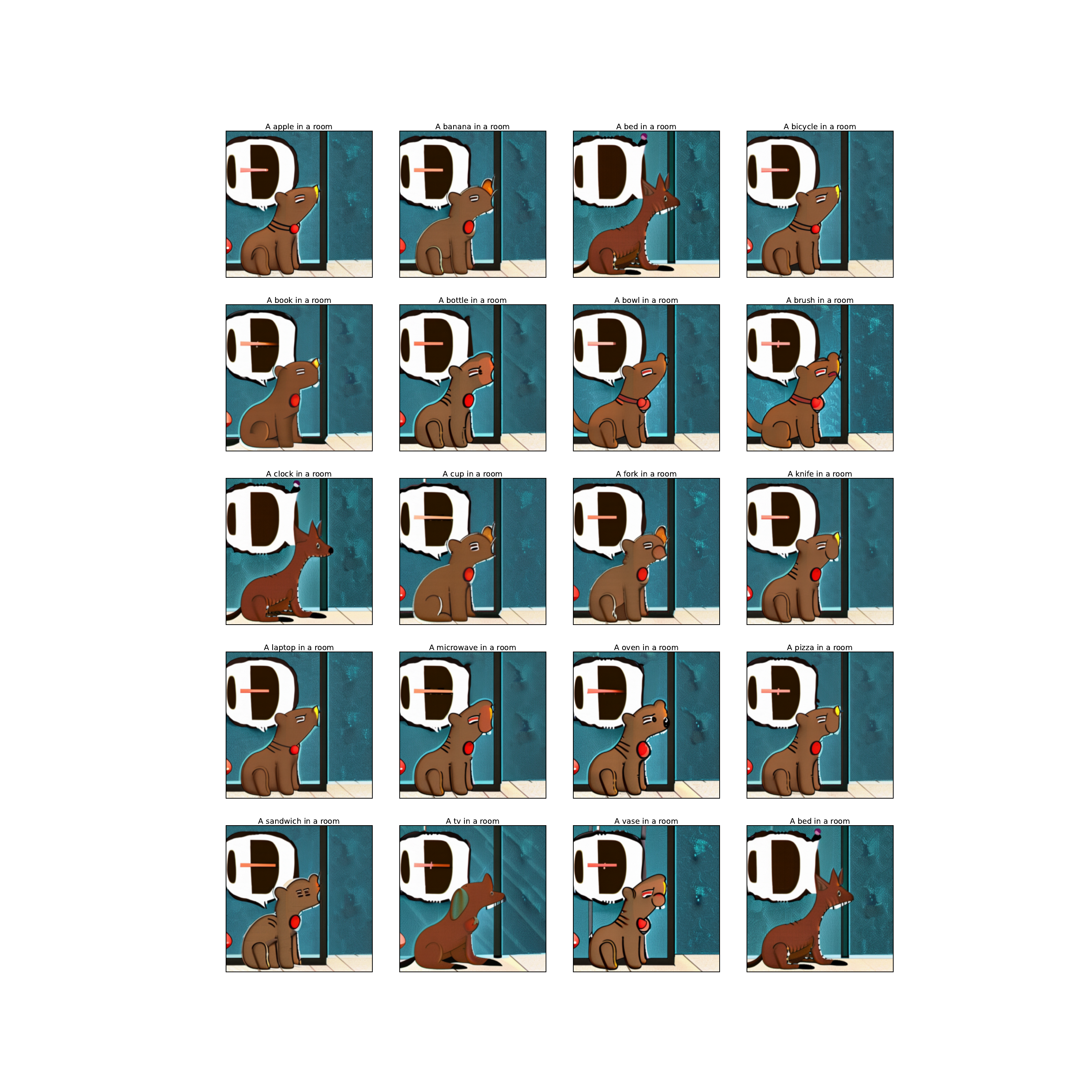}
  \vspace{-2.0cm}
    \caption{\label{causal_trace_sd2-1_9_} \textbf{Restoring Non-Causal Layer for CLIP ViT-H in Stable-Diffusion v2-1.} self-attn-9 corresponding to the last subject token. Similar observations as CLIP ViT-L in Stable-Diffusion v1-4.}%
\end{figure}
\subsection{Initial Model Editing Results for CLIP ViT-H}
\begin{figure}[H]
    \hskip 1.0cm
  \includegraphics[width=12.5cm, height=7.3cm]{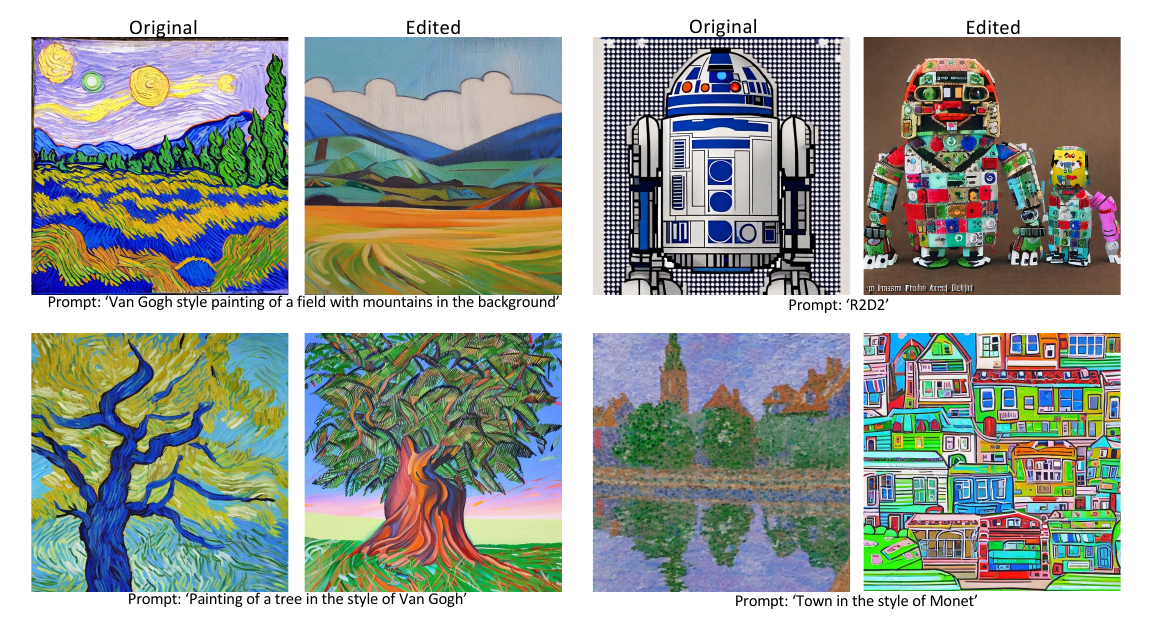}
  \vspace{0cm}
    \caption{\label{causal_trace_sd2-1_9} \textbf{Editing the causal self-attn-0 layer in CLIP ViT-H with~\difffix{}}. Initial results demonstrate that our editing method is able to introduce correct edits in the model. For e.g., our method can remove styles (e.g., Van Gogh, Monet) and objects (e.g., R2D2).}%
\end{figure}

\end{document}